%% file: main.tex
\documentclass[runningheads]{llncs}

\usepackage{style} 
\usepackage{abbrv} 
\usepackage[accsupp]{axessibility}  
\input{preamble}
 
\usepackage{orcidlink}

\begin{document}

\title{\texttt{Brain-ID}: Learning Contrast-agnostic Anatomical Representations for Brain Imaging}

\author{Peirong Liu\textsuperscript{1} \quad Oula Puonti\textsuperscript{1} \quad Xiaoling Hu\textsuperscript{1} \\ Daniel C. Alexander\textsuperscript{2} \quad Juan E. Iglesias\textsuperscript{1,2}
}
\institute{\textsuperscript{1}Harvard Medical School and Massachusetts General Hospital \quad \textsuperscript{2}UCL
}

\titlerunning{\texttt{Brain-ID}}
\authorrunning{Liu et al.}

\maketitle

\begin{center}
   \input{sec/showcase}

\end{center}

\input{sec/abstract}

\input{sec/intro}

\input{sec/related_work}

\input{sec/method/main}

\input{sec/exp/main}
\input{sec/con}


\bibliographystyle{splncs04}
\bibliography{reference}

\input{sec/appendix/main}


\end{document}

%% file: preamble.tex
\usepackage{amsthm}
\usepackage{amsmath} 
\usepackage{amssymb} 
\usepackage{pifont}
\newcommand{\cmark}{\ding{51}}%
\newcommand{\xmark}{\ding{55}}%

\usepackage{booktabs}
\usepackage{graphicx}
\usepackage{wrapfig}

\usepackage[dvipsnames]{xcolor}

\usepackage[pagebackref,breaklinks,colorlinks]{hyperref}
\usepackage[capitalize]{cleveref}
\crefname{section}{Sec.}{Secs.}
\Crefname{section}{Section}{Sections}
\Crefname{table}{Table}{Tables}
\crefname{table}{Tab.}{Tabs.}

\usepackage{xcolor}
\usepackage{tikz}
\usepackage{changepage}
\usepackage{tikz-3dplot}
\usepackage{tikzscale}
\usetikzlibrary{plotmarks}
\usepackage{pgfplots}
\usetikzlibrary{fadings}
\usetikzlibrary{tikzmark,calc} 
\usetikzlibrary{shapes.arrows}
\usetikzlibrary{decorations.pathreplacing, decorations.markings}
\usepgfplotslibrary{colorbrewer}
\usepgfmodule{nonlineartransformations} 
\makeatletter
\def\latticetilt{%
\pgf@xa=\pgf@x%
\pgf@ya=\pgf@y%
\pgfmathsetmacro{\myx}{\pgf@xa+\pgfkeysvalueof{/tikz/lattice/amplitude}*sin((\pgf@ya/\pgfkeysvalueof{/tikz/lattice/spacing})*360/\pgfkeysvalueof{/tikz/lattice/superlattice period})}%
\pgf@x=\myx pt%
\pgfmathsetmacro{\myy}{\pgf@ya+\pgfkeysvalueof{/tikz/lattice/amplitude}*sin((\pgf@xa/\pgfkeysvalueof{/tikz/lattice/spacing})*360/\pgfkeysvalueof{/tikz/lattice/superlattice period})}%
\pgf@y=\myy pt}

\usepackage[capbesideposition=right]{floatrow}

\xdefinecolor{navyblue}{RGB}{19 41 75}
\xdefinecolor{carolinablue}{RGB}{123 175 212}
\xdefinecolor{cb}{RGB}{123 175 212}
\xdefinecolor{flatgrey}{RGB}{162 170 173}
\xdefinecolor{matcha}{RGB}{183 224 176}

\definecolor{mygreen}{HTML}{DCEDC8}
\definecolor{lightgrey}{HTML}{E1E1E1}
\definecolor{darkblue}{HTML}{00275D}
\definecolor{mint}{HTML}{99EDC3}

\xdefinecolor{myyellow}{HTML}{FFE082}
\xdefinecolor{myblue}{RGB}{181, 222, 255}
\xdefinecolor{mypurple}{RGB}{202, 184, 255}
\xdefinecolor{myorange}{RGB}{255, 171, 145}
\xdefinecolor{tomato}{RGB}{255, 99, 71}

\definecolor{applegreen}{rgb}{0.4, 0.81, 0.0}
\definecolor{bittersweet}{rgb}{1.0, 0.2, 0.3}
\definecolor{mayablue}{rgb}{0.4, 0.5, 0.98}
\definecolor{citrine}{rgb}{1, 0.62, 0.14}
\definecolor{bubblegum}{rgb}{0.99, 0.76, 0.8}
\definecolor{olive}{rgb}{0.42, 0.56, 0.14}

\def\connect#1#2#3{%
    \path let
      \p1 = ($(#2)-(#1)$),
      \n1 = {veclen(\p1)-0.cm},
      \n2 = {atan2(\y1,\x1)} 
    in
      (#1) -- (#2) node[#3, midway, rotate = \n2, shading angle = \n2+90, minimum height=\n1, minimum    width=1pt, inner sep=1pt] {};
            }
            
\newcommand{\dashedLayer}[6]{
			\def\a{#1} 
			\def\b{0.02}
			\def\c{#2} 
			\def\t{#3} 
			\def\d{#4} 

			\draw[line width=0.15mm, dash pattern=on 2.25pt off 1.8pt](\c+\t,0,\d) -- (\c+\t,\a,\d) -- (\t,\a,\d);                                                      
			\draw[line width=0.15mm, dash pattern=on 2.25pt off 1.8pt](\t,0,\a+\d) -- (\c+\t,0,\a+\d) node[midway,below] {#6} -- (\c+\t,\a,\a+\d) -- (\t,\a,\a+\d) -- (\t,0,\a+\d); 
			\draw[line width=0.15mm, dash pattern=on 2.25pt off 1.8pt](\c+\t,0,\d) -- (\c+\t,0,\a+\d);
			\draw[line width=0.15mm, dash pattern=on 2.25pt off 1.8pt](\c+\t,\a,\d) -- (\c+\t,\a,\a+\d);
			\draw[line width=0.15mm, dash pattern=on 2.25pt off 1.8pt](\t,\a,\d) -- (\t,\a,\a+\d);
			
			\draw[line width=0.15mm] (\c+\t,0,\d) -- (\c+\t,\a,\d);
			\draw[line width=0.15mm] (\c+\t,0,\d) -- (\c+\t,0,\a+\d);
			\draw[line width=0.15mm] (\c+\t,\a,\d) -- (\c+\t,\a,\a+\d);
			\draw[line width=0.15mm] (\c+\t,0,\a+\d) -- (\c+\t,\a,\a+\d);
			
			\filldraw[#5] (\t+\b,\b,\a+\d) -- (\c+\t-\b,\b,\a+\d) -- (\c+\t-\b,\a-\b,\a+\d) -- (\t+\b,\a-\b,\a+\d) -- (\t+\b,\b,\a+\d); 
			\filldraw[#5] (\t+\b,\a,\a-\b+\d) -- (\c+\t-\b,\a,\a-\b+\d) -- (\c+\t-\b,\a,\b+\d) -- (\t+\b,\a,\b+\d);
			\ifthenelse {\equal{#5} {}}
			{} 
			{\filldraw[#5] (\c+\t,\b,\a-\b+\d) -- (\c+\t,\b,\b+\d) -- (\c+\t,\a-\b,\b+\d) -- (\c+\t,\a-\b,\a-\b+\d);} 
		}

\newcommand{\networkLayer}[6]{
			\def\a{#1} 
			\def\b{0.02}
			\def\c{#2} 
			\def\t{#3} 
			\def\d{#4} 

			\draw[line width=0.15mm](\c+\t,0,\d) -- (\c+\t,\a,\d) -- (\t,\a,\d);                                                      
			\draw[line width=0.15mm](\t,0,\a+\d) -- (\c+\t,0,\a+\d) node[midway,below] {#6} -- (\c+\t,\a,\a+\d) -- (\t,\a,\a+\d) -- (\t,0,\a+\d); 
			\draw[line width=0.15mm](\c+\t,0,\d) -- (\c+\t,0,\a+\d);
			\draw[line width=0.15mm](\c+\t,\a,\d) -- (\c+\t,\a,\a+\d);
			\draw[line width=0.15mm](\t,\a,\d) -- (\t,\a,\a+\d);

			\filldraw[#5] (\t+\b,\b,\a+\d) -- (\c+\t-\b,\b,\a+\d) -- (\c+\t-\b,\a-\b,\a+\d) -- (\t+\b,\a-\b,\a+\d) -- (\t+\b,\b,\a+\d); 
			\filldraw[#5] (\t+\b,\a,\a-\b+\d) -- (\c+\t-\b,\a,\a-\b+\d) -- (\c+\t-\b,\a,\b+\d) -- (\t+\b,\a,\b+\d);
			\ifthenelse {\equal{#5} {}}
			{} 
			{\filldraw[#5] (\c+\t,\b,\a-\b+\d) -- (\c+\t,\b,\b+\d) -- (\c+\t,\a-\b,\b+\d) -- (\c+\t,\a-\b,\a-\b+\d);} 
		}
		
\usepackage[linesnumbered,ruled,vlined]{algorithm2e}

\SetCommentSty{mycommfont}
\SetKwInput{Input}{Input}                
\SetKwInput{Output}{Output}         
\SetKwInput{Dataset}{Dataset}     
\SetKwInput{Parameters}{Parameters}  
\SetKwInput{Settings}{Settings}      
\SetKwInput{Initialization}{Initialization}      
\SetAlgorithmName{Alg.}{Alg.}{List of Algorithms} 
\makeatletter
\newcommand{\algorithmfootnote}[2][\footnotesize]{%
  \let\old@algocf@finish\@algocf@finish
  \def\@algocf@finish{\old@algocf@finish
    \leavevmode\rlap{\begin{minipage}{\linewidth}
    #1#2
    \end{minipage}}%
  }%
}
\makeatother

\usepackage{floatrow}
\usepackage{makecell}

\usepackage{booktabs}
\usepackage{multirow}
\newcolumntype{P}[1]{>{\centering\arraybackslash}p{#1}}

\usepackage{hyperref} 
  
\newtheoremstyle{bfnote}%
  {}{}
  {\itshape}{}
  {\bfseries}{.}
  { }{\thmname{#1}\thmnumber{ #2}\thmnote{ (#3)}}
\theoremstyle{bfnote}

\usepackage{url}

\pgfplotsset{compat=1.18}

%% file: sec/showcase.tex
 
\centering  
\captionsetup{type=figure}

\resizebox{\linewidth}{!}{
	\begin{tikzpicture}
		\tikzstyle{myarrows}=[line width=0.8mm,draw=blue!50,-triangle 45,postaction={draw, line width=0.05mm, shorten >=0.02mm, -}]
		\tikzstyle{mylines}=[line width=0.8mm]
  


	\pgfmathsetmacro{\cubex}{0.5*3}
	\pgfmathsetmacro{\cubey}{0.5*3}

	\pgfmathsetmacro{\shift}{-3.}
	\foreach \i/\cubez in {-5.9/0.15, -3.9/0.3, -1.5/0.2, 0.5/0.45}
	{
	\draw[black,fill=gray!30, line width = 0.02mm] (\shift+0.1*3,\i+0.15*3,0.1*3) -- ++(-\cubex,0,0) -- ++(0,-\cubey,0) -- ++(\cubex,0,0) -- cycle;
	\draw[black,fill=gray!35, line width = 0.02mm] (\shift+0.1*3,\i+0.15*3,0.1*3) -- ++(0,0,-\cubez) -- ++(0,-\cubey,0) -- ++(0,0,\cubez) -- cycle;
	\draw[black,fill=gray!35, line width = 0.02mm] (\shift+0.1*3,\i+0.15*3,0.1*3) -- ++(-\cubex,0,0) -- ++(0,0,-\cubez) -- ++(\cubex,0,0) -- cycle;
 
        \draw [<->, color = matcha!150, line width = 0.5mm](\shift+0.6,\i-0.15,0.1*3) -- (\shift+1.5,\i-0.15,0.1*3);  
    
	}
 
	\node at (\shift+0.1*3, -5.9+0.15*3, 0.75*3) {\includegraphics[width=0.1220\textwidth]{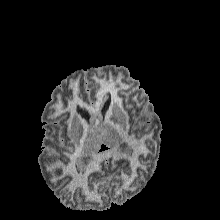}};
	\node at (\shift+0.1*3, -3.9+0.15*3, 0.75*3) {\includegraphics[width=0.1220\textwidth]{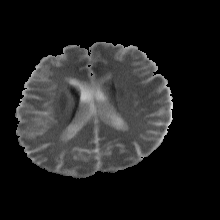}};
	\node at (\shift+0.1*3, -1.5+0.15*3, 0.75*3) {\includegraphics[width=0.1220\textwidth]{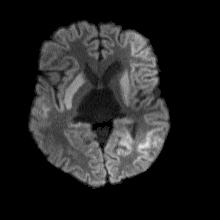}}; 
	\node at (\shift+0.1*3, 0.5+0.15*3, 0.75*3) {\includegraphics[width=0.1220\textwidth]{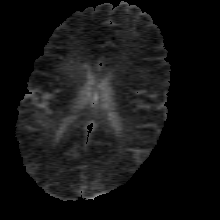}};
        \node at (0.+\shift+0.3-1.6, -0.5+0.15*3, 0.75*3) {ID-$i$};
        \node at (0.+\shift+0.3-1.6, -4.9+0.15*3, 0.75*3) {ID-$ii$};
 
    \draw [decorate,decoration={brace,amplitude=5pt,mirror,raise=6ex},line width=2.pt,color = gray] (-0.5+\shift, 0.9) -- (-0.5+\shift, -2.7);
    \draw [decorate,decoration={brace,amplitude=5pt,mirror,raise=6ex},line width=2.pt,color = gray] (-0.5+\shift, 0.9-4.4) -- (-0.5+\shift, -2.7-4.4);
    
 

	\pgfmathsetmacro{\cubez}{0.12}  
	\pgfmathsetmacro{\shift}{-3.2}
 
	\foreach \i/\j in {3.2/-5.9, 3.2/-3.9, 3.2/-1.5, 3.2/0.5,  5.2/-5.9, 5.2/-3.9, 5.2/-1.5, 5.2/0.5,  7.2/-5.9, 7.2/-3.9, 7.2/-1.5, 7.2/0.5,  9.2/-5.9, 9.2/-3.9, 9.2/-1.5, 9.2/0.5,  12.6/-5.9, 12.6/-3.9, 12.6/-1.5, 12.6/0.5}
	{
	\draw[black,fill=gray!30, line width = 0.02mm] (\i+\shift+0.1*3,\j+0.15*3,0.1*3) -- ++(-\cubex,0,0) -- ++(0,-\cubey,0) -- ++(\cubex,0,0) -- cycle;
	\draw[black,fill=gray!35, line width = 0.02mm] (\i+\shift+0.1*3,\j+0.15*3,0.1*3) -- ++(0,0,-\cubez) -- ++(0,-\cubey,0) -- ++(0,0,\cubez) -- cycle;
	\draw[black,fill=gray!35, line width = 0.02mm] (\i+\shift+0.1*3,\j+0.15*3,0.1*3) -- ++(-\cubex,0,0) -- ++(0,0,-\cubez) -- ++(\cubex,0,0) -- cycle;
        
        \draw[black,fill=black] (10.9-0.2+\shift+0.3, \j+0.15*3, 0.75*3) circle (0.8pt);
        \draw[black,fill=black] (10.9+\shift+0.3, \j+0.15*3, 0.75*3) circle (0.8pt);
        \draw[black,fill=black] (10.9+0.2+\shift+0.3, \j+0.15*3, 0.75*3) circle (0.8pt);

	}

        \pgfmathsetmacro{\dx}{-1.5}
        \foreach \dy in {-2.2, -0.2, 2.2, 4.2}
        {
	\draw[dashed, color = matcha!150, line width=0.4mm] (0+\dx, -5.+\dy) -- (11.3+\dx, -5.+\dy) -- (11.3+\dx, -3.2+\dy) -- (0+\dx, -3.2+\dy) -- (0+\dx, -5.+\dy); 
        }
        
	\node at (3.2+\shift+0.1*3, 0.5+0.15*3, 0.75*3) {\includegraphics[width=0.1220\textwidth]{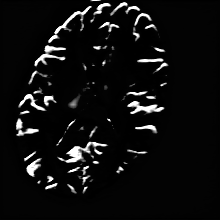}};
	\node at (3.2+\shift+0.1*3, -1.5+0.15*3, 0.75*3) {\includegraphics[width=0.1220\textwidth]{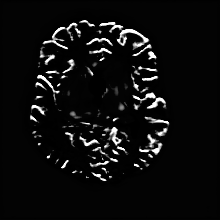}};
	\node at (3.2+\shift+0.1*3, -3.9+0.15*3, 0.75*3) {\includegraphics[width=0.1220\textwidth]{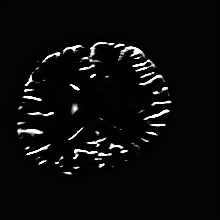}};
	\node at (3.2+\shift+0.1*3, -5.9+0.15*3, 0.75*3) {\includegraphics[width=0.1220\textwidth]{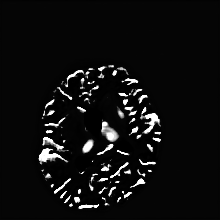}};
 
	\node at (5.2+\shift+0.1*3, 0.5+0.15*3, 0.75*3) {\includegraphics[width=0.1220\textwidth]{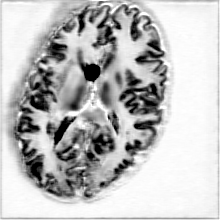}};
	\node at (5.2+\shift+0.1*3, -1.5+0.15*3, 0.75*3) {\includegraphics[width=0.1220\textwidth]{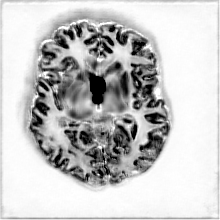}};
	\node at (5.2+\shift+0.1*3, -3.9+0.15*3, 0.75*3) {\includegraphics[width=0.1220\textwidth]{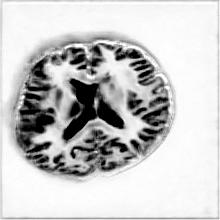}};
	\node at (5.2+\shift+0.1*3, -5.9+0.15*3, 0.75*3) {\includegraphics[width=0.1220\textwidth]{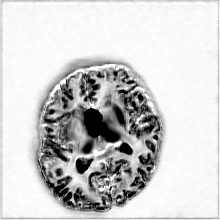}};
        
	\node at (7.2+\shift+0.1*3, 0.5+0.15*3, 0.75*3) {\includegraphics[width=0.1220\textwidth]{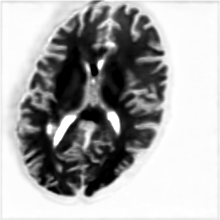}};
	\node at (7.2+\shift+0.1*3, -1.5+0.15*3, 0.75*3) {\includegraphics[width=0.1220\textwidth]{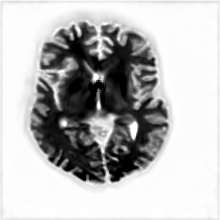}};
	\node at (7.2+\shift+0.1*3, -3.9+0.15*3, 0.75*3) {\includegraphics[width=0.1220\textwidth]{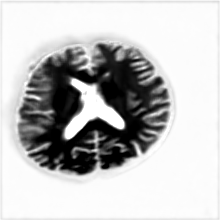}};
	\node at (7.2+\shift+0.1*3, -5.9+0.15*3, 0.75*3) {\includegraphics[width=0.1220\textwidth]{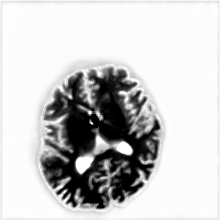}};
 
	\node at (9.2+\shift+0.1*3, 0.5+0.15*3, 0.75*3) {\includegraphics[width=0.1220\textwidth]{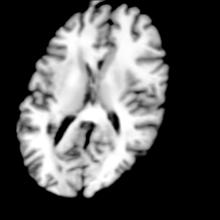}};
	\node at (9.2+\shift+0.1*3, -1.5+0.15*3, 0.75*3) {\includegraphics[width=0.1220\textwidth]{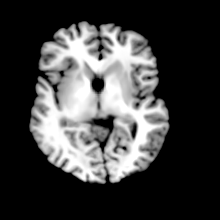}};
	\node at (9.2+\shift+0.1*3, -3.9+0.15*3, 0.75*3) {\includegraphics[width=0.1220\textwidth]{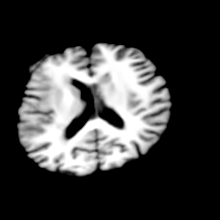}};
	\node at (9.2+\shift+0.1*3, -5.9+0.15*3, 0.75*3) {\includegraphics[width=0.1220\textwidth]{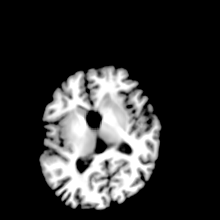}};
 
	\node at (12.6+\shift+0.1*3, 0.5+0.15*3, 0.75*3) {\includegraphics[width=0.1220\textwidth]{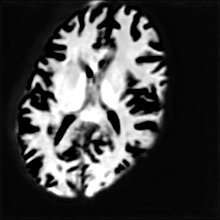}};
	\node at (12.6+\shift+0.1*3, -1.5+0.15*3, 0.75*3) {\includegraphics[width=0.1220\textwidth]{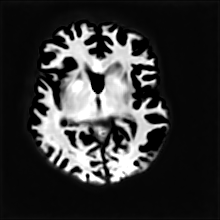}};
	\node at (12.6+\shift+0.1*3, -3.9+0.15*3, 0.75*3) {\includegraphics[width=0.1220\textwidth]{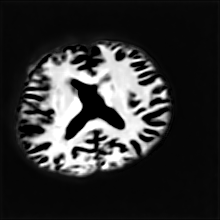}};
	\node at (12.6+\shift+0.1*3, -5.9+0.15*3, 0.75*3) {\includegraphics[width=0.1220\textwidth]{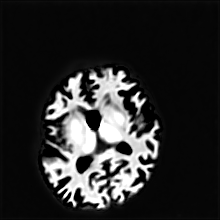}};


    \pgfmathsetmacro{\dx}{2.5}
    \pgfmathsetmacro{\dy}{0}
    \draw [myarrows, color = matcha!150](-4.1+\dx, -7.45+\dy) -- (7.5+\dx, -7.45+\dy);  
    \node[anchor=north] at (1.7+\dx, -7.6+\dy) {\textcolor{matcha!150}{\textbf{Feature Channel}}}; 
    \node[anchor=north] at (1.7+\dx, 1.6+\dy) {\textcolor{matcha!150}{\textbf{\texttt{Brain-ID} Subject-specific Features}}};

        
	\pgfmathsetmacro{\cubez}{0.12}  
	\pgfmathsetmacro{\shift}{-3.2}
 
	\foreach \i/\j in {15.6/-5.9, 15.6/-3.9, 15.6/-1.5, 15.6/0.5, 17.8/-5.9, 17.8/-3.9, 17.8/-1.5, 17.8/0.5}
	{
	\draw[black,fill=gray!30, line width = 0.02mm] (\i+\shift+0.1*3,\j+0.15*3,0.1*3) -- ++(-\cubex,0,0) -- ++(0,-\cubey,0) -- ++(\cubex,0,0) -- cycle;
	\draw[black,fill=gray!35, line width = 0.02mm] (\i+\shift+0.1*3,\j+0.15*3,0.1*3) -- ++(0,0,-\cubez) -- ++(0,-\cubey,0) -- ++(0,0,\cubez) -- cycle;
	\draw[black,fill=gray!35, line width = 0.02mm] (\i+\shift+0.1*3,\j+0.15*3,0.1*3) -- ++(-\cubex,0,0) -- ++(0,0,-\cubez) -- ++(\cubex,0,0) -- cycle;
        
        \draw[black,fill=black] (10.9-0.2+\shift+0.3, \j+0.15*3, 0.75*3) circle (0.8pt);
        \draw[black,fill=black] (10.9+\shift+0.3, \j+0.15*3, 0.75*3) circle (0.8pt);
        \draw[black,fill=black] (10.9+0.2+\shift+0.3, \j+0.15*3, 0.75*3) circle (0.8pt);

        \draw [->, color = matcha!150, line width = 0.5mm](13.7+\shift+0.3, \j+0.15*3, 0.75*3) -- (13.7+0.9+\shift+0.25, \j+0.15*3, 0.75*3);  
        
	}

	\pgfmathsetmacro{\i}{15.6}   
	\node at (\i+\shift+0.1*3, 0.5+0.15*3, 0.75*3) {\includegraphics[width=0.1220\textwidth]{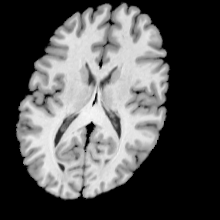}};
	\node at (\i+\shift+0.1*3, -1.5+0.15*3, 0.75*3) {\includegraphics[width=0.1220\textwidth]{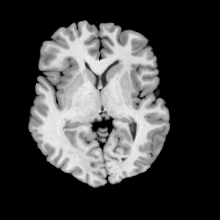}};
	\node at (\i+\shift+0.1*3, -3.9+0.15*3, 0.75*3) {\includegraphics[width=0.1220\textwidth]{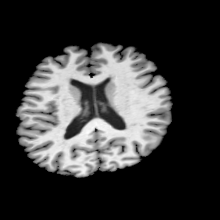}};
	\node at (\i+\shift+0.1*3, -5.9+0.15*3, 0.75*3) {\includegraphics[width=0.1220\textwidth]{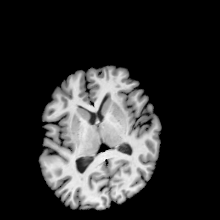}};
 
	\pgfmathsetmacro{\i}{17.8}   
	\node at (\i+\shift+0.1*3, 0.5+0.15*3, 0.75*3) {\includegraphics[width=0.1220\textwidth]{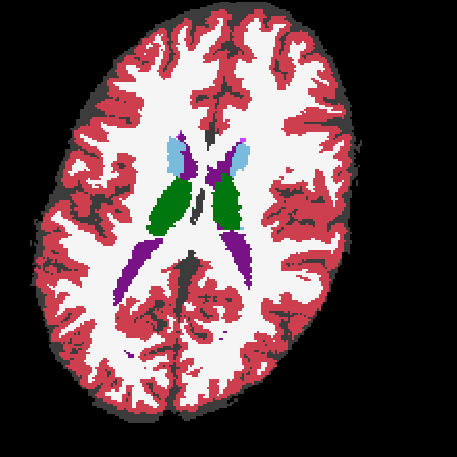}};
	\node at (\i+\shift+0.1*3, -1.5+0.15*3, 0.75*3) {\includegraphics[width=0.1220\textwidth]{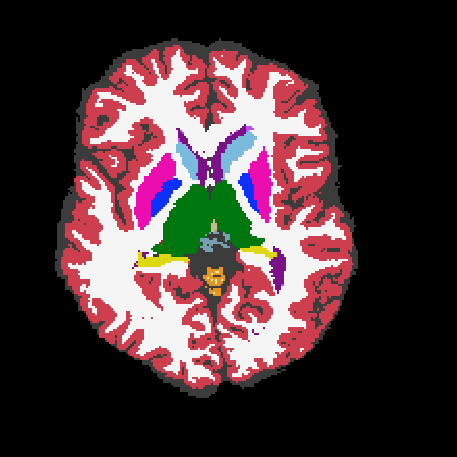}};
	\node at (\i+\shift+0.1*3, -3.9+0.15*3, 0.75*3) {\includegraphics[width=0.1220\textwidth]{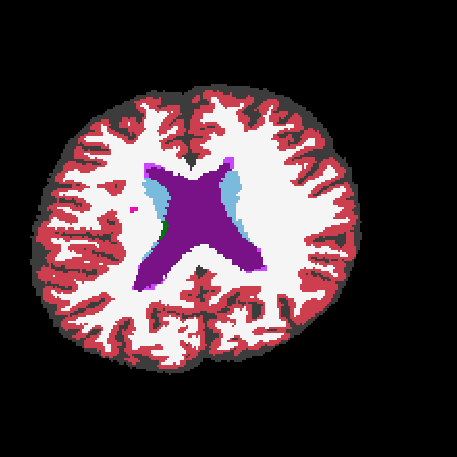}};
	\node at (\i+\shift+0.1*3, -5.9+0.15*3, 0.75*3) {\includegraphics[width=0.1220\textwidth]{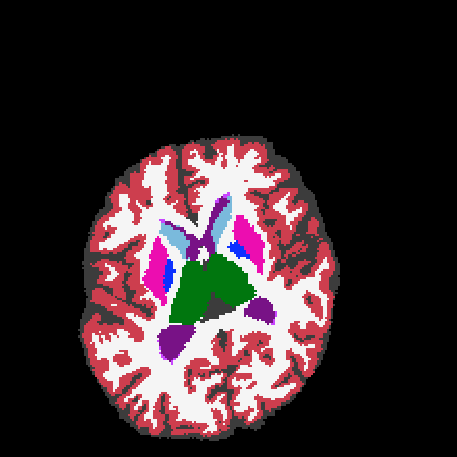}};

        \pgfmathsetmacro{\dx}{6.9}
        \node[anchor=north] at (4.9+\dx, 1.58+\dy) {\color{orange!80}\textbf{Recon}};
        \node[anchor=north] at (7.1+\dx, 1.58+\dy) {\color{orange!80}\textbf{Seg}};
        
        \draw [decorate,decoration={brace,amplitude=5pt,mirror,raise=6ex},line width=2.pt,color = orange!80] (3.9+\dx, -6.4) -- (8.2+\dx, -6.4);
        \node[anchor=north] at (6.1+\dx, -7.6+\dy) {\textcolor{orange!80}{\textbf{Contrast-independent}}};

        \foreach \dy in {-0.2, 4.2}
        {
	\draw[dashed, color = orange!80, line width=0.4mm] (4+\dx, -7+\dy) -- (5.9+\dx, -7+\dy) -- (5.9+\dx, -3.2+\dy) -- (4+\dx, -3.2+\dy) -- (4+\dx, -7+\dy); 
        }
        \pgfmathsetmacro{\dx}{9.1}
        \foreach \dy in {-0.2, 4.2}
        {
	\draw[dashed, color = orange!80, line width=0.4mm] (4+\dx, -7+\dy) -- (5.9+\dx, -7+\dy) -- (5.9+\dx, -3.2+\dy) -- (4+\dx, -3.2+\dy) -- (4+\dx, -7+\dy); 
        } 
 
        
	\pgfmathsetmacro{\cubez}{0.12}  
	\pgfmathsetmacro{\shift}{1.3}
 
	\foreach \i/\j in {15.6/-5.9, 15.6/-3.9, 15.6/-1.5, 15.6/0.5, 17.8/-5.9, 17.8/-3.9, 17.8/-1.5, 17.8/0.5}
	{
	\draw[black,fill=gray!30, line width = 0.02mm] (\i+\shift+0.1*3,\j+0.15*3,0.1*3) -- ++(-\cubex,0,0) -- ++(0,-\cubey,0) -- ++(\cubex,0,0) -- cycle;
	\draw[black,fill=gray!35, line width = 0.02mm] (\i+\shift+0.1*3,\j+0.15*3,0.1*3) -- ++(0,0,-\cubez) -- ++(0,-\cubey,0) -- ++(0,0,\cubez) -- cycle;
	\draw[black,fill=gray!35, line width = 0.02mm] (\i+\shift+0.1*3,\j+0.15*3,0.1*3) -- ++(-\cubex,0,0) -- ++(0,0,-\cubez) -- ++(\cubex,0,0) -- cycle;
        
        \draw[black,fill=black] (10.9-0.2+\shift+0.3, \j+0.15*3, 0.75*3) circle (0.8pt);
        \draw[black,fill=black] (10.9+\shift+0.3, \j+0.15*3, 0.75*3) circle (0.8pt);
        \draw[black,fill=black] (10.9+0.2+\shift+0.3, \j+0.15*3, 0.75*3) circle (0.8pt);

	}

	\pgfmathsetmacro{\i}{15.6}    
	\node at (\i+\shift+0.1*3, 0.5+0.15*3, 0.75*3) {\includegraphics[width=0.1220\textwidth]{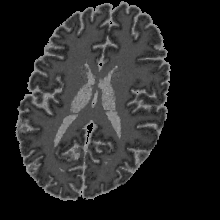}};
	\node at (\i+\shift+0.1*3, -1.5+0.15*3, 0.75*3) {\includegraphics[width=0.1220\textwidth]{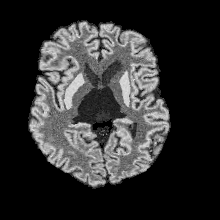}};
	\node at (\i+\shift+0.1*3, -3.9+0.15*3, 0.75*3) {\includegraphics[width=0.1220\textwidth]{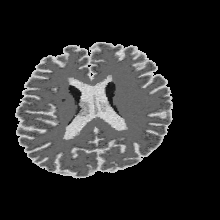}};
	\node at (\i+\shift+0.1*3, -5.9+0.15*3, 0.75*3) {\includegraphics[width=0.1220\textwidth]{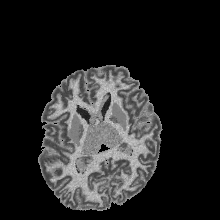}};

	\pgfmathsetmacro{\i}{17.8}   
	\node at (\i+\shift+0.1*3, 0.5+0.15*3, 0.75*3) {\includegraphics[width=0.1220\textwidth]{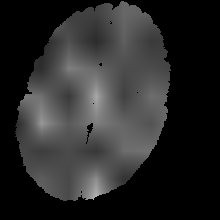}};
	\node at (\i+\shift+0.1*3, -1.5+0.15*3, 0.75*3) {\includegraphics[width=0.1220\textwidth]{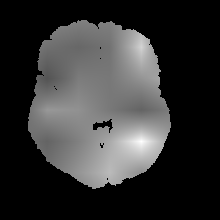}};
	\node at (\i+\shift+0.1*3, -3.9+0.15*3, 0.75*3) {\includegraphics[width=0.1220\textwidth]{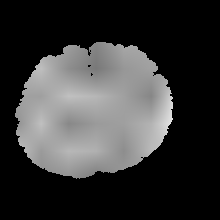}};
	\node at (\i+\shift+0.1*3, -5.9+0.15*3, 0.75*3) {\includegraphics[width=0.1220\textwidth]{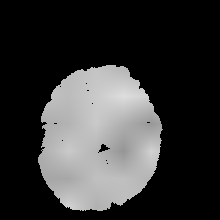}};

        \pgfmathsetmacro{\dx}{11.4}
        \node[anchor=north] at (4.9+\dx, 1.58+\dy) {\textbf{\color{cb!100}SR}};
        \node[anchor=north] at (7.1+\dx, 1.58+\dy) {\textbf{\color{cb!100}BF}};
        
        \draw [decorate,decoration={brace,amplitude=5pt,mirror,raise=6ex},line width=2.pt,color = cb!100] (3.9+\dx, -6.4) -- (8.2+\dx, -6.4);
        \node[anchor=north] at (6.1+\dx, -7.6+\dy) {\textcolor{cb!100}{\textbf{Contrast-dependent}}}; 
        
        \pgfmathsetmacro{\dx}{11.4}
        \foreach \dy in {-2.2, -0.2, 2.2, 4.2}
        {
	\draw[dashed, color = cb!100, line width=0.4mm] (4+\dx, -5.+\dy) -- (8.1+\dx, -5.+\dy) -- (8.1+\dx, -3.2+\dy) -- (4+\dx, -3.2+\dy) -- (4+\dx, -5.+\dy); 
        }
        

	\end{tikzpicture}
	}  
 
	\vspace{-0.2cm}
\caption{\texttt{Brain-ID} features serve as the distinctive identity of each subject (ID) regardless of appearance (contrast, deformation, corruptions, etc) and quickly adapt to downstream tasks -- either contrast-independent (anatomy {\textbf{\color{orange!80}Recon}}struction, {\color{orange!80}\textbf{Seg}}mentation), or contrast-dependent ({\color{cb}\textbf{S}}uper-{\color{cb}\textbf{R}}esolution, {\color{cb}\textbf{B}}ias {\color{cb}\textbf{F}}ield estimation), all through one layer.}  
	 \label{showcase}

%% file: sec/abstract.tex
\vspace*{-0.45cm}
\begin{abstract} 
Recent learning-based approaches have made astonishing advances in calibrated medical imaging like computerized tomography (CT), yet they struggle to generalize in uncalibrated modalities -- notably magnetic resonance (MR) imaging, where performance is highly sensitive to the differences in MR contrast, resolution, and orientation. This prevents broad applicability to diverse real-world clinical protocols. We introduce \texttt{Brain-ID}, an anatomical representation learning model for brain imaging. With the proposed ``mild-to-severe'' intra-subject generation, \texttt{Brain-ID} is robust to the subject-specific brain anatomy regardless of the appearance of acquired images (e.g., contrast, deformation, resolution, artifacts). Trained entirely on synthetic data, \texttt{Brain-ID} readily adapts to various downstream tasks through only one layer. We present new metrics to validate the intra- and inter-subject robustness of \texttt{Brain-ID} features, and evaluate their performance on four downstream applications, covering contrast-independent (anatomy reconstruction/contrast synthesis, brain segmentation), and contrast-dependent (super-resolution, bias field estimation) tasks~(\cref{showcase}). Extensive experiments on six public datasets demonstrate that \texttt{Brain-ID} achieves state-of-the-art performance in all tasks on different MRI modalities and CT, and more importantly, preserves its performance on low-resolution and small datasets. Code is available at \href{https://github.com/peirong26/Brain-ID}{https://github.com/peirong26/Brain-ID}. 
\end{abstract}

%% file: sec/intro.tex
\section{Introduction}
\label{sec: intro}  
\vspace{-0.2cm}
Magnetic resonance imaging (MRI) enables in vivo noninvasive imaging of the human brain with exquisite and tunable soft-tissue contrast~\cite{BrantZawadzki1992MPRA}. Recent machine learning based methods have achieved great improvements in faster and more accurate image analysis of brain MRI~\cite{Iglesias2023SynthSRAP}, such as image segmentation~\cite{Ronneberger2015UNetCN, Milletar2016VNetFC, Kamnitsas2016EfficientM3, Ding_2021_ICCV}, registration~\cite{Balakrishnan2018VoxelMorphAL,Yang2017QuicksilverFP,de2019deep,Shen2021Accurate}, super-resolution~\cite{Tian2020ImprovingIV,Tanno2020UncertaintyMI}, and connectivity studies~\cite{Mller2011UnderconnectedBH}. However, most existing MRI analysis methods are specific to certain MR contrast(s) and often require near-isotropic acquisitions. Therefore, models face sharp performance drops when voxel size and anisotropy increase, or are being used for a different contrast than what seen during training~\cite{wang2018deep}. This reduces model generalizability and results in duplicate data collection and training efforts given new datasets. By resorting to synthetic data, recent contrast-agnostic models~\cite{Iglesias2020JointSA,Liu2021YETI,Liu2022SONATA,Iglesias2023SynthSRAP,Billot2021SynthSegSO,Hoffmann2020SynthMorphLC,Laso2023WMH} achieve impressive results and largely extend the applicability of models to heterogeneous clinical acquisition protocols. However, these models are only applicable to the tasks they were trained for.

Meanwhile, task-agnostic foundation models in computer vision and natural language processing have witnessed remarkable success, along with the fast developments of large-scale datasets~\cite{Brown2020LanguageMA, Chowdhery2022PaLMSL, Kirillov2023SegmentA}. Designed in a task-agnostic manner, foundation models have shown impressive performance in obtaining more general feature representations and can be quickly adapted (e.g., fine-tuned) to a wide range of downstream tasks~\cite{Bommasani2021OnTO, Awais2023FoundationalMD}. However, due to different acquisition protocols, processing pipelines, and privacy requirements across institutions, large-scale datasets in medical imaging require significantly more effort than natural imaging/language. As a result, medical foundation models are not as well developed. The MONAI~\cite{cardoso2022monai} project includes a model zoo containing pre-trained models for various tasks, yet they are all highly task-oriented and sensitive to specific image contrasts. Zhou et al.~\cite{Zhou2023AFM} constructed a medical foundation model, yet it is designed for the detection of eye and systemic health conditions from retinal scans, and only works on the modalities of color fundus photography and optical coherence tomography. Lately, generalist biomedical AI systems~\cite{Moor2023FoundationMF,Singhal2022LargeLM, Tu2023TowardsGB} have shown great potential in biomedical tasks under vision-language context, e.g., (visual) question answering, image classification, radiology report generation and summarization. However, they have not explored the challenging vision tasks such as reconstruction, segmentation, super-resolution, and registration.

Here, we introduce \texttt{Brain-ID}, a subject-specific, contrast- and task-agnostic anatomical representation approach which is trained \textit{entirely} on synthetic data, and can swiftly adapt to diverse downstream tasks through only \textit{one} layer.
\vspace{-0.2cm}
\begin{itemize}
    \item[1)] We introduce our on-the-fly, intra-subject data generator capable of sythesizing \textit{any} contrast, employing a \textit{mild-to-severe} increasing-corruption generation, all from the \textit{same} brain anatomy~(\cref{fig: augment}, \cref{parag: sample}). Unlike real-world datasets that are constrained by the images acquired per subject, \texttt{Brain-ID} learns in a much more expansive and diverse space.
    \vspace{0.1cm}
    
    \item[2)] We design a feature representation learning framework guided by the \textit{unique anatomy} of each subject. Demonstrated by our proposed evaluation approach for both inter- and intra-subject robustness, \texttt{Brain-ID}'s high-resolution features are consistently robust to the superficial perturbations of the input image's appearance~(Sec.~\ref{exp: feat}) such as contrast, resolution, orientation, artifacts. We further present straightforward one-layer adaptions that enable \texttt{Brain-ID} features to seamlessly adapt to downstream tasks~(Sec.~\ref{sec: downstream}).
    \vspace{0.1cm}
    
    \item[3)] We extensively evaluate \texttt{Brain-ID} on both contrast-independent (anatomy reconstruction, brain segmentation) and contrast-dependent (super-resolution, bias field estimation) tasks, across six public datasets (around 8,000 images in total) including modalities of MR (T1w, T2w, FLAIR) and CT. \texttt{Brain-ID} achieves state-of-the-art performance in all tasks (\cref{tab: task}). More importantly, it maintains its high performance even on low-resolution data~(\cref{fig: feat_comp}) and datasets with limited size~(\cref{fig: comp_curve}).
    
\end{itemize}

%% file: sec/related_work.tex
\section{Related work}
\label{sec: related_work}
 \vspace{-0.05cm}
\subsubsection{Feature Representation in Medical Imaging}
\label{related: feat_repre} 
As discussed in Sec.~\ref{sec: intro}, general feature representation learning in the medical imaging can be more challenging than in the natural image domain, due to limited data availability. Xu et al.~\cite{Xu2014DeepLO} introduced a multiple instance learning model for feature representation in medical imaging, but it is designed specifically for classification, and only applies to histopathology images. You et al.~\cite{You2021MomentumCV} presented CVRL, a semi-supervised approach for voxel-wise representations, which is designed for image segmentation, but requires CT inputs to extract anatomical information. SAM~\cite{Yan2020SAMSL} is a self-supervised framework to encode anatomical information from CT images for feature embeddings, which has shown to be effective in downstream tasks such as registration (SAMConvex~\cite{Li2023SAMConvexFD}). However, same as CVRL, SAM only works on CT. BrainPrint~\cite{Wachinger2014BrainPrintI} is a compact and discriminative representation of brain morphology, which is specifically designed for cortical surface analyses. To our best knowledge, CIFL~\cite{chua2023contrast} is the only existing work on learning contrast-agnostic and task-independent brain feature representations. CIFL relies on contrastive learning alone, and is insufficient to outperform task/contrast-specific supervised models in downstream applications as shown by our experiments~(\cref{tab: task}).

\vspace{-0.2cm}
\subsubsection{Contrast-invariant Learning for MR Images}
\label{related: contrast_agnostic} 
MRI scans acquired across sites vary substantially in contrasts, resolutions, orientations, etc. When given a new dataset, heterogeneity leads to duplicate training efforts for approaches that are sensitive to specific combinations of MR contrast, resolution, or orientation. Classical brain segmentation models used Bayesian inference for contrast-robustness~\cite{Leemput2003AUF, Fischl2002WholeBS}, which requires a long processing time and struggles with low or anisotropic resolutions~\cite{Puonti2016FastAS,Iglesias2023SynthSRAP}. Recently, SynthSeg~\cite{Billot2021SynthSegSO,Laso2023WMH} was proposed for contrast-agnostic segmentation and achieves impressive results with a synthetic generator that simulates widely diverse contrasts. Meanwhile, there have been works with similar ideas of using synthetic data to achieve contrast-invariance in tasks like image registration~\cite{Hoffmann2020SynthMorphLC}, super-resolution~\cite{Iglesias2020JointSA}, and  skullstripping~\cite{Hoopes2022SynthStripSF}. However, all the above-mentioned methods are trained in a task-specific manner, whose features therefore cannot be readily applied to other domains.

%% file: sec/method/main.tex
\input{sec/method/fig/fw_gen}


\vspace{-0.2cm}
\section{\texttt{Brain-ID}: Learning Anatomy-specific Brain Features}
\label{sec: method}
\vspace{-0.2cm}

As discussed in \cref{sec: intro}, the main challenges to obtain a general and robust feature representation for MR imaging lie in \textit{(i)}~the practical restrictions of building large-scale datasets with diverse contrasts; and \textit{(ii)}~the nature of most medical imaging models that are task-oriented and specific to data type (contrast, resolution, orientation, etc). We aim to learn a brain feature representation that is:\vspace{0.05cm}\\
\textit{(i)} \textit{Robust}: features should be robust to each subject's distinct anatomy, unaffected by variations in poses/deformations, contrasts, resolutions, or artifacts.\vspace{0.05cm}\\
\textit{(ii)} \textit{Expressive}: features should also exhibit high expressiveness, containing rich information that facilitates easy and effective adaptation to diverse downstream tasks, eliminating the necessity for extensive training data.

We first introduce \texttt{Brain-ID}'s data generator (\cref{sec: generator}) and training framework (\cref{sec: framework}) to achieve the above two aims. Then, we present our one-layer solutions for adapting \texttt{Brain-ID} features to downstream tasks that could be either dependent or independent of the contrast of input images~(\cref{sec: downstream}).


\input{sec/method/generator}

\input{sec/method/fig/fw_train}

\input{sec/method/trainer}

\input{sec/method/downstream}


%% file: sec/method/fig/fw_gen.tex
 
\begin{figure*}[t]
\centering
\resizebox{\textwidth}{!}{
	
	\begin{tikzpicture}[lattice/.cd,spacing/.initial=4,superlattice
  period/.initial=12,amplitude/.initial=2]
	\pgfmathsetmacro{\cubex}{0.29*3}
	\pgfmathsetmacro{\cubey}{0.29*3}
	\pgfmathsetmacro{\cubez}{0.028*3}
	

	\pgfmathsetmacro{\shift}{0.3}
	\foreach \i in {-1.15}
	{
	\draw[black,fill=gray!30, line width = 0.02mm] (\i+\shift+0.1*3,0.15*3,0.1*3) -- ++(-\cubex,0,0) -- ++(0,-\cubey,0) -- ++(\cubex,0,0) -- cycle;
	\draw[black,fill=gray!35, line width = 0.02mm] (\i+\shift+0.1*3,0.15*3,0.1*3) -- ++(0,0,-\cubez) -- ++(0,-\cubey,0) -- ++(0,0,\cubez) -- cycle;
	\draw[black,fill=gray!35, line width = 0.02mm] (\i+\shift+0.1*3,0.15*3,0.1*3) -- ++(-\cubex,0,0) -- ++(0,0,-\cubez) -- ++(\cubex,0,0) -- cycle;
	}
 
	\node at (-1.15+\shift+0.2975*3, 0.3475*3, 0.99*3) {\includegraphics[width=0.071\textwidth]{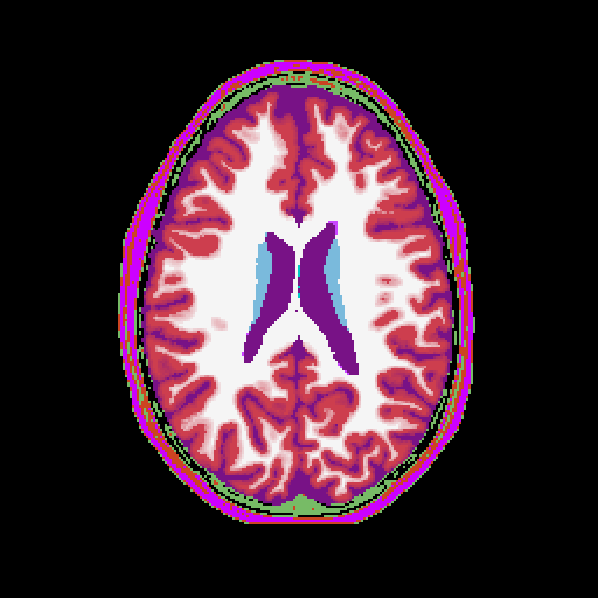}};

	\draw[-latex] (-1.17+0.6+\shift, 0.05, 0.25*3) -- (0.1+2., 0.05, 0.25*3);
	\node at (-1.15+0.93+\shift, 0.49, 3.1) {\tiny Label ($L$)}; 
	\node at (-2.01+1.8+\shift, 0.2, 0.25*3) {\tiny \cref{eq: deform}};
	\node at (-1.0+2.1+\shift, 0.2, 0.25*3) {\tiny \text{Deformation}};
 
	\node at (-2+1.8+\shift, -0.1, 0.25*3) {\tiny \cref{eq: contrast}}; 
	\node at (-1.13+2.04+\shift, -0.1, 0.25*3) {\tiny \text{Contrast}};

	\pgfmathsetmacro{\dx}{-5.5}
	\pgfmathsetmacro{\dy}{2.7}
        
        \begin{scope}[xshift=-5]
            \pgftransformnonlinear{\latticetilt}
            \draw[orange!80,step=0.08,thin] (0.15,0.15) grid (0.88,0.88);
            \draw[->, black] (0.15,0.15) -- +(0.85,0);
            \draw[->, black] (0.15,0.15) -- +(0,0.85);
            
        \end{scope} 
    
	\draw[->, line width = 0.2mm, color=orange!80] (5.5+\dx+\shift, -2.6+\dy) -- (5.5+\dx+\shift, -2.9+\dy);
 
	\node at (0.35+\shift+0.2975*3, 0.26, 3.2) {\includegraphics[width=0.115\textwidth]{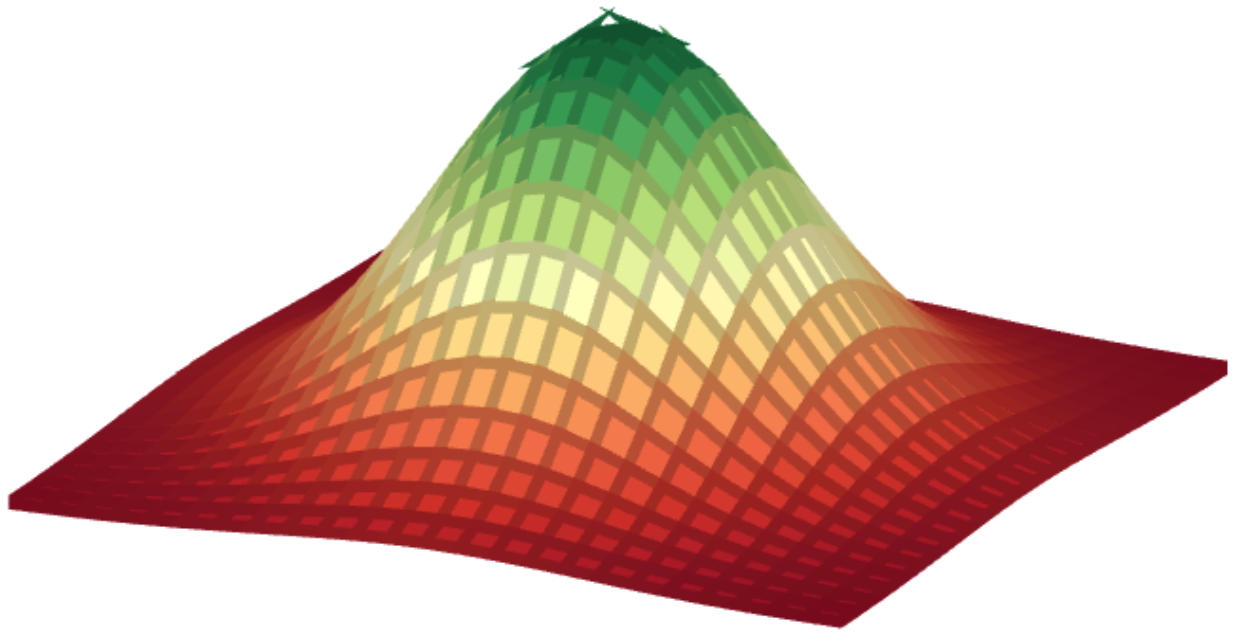}};

	\draw[->, line width = 0.2mm, color=matcha!200] (5.5+\dx+\shift, -3.28+\dy) -- (5.5+\dx+\shift, -2.98+\dy);
	
	
	\pgfmathsetmacro{\shift}{1.3}
	\foreach \i in {1.3}
	{
	\draw[black,fill=gray!30, line width = 0.02mm] (\i+\shift+0.4*3,0.5*3,0.5*3) -- ++(-\cubex,0,0) -- ++(0,-\cubey,0) -- ++(\cubex,0,0) -- cycle;
	\draw[black,fill=gray!35, line width = 0.02mm] (\i+\shift+0.4*3,0.5*3,0.5*3) -- ++(0,0,-\cubez) -- ++(0,-\cubey,0) -- ++(0,0,\cubez) -- cycle;
	\draw[black,fill=gray!35, line width = 0.02mm] (\i+\shift+0.4*3,0.5*3,0.5*3) -- ++(-\cubex,0,0) -- ++(0,0,-\cubez) -- ++(\cubex,0,0) -- cycle;
	\node at (\i+\shift+0.409*3, 0.509*3, 0.9*3) {\includegraphics[width=0.071\textwidth]{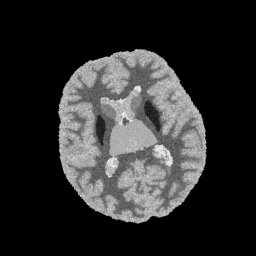}};
	\node at (\i+\shift-0.35, 0.26, 0.13) {\tiny $I_{i}$};

	\draw[black,fill=gray!30, line width = 0.02mm] (\i+\shift-0.1*3,-0.2*3,-0.2*3) -- ++(-\cubex,0,0) -- ++(0,-\cubey,0) -- ++(\cubex,0,0) -- cycle;
	\draw[black,fill=gray!35, line width = 0.02mm] (\i+\shift-0.1*3, -0.2*3,-0.2*3) -- ++(0,0,-\cubez) -- ++(0,-\cubey,0) -- ++(0,0,\cubez) -- cycle;
	\draw[black,fill=gray!35, line width = 0.02mm] (\i+\shift-0.1*3,-0.2*3,-0.2*3) -- ++(-\cubex,0,0) -- ++(0,0,-\cubez) -- ++(\cubex,0,0) -- cycle;	
	\node at (\i+\shift-0.75, -0.35*3, -0.212*3) {\includegraphics[width=0.071\textwidth]{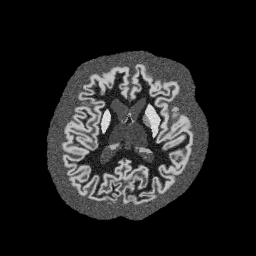}}; 
 
	\draw[black,fill=black] (\i+\shift+0.45, 0., 0.2*3) circle (0.4pt);
	\draw[black,fill=black] (\i+\shift+0.45, 0., 0.25*3) circle (0.4pt);
	\draw[black,fill=black] (\i+\shift+0.45, 0., 0.3*3) circle (0.4pt);
	\node at (\i+\shift+0.1, 0.08, 3.1) {\tiny $I_{1}$}; 
	
	\draw[-latex] (\i+\shift+0.7, 0.05, 0.25*3) -- (\i+\shift+1.93, 0.05, 0.25*3); 
	\node at (\i+\shift+1.30, 0.2, 0.25*3) {\tiny \text{Corruption}};
	\node at (\i+\shift+1.20, -0.1, 0.25*3) {\tiny \text{Resampling}};
	\node at (\i+\shift+1.19, -0.3, 0.25*3) {\tiny \text{({Resolution}}};
	\node at (\i+\shift+1.21, -0.45, 0.25*3) {\tiny \text{simulation)}};
	
	}
	
	
	\pgfmathsetmacro{\cubeza}{0.05*3}
	\pgfmathsetmacro{\cubezb}{0.04*3}
	\pgfmathsetmacro{\shift}{4.2}
	\foreach \i in {1}
	{
	\draw[black,fill=gray!30, line width = 0.02mm] (\i+\shift+0.4*3,0.5*3,0.5*3) -- ++(-\cubex,0,0) -- ++(0,-\cubey,0) -- ++(\cubex,0,0) -- cycle;
	\draw[black,fill=gray!35, line width = 0.02mm] (\i+\shift+0.4*3,0.5*3,0.5*3) -- ++(0,0,-\cubeza) -- ++(0,-\cubey,0) -- ++(0,0,\cubeza) -- cycle;
	\draw[black,fill=gray!35, line width = 0.02mm] (\i+\shift+0.4*3,0.5*3,0.5*3) -- ++(-\cubex,0,0) -- ++(0,0,-\cubeza) -- ++(\cubex,0,0) -- cycle;
	\node at (\i+\shift+0.4086*3, 0.509*3, 0.9*3) {\includegraphics[width=0.071\textwidth]{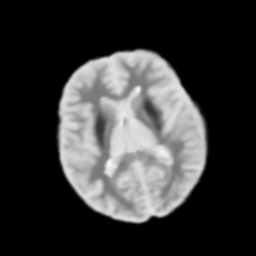}};
	\node at (\i+\shift+0.2, -0.05, 0.13) {\tiny $S_{i-\i}$};

	\draw[black,fill=gray!30, line width = 0.02mm] (\i+\shift-0.1*3,-0.2*3,-0.2*3) -- ++(-\cubex,0,0) -- ++(0,-\cubey,0) -- ++(\cubex,0,0) -- cycle;
	\draw[black,fill=gray!35, line width = 0.02mm] (\i+\shift-0.1*3, -0.2*3,-0.2*3) -- ++(0,0,-\cubezb) -- ++(0,-\cubey,0) -- ++(0,0,\cubezb) -- cycle;
	\draw[black,fill=gray!35, line width = 0.02mm] (\i+\shift-0.1*3,-0.2*3,-0.2*3) -- ++(-\cubex,0,0) -- ++(0,0,-\cubezb) -- ++(\cubex,0,0) -- cycle;
	
	\node at (\i+\shift-0.241*3, -0.341*3, -0.19*3) {\includegraphics[width=0.071\textwidth]{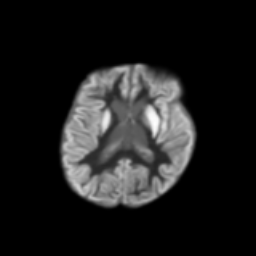}};
	\node at (\i+\shift+0.65, -0.23, 3.1) {\tiny $S_{1-\i}$};
	}

	\pgfmathsetmacro{\cubeza}{0.03*3}
	\pgfmathsetmacro{\cubezb}{0.1*3}
	\pgfmathsetmacro{\shift}{4.2}
	\foreach \i in {2}
	{
	\draw[black,fill=gray!30, line width = 0.02mm] (\i+\shift+0.4*3,0.5*3,0.5*3) -- ++(-\cubex,0,0) -- ++(0,-\cubey,0) -- ++(\cubex,0,0) -- cycle;
	\draw[black,fill=gray!35, line width = 0.02mm] (\i+\shift+0.4*3,0.5*3,0.5*3) -- ++(0,0,-\cubeza) -- ++(0,-\cubey,0) -- ++(0,0,\cubeza) -- cycle;
	\draw[black,fill=gray!35, line width = 0.02mm] (\i+\shift+0.4*3,0.5*3,0.5*3) -- ++(-\cubex,0,0) -- ++(0,0,-\cubeza) -- ++(\cubex,0,0) -- cycle;
	\node at (\i+\shift+0.4086*3, 0.509*3, 0.9*3) {\includegraphics[width=0.071\textwidth]{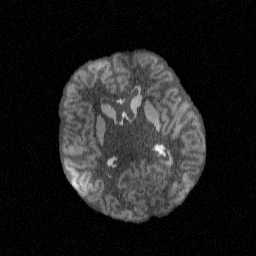}};
	\node at (\i+\shift+0.2, -0.05, 0.13) {\tiny $S_{i-\i}$};

	\draw[black,fill=gray!30, line width = 0.02mm] (\i+\shift-0.1*3,-0.2*3,-0.2*3) -- ++(-\cubex,0,0) -- ++(0,-\cubey,0) -- ++(\cubex,0,0) -- cycle;
	\draw[black,fill=gray!35, line width = 0.02mm] (\i+\shift-0.1*3, -0.2*3,-0.2*3) -- ++(0,0,-\cubezb) -- ++(0,-\cubey,0) -- ++(0,0,\cubezb) -- cycle;
	\draw[black,fill=gray!35, line width = 0.02mm] (\i+\shift-0.1*3,-0.2*3,-0.2*3) -- ++(-\cubex,0,0) -- ++(0,0,-\cubezb) -- ++(\cubex,0,0) -- cycle;
	
	\node at (\i+\shift-0.241*3, -0.341*3, -0.19*3) {\includegraphics[width=0.071\textwidth]{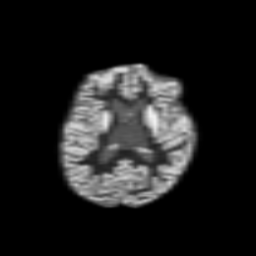}};
	\node at (\i+\shift+0.65, -0.23, 3.1) {\tiny $S_{1-\i}$};
	}

	\pgfmathsetmacro{\cubeza}{0.05*3}
	\pgfmathsetmacro{\cubezb}{0.04*3}
	\pgfmathsetmacro{\shift}{4.2}
	\foreach \i in {3}
	{
	\draw[black,fill=gray!30, line width = 0.02mm] (\i+\shift+0.4*3,0.5*3,0.5*3) -- ++(-\cubex,0,0) -- ++(0,-\cubey,0) -- ++(\cubex,0,0) -- cycle;
	\draw[black,fill=gray!35, line width = 0.02mm] (\i+\shift+0.4*3,0.5*3,0.5*3) -- ++(0,0,-\cubeza) -- ++(0,-\cubey,0) -- ++(0,0,\cubeza) -- cycle;
	\draw[black,fill=gray!35, line width = 0.02mm] (\i+\shift+0.4*3,0.5*3,0.5*3) -- ++(-\cubex,0,0) -- ++(0,0,-\cubeza) -- ++(\cubex,0,0) -- cycle;
	\node at (\i+\shift+0.4086*3, 0.509*3, 0.9*3) {\includegraphics[width=0.071\textwidth]{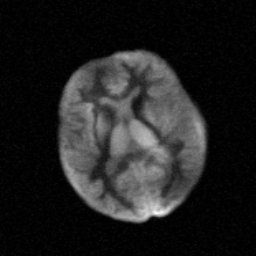}};
	\node at (\i+\shift+0.2, -0.05, 0.13) {\tiny $S_{i-\i}$};
	\draw[black,fill=black] (\i+\shift+1.1, 0.4, 0.25*3) circle (0.4pt);
	\draw[black,fill=black] (\i+\shift+1.2, 0.4, 0.25*3) circle (0.4pt);
	\draw[black,fill=black] (\i+\shift+1.3, 0.4, 0.25*3) circle (0.4pt);

	\draw[black,fill=gray!30, line width = 0.02mm] (\i+\shift-0.1*3,-0.2*3,-0.2*3) -- ++(-\cubex,0,0) -- ++(0,-\cubey,0) -- ++(\cubex,0,0) -- cycle;
	\draw[black,fill=gray!35, line width = 0.02mm] (\i+\shift-0.1*3, -0.2*3,-0.2*3) -- ++(0,0,-\cubezb) -- ++(0,-\cubey,0) -- ++(0,0,\cubezb) -- cycle;
	\draw[black,fill=gray!35, line width = 0.02mm] (\i+\shift-0.1*3,-0.2*3,-0.2*3) -- ++(-\cubex,0,0) -- ++(0,0,-\cubezb) -- ++(\cubex,0,0) -- cycle;
	
	\node at (\i+\shift-0.241*3, -0.341*3, -0.19*3) {\includegraphics[width=0.071\textwidth]{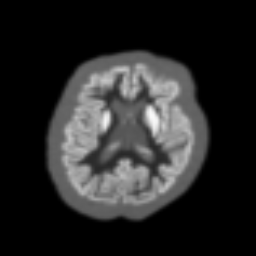}};
	\node at (\i+\shift+0.65, -0.23, 3.1) {\tiny $S_{1-\i}$};
	}

	\pgfmathsetmacro{\cubeza}{0.07*3}
	\pgfmathsetmacro{\cubezb}{0.02*3}
	\pgfmathsetmacro{\shift}{4.2}
	\foreach \i in {4}
	{
	\draw[black,fill=gray!30, line width = 0.02mm] (\i+\shift+0.4*3,0.5*3,0.5*3) -- ++(-\cubex,0,0) -- ++(0,-\cubey,0) -- ++(\cubex,0,0) -- cycle;
	\draw[black,fill=gray!35, line width = 0.02mm] (\i+\shift+0.4*3,0.5*3,0.5*3) -- ++(0,0,-\cubeza) -- ++(0,-\cubey,0) -- ++(0,0,\cubeza) -- cycle;
	\draw[black,fill=gray!35, line width = 0.02mm] (\i+\shift+0.4*3,0.5*3,0.5*3) -- ++(-\cubex,0,0) -- ++(0,0,-\cubeza) -- ++(\cubex,0,0) -- cycle;
	\node at (\i+\shift+0.4086*3, 0.509*3, 0.9*3) {\includegraphics[width=0.071\textwidth]{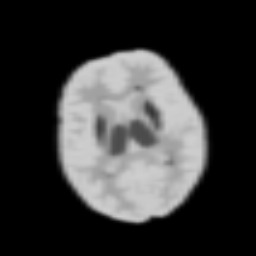}};
	\node at (\i+\shift+0.2, -0.05, 0.13) {\tiny $S_{i-\i}$};
	\draw[black,fill=black] (\i+\shift+1.1, 0.4, 0.25*3) circle (0.4pt);
	\draw[black,fill=black] (\i+\shift+1.2, 0.4, 0.25*3) circle (0.4pt);
	\draw[black,fill=black] (\i+\shift+1.3, 0.4, 0.25*3) circle (0.4pt);

	\draw[black,fill=gray!30, line width = 0.02mm] (\i+\shift-0.1*3,-0.2*3,-0.2*3) -- ++(-\cubex,0,0) -- ++(0,-\cubey,0) -- ++(\cubex,0,0) -- cycle;
	\draw[black,fill=gray!35, line width = 0.02mm] (\i+\shift-0.1*3, -0.2*3,-0.2*3) -- ++(0,0,-\cubezb) -- ++(0,-\cubey,0) -- ++(0,0,\cubezb) -- cycle;
	\draw[black,fill=gray!35, line width = 0.02mm] (\i+\shift-0.1*3,-0.2*3,-0.2*3) -- ++(-\cubex,0,0) -- ++(0,0,-\cubezb) -- ++(\cubex,0,0) -- cycle;
	
	\node at (\i+\shift-0.241*3, -0.341*3, -0.19*3) {\includegraphics[width=0.071\textwidth]{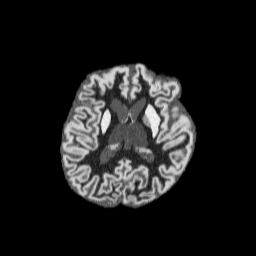}};
	\node at (\i+\shift+0.65, -0.23, 3.1) {\tiny $S_{1-\i}$};
	\draw[black,fill=black] (\i+\shift+0.36, -0.93, 0.25*3) circle (0.4pt);
	\draw[black,fill=black] (\i+\shift+0.46, -0.93, 0.25*3) circle (0.4pt);
	\draw[black,fill=black] (\i+\shift+0.56, -0.93, 0.25*3) circle (0.4pt);
	}
	\end{tikzpicture}
	
} 
\caption{\texttt{Brain-ID}'s data generator on the fly. Given the brain segmentation labels of a subject, we randomly generate a deformation field, and synthesize intra-subjecct samples featuring various contrast intensities and corruption levels (\cref{sec: generator}).}
\label{fig: augment}
\vspace{-0.3cm}
\end{figure*}

%% file: sec/method/generator.tex
\vspace{-0.2cm}
\subsection{Enrich the Intra-subject Learning Space}
\label{sec: generator}
\vspace{-0.2cm}

A good representation relies on large-scale data, however, different acquisition protocols, processing pipelines, and privacy requirements across institutions make large-scale data less accessible and require significantly more effort, e.g., via federated learning~\cite{Sheller2020Federated,Adnan2022Federated,Darzidehkalani2022Federated}. Moreover, for each subject, there are usually only limited acquired images/modalities available. This lack of data consistency is a significant barrier to obtaining an subject-specific, contrast-invariant feature representation. \texttt{Brain-ID} avoids these barriers through the use of synthetic data. 

In order to generate images with complex brain structures, we start from high-resolution brain segmentation images that provide labels of brain structures and extracerebral regions ($L$ in \cref{fig: augment}). The generatation consists of three steps, \textit{(i)}~deformation generation, \textit{(ii)}~contrast synthesis, and \textit{(iii)}~data corruption (including lower resolution resampling). For simplicity, $\Theta$ denotes the parameter group of the generation process described below.


\vspace{-0.5cm}
\subsubsection{Deformation Generation}
\label{sec: deform}
We first generate a random deformation field ($\phi\vert_{\theta_\phi}$) consisting of an affine transformation and a non-linear displacement field:
\vspace{-0.23cm}
\begin{equation}
    \phi\vert_{\theta_\phi} = \mathcal{T}\vert_{\theta_\phi} \circ \mathcal{A}\vert_{\theta_\phi}\,,
    \label{eq: deform}
\vspace{-0.25cm}
\end{equation}
where $\mathcal{A}\vert_{\theta_\phi}$ denotes an affine transformation matrix which includes linear (rotation, scaling, shearing) transformation and translation, $\mathcal{T}\vert_{\theta_\phi}$ refers to a non-linear displacement field computed as the integration of a stationary velocity field (SVF) that is smooth and invertible everywhere and thus preserves the topology of the brain anatomy~\cite{Iglesias2020JointSA}. $\theta_\phi \in \Theta$ controls the transformation ranges.


\vspace{-0.4cm}
\subsubsection{MR Contrast Synthesis}
\label{sec: contrast}
Then, we synthesize images ($I(x),\, x \in \Omega$) by randomly ``painting'' intensities on the segmentation maps according to their brain structure labels ($l\in L$). Specifically, the regional intensities are generated by separately sampling a Gaussian distribution on each labeled region:
\vspace{-0.25cm}
\begin{equation} 
    \begin{cases}
    \vspace{-0.1cm}
        I(x) \sim \mathcal{N}(\mu_l,\, \sigma_l ) \,, \quad l \in L\,,\\ 
        \mu_l \sim \mathcal{N}(0,\, 1 \, \vert\, \theta_{\mu}, \, \theta_{l}) \,,~
        \sigma_l \sim \mathcal{N}(0,\, 1 \, \vert \, \theta_{\sigma}, \, \theta_{l}) \,,
    \end{cases}
    \label{eq: contrast}
\vspace{-0.25cm}
\end{equation}
where $\mu_l$ and $\sigma_l$ refer to the mean and standard deviation of each segmentation label $l$, and are independently sampled from Gaussian distributions at each voxel, with $\theta_{l},\, \theta_{\mu},\, \theta_{\sigma} \in \Theta$ controlling the shifts and scales of their values.


\vspace{-0.4cm}
\subsubsection{Resolution Simulation and Data Corruption}
\label{sec: corrupt}
Given a deformed, contrast-synthesized image ($I$), we adopt the standard data corruption pipeline~\cite{Iglesias2023SynthSRAP}, which further augments images with different levels of resolution and scanning artifacts that are commonly found in real-world clinical protocols.

As illustrated in \cref{fig: augment}, we are able to generate an infinite number of variations from a single subject with its unique brain anatomy labels. By generating images with randomized contrast/resolution/orientation on the fly for each subject, we enormously enrich the learning space for a robust representation, and focus the learning on intrinsic subject-specific features rather than superficial aspects of images that depend on acquisition parameters and conditions. (In \cref{exp: ablation}, we conduct experiments to provide insights on how different choices of data corruption levels would affect the robustness of \texttt{Brain-ID} features.)

%% file: sec/method/fig/fw_train.tex
\begin{figure*}[t] 
\centering  
\resizebox{\linewidth}{!}{
	\begin{tikzpicture}
		\tikzstyle{myarrows}=[line width=0.8mm,draw=blue!50,-triangle 45,postaction={draw, line width=0.05mm, shorten >=0.02mm, -}]
		\tikzstyle{mylines}=[line width=0.8mm]
  


	\pgfmathsetmacro{\dx}{-1.5}
	\pgfmathsetmacro{\dy}{0}
        \connect{(-0+\dx, -7.13+\dy}{(-0+\dx, 2+\dy}{single arrow, top color=cb!10, bottom color=cb!150};
    \node at (0.85+\dx, -7.2+0.15*3, 0.75*3) {(\cref{sec: generator})};
    
	\node[rotate=90,anchor=north] at (-0.95+\dx, -2.6+\dy) {\textcolor{cb!120}{\large\textbf{Corruption Level}}};
 
	\node[] at (-0.75+\dx, -6.2+\dy) {\textcolor{cb!80}{\large\textbf{Mild}}};
	\node[] at (-0.75+\dx, 1.+\dy) {\textcolor{cb!150}{\large\textbf{Severe}}};


	\pgfmathsetmacro{\cubex}{0.5*3}
	\pgfmathsetmacro{\cubey}{0.5*3}

	\pgfmathsetmacro{\shift}{1.}
	\foreach \i/\cubez in {-5.9/0.12, -3.9/0.2, -0.5/0.3, 1.5/0.45}
	{
	\draw[black,fill=gray!35, line width = 0.02mm] (\shift+0.1*3,\i+0.15*3,0.1*3) -- ++(-\cubex,0,0) -- ++(0,-\cubey,0) -- ++(\cubex,0,0) -- cycle;
	\draw[black,fill=gray!35, line width = 0.02mm] (\shift+0.1*3,\i+0.15*3,0.1*3) -- ++(0,0,-\cubez) -- ++(0,-\cubey,0) -- ++(0,0,\cubez) -- cycle;
	\draw[black,fill=gray!35, line width = 0.02mm] (\shift+0.1*3,\i+0.15*3,0.1*3) -- ++(-\cubex,0,0) -- ++(0,0,-\cubez) -- ++(\cubex,0,0) -- cycle;
	}
 
	\node at (\shift+0.1*3, -5.9+0.15*3, 0.75*3) {\includegraphics[width=0.1220\textwidth]{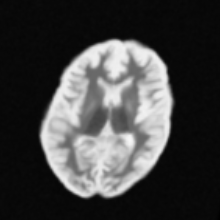}};
	\node at (\shift+0.1*3, -3.9+0.15*3, 0.75*3) {\includegraphics[width=0.1220\textwidth]{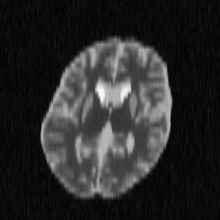}};
	\node at (\shift+0.1*3, -0.5+0.15*3, 0.75*3) {\includegraphics[width=0.1220\textwidth]{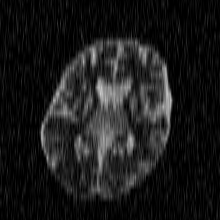}}; 
	\node at (\shift+0.1*3, 1.5+0.15*3, 0.75*3) {\includegraphics[width=0.1220\textwidth]{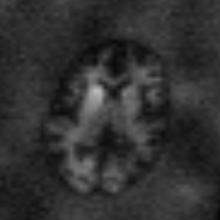}};
        \node at (0.+\shift+0.3-1.3, 1.5+0.15*3, 0.75*3) {\large$\mathbf{S}_N$};
        \node at (0.+\shift+0.3-1.3, -0.5+0.15*3, 0.75*3) {\large$\mathbf{S}_{N-1}$};
        \node at (0.+\shift+0.3-1.3, -3.9+0.15*3, 0.75*3) {\large$\mathbf{S}_2$};
        \node at (0.+\shift+0.3-1.3, -5.9+0.15*3, 0.75*3) {\large$\mathbf{S}_1$};
 
    \draw[black,fill=black] (\shift+0.3, -2.+0.15*3, 0.75*3) circle (0.8pt);
    \draw[black,fill=black] (\shift+0.3, -2.2+0.15*3, 0.75*3) circle (0.8pt);
    \draw[black,fill=black] (\shift+0.3, -2.4+0.15*3, 0.75*3) circle (0.8pt);

    \node at (\shift+0.3, -7.2+0.15*3, 0.75*3) {\large ID-$idx$};

    \node at (\shift+3.58, -6.2+0.15*3, 0.75*3) {(Full setup: Appendix~\ref{app: implement})};
    \node at (\shift+2.2, -5.7+0.15*3, 0.75*3) {\textcolor{cb!80}{\textbf{Mild}}};
    \node at (\shift+3.67, -5.7+0.15*3, 0.75*3) {: $\sigma_{\text{noise}} = 1,\, \cdots$};
    \node at (\shift+2.35, -5.2+0.15*3, 0.75*3) {\textcolor{cb!150}{\textbf{Severe}}};
    \node at (\shift+4.07, -5.2+0.15*3, 0.75*3) {: $\sigma_{\text{noise}} = 10,\, \cdots$};

        \pgfmathsetmacro{\dx}{-2.17}
        \pgfmathsetmacro{\dy}{-2.}
	\draw[dashed, color = cb, line width=0.4mm] (4+\dx, -5.+\dy) -- (7.75+\dx, -5.+\dy) -- (7.75+\dx, -3.2+\dy) -- (4+\dx, -3.2+\dy) -- (4+\dx, -5.+\dy);

    \draw [decorate,decoration={brace,amplitude=5pt,raise=6ex},line width=2.pt,color = cb] (0.7, 2) -- (0.7, -7.1);

    
    \pgfmathsetmacro{\sx}{2.1}
    \pgfmathsetmacro{\sy}{5.4}
    \pgfmathsetmacro{\dx}{\sx+0}
    \pgfmathsetmacro{\dy}{\sy+0.2} 
    \pgfmathsetmacro{\ddy}{-3.5} 
    \pgfmathsetmacro{\dxt}{\dx+0.5} 
    
    \node at (1.5+\sx, -6.+\dy+\ddy){\large $\mathcal{F}$};
    \pgfmathsetmacro{\dy}{\dy+\ddy} 
    \networkLayer{2}{0.1}{5.1-\dy+\dxt}{13-2.6*\dy}{color=myblue!80}{}
    \networkLayer{1.6}{0.2}{5.2-\dy+\dxt}{13-2.6*\dy}{color=myblue!60}{}
    \networkLayer{1.2}{0.4}{5.4-\dy+\dxt}{13-2.6*\dy}{color=myblue!40}{}
    \networkLayer{0.8}{0.8}{5.8-\dy+\dxt}{13-2.6*\dy}{color=myblue!20}{}
    
    \pgfmathsetmacro{\dxs}{\dx+0}
    \pgfmathsetmacro{\dys}{\dy+0} 
    \networkLayer{0.8}{0.8}{7.2-\dys+\dxs}{13-2.6*\dys}{color=matcha!20}{}
    \networkLayer{1.2}{0.4}{7.6-\dys+\dxs}{13-2.6*\dys}{color=matcha!40}{}
    \networkLayer{1.6}{0.2}{8.-\dys+\dxs}{13-2.6*\dys}{color=matcha!60}{}
    \networkLayer{2}{0.1}{8.3-\dys+\dxs}{13-2.6*\dys}{color=matcha!80}{}

 
    \draw [decorate,decoration={brace,amplitude=5pt,mirror,raise=6ex},line width=2.pt,color = matcha!150] (6.7, 2) -- (6.7, -7.1);

	\pgfmathsetmacro{\cubez}{0.12}  
	\pgfmathsetmacro{\shift}{1.2}
 
	\foreach \i/\j in {7.2/-5.9, 7.2/-3.9, 7.2/-0.5, 7.2/1.5,  9.2/-5.9, 9.2/-3.9, 9.2/-0.5, 9.2/1.5,  12.6/-5.9, 12.6/-3.9, 12.6/-0.5, 12.6/1.5}
	{
	\draw[black,fill=gray!35, line width = 0.02mm] (\i+\shift+0.1*3,\j+0.15*3,0.1*3) -- ++(-\cubex,0,0) -- ++(0,-\cubey,0) -- ++(\cubex,0,0) -- cycle;
	\draw[black,fill=gray!35, line width = 0.02mm] (\i+\shift+0.1*3,\j+0.15*3,0.1*3) -- ++(0,0,-\cubez) -- ++(0,-\cubey,0) -- ++(0,0,\cubez) -- cycle;
	\draw[black,fill=gray!35, line width = 0.02mm] (\i+\shift+0.1*3,\j+0.15*3,0.1*3) -- ++(-\cubex,0,0) -- ++(0,0,-\cubez) -- ++(\cubex,0,0) -- cycle;
 
        \draw[black,fill=black] (\i+\shift+0.3, -2.+0.15*3, 0.75*3) circle (0.8pt);
        \draw[black,fill=black] (\i+\shift+0.3, -2.2+0.15*3, 0.75*3) circle (0.8pt);
        \draw[black,fill=black] (\i+\shift+0.3, -2.4+0.15*3, 0.75*3) circle (0.8pt);
        
        \draw[black,fill=black] (10.9-0.2+\shift+0.3, \j+0.15*3, 0.75*3) circle (0.8pt);
        \draw[black,fill=black] (10.9+\shift+0.3, \j+0.15*3, 0.75*3) circle (0.8pt);
        \draw[black,fill=black] (10.9+0.2+\shift+0.3, \j+0.15*3, 0.75*3) circle (0.8pt);
        
        \draw[black,fill=black] (10.9-0.2+\shift+0.3, -2.+0.15*3, 0.75*3) circle (0.8pt);
        \draw[black,fill=black] (10.9+\shift+0.3, -2.2+0.15*3, 0.75*3) circle (0.8pt);
        \draw[black,fill=black] (10.9+0.2+\shift+0.3, -2.4+0.15*3, 0.75*3) circle (0.8pt);
	}

        \pgfmathsetmacro{\dx}{2.9}
        \foreach \dy in {-2.2, -0.2, 3.2, 5.2}
        {
	\draw[dashed, color = matcha!150, line width=0.4mm] (4+\dx, -5.+\dy) -- (11.3+\dx, -5.+\dy) -- (11.3+\dx, -3.2+\dy) -- (4+\dx, -3.2+\dy) -- (4+\dx, -5.+\dy); 
        }
        
	\node at (7.2+\shift+0.1*3, 1.5+0.15*3, 0.75*3) {\includegraphics[width=0.1220\textwidth]{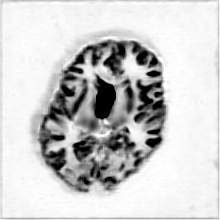}};
	\node at (7.2+\shift+0.1*3, -0.5+0.15*3, 0.75*3) {\includegraphics[width=0.1220\textwidth]{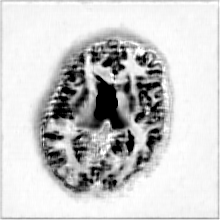}};
	\node at (7.2+\shift+0.1*3, -3.9+0.15*3, 0.75*3) {\includegraphics[width=0.1220\textwidth]{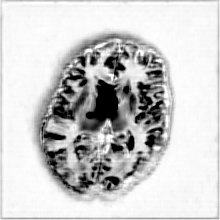}};
	\node at (7.2+\shift+0.1*3, -5.9+0.15*3, 0.75*3) {\includegraphics[width=0.1220\textwidth]{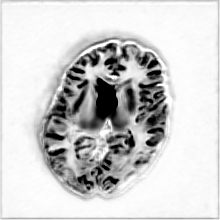}};
 
	\node at (9.2+\shift+0.1*3, 1.5+0.15*3, 0.75*3) {\includegraphics[width=0.1220\textwidth]{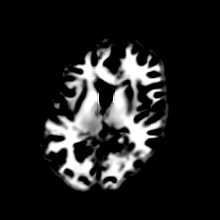}};
	\node at (9.2+\shift+0.1*3, -0.5+0.15*3, 0.75*3) {\includegraphics[width=0.1220\textwidth]{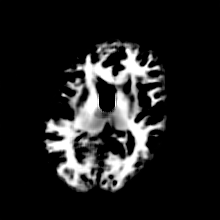}};
	\node at (9.2+\shift+0.1*3, -3.9+0.15*3, 0.75*3) {\includegraphics[width=0.1220\textwidth]{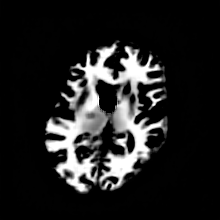}};
	\node at (9.2+\shift+0.1*3, -5.9+0.15*3, 0.75*3) {\includegraphics[width=0.1220\textwidth]{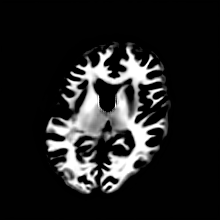}};
 
	\node at (12.6+\shift+0.1*3, 1.5+0.15*3, 0.75*3) {\includegraphics[width=0.1220\textwidth]{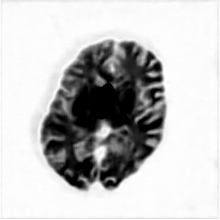}};
	\node at (12.6+\shift+0.1*3, -0.5+0.15*3, 0.75*3) {\includegraphics[width=0.1220\textwidth]{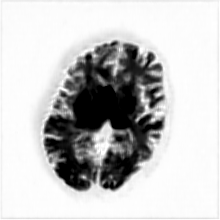}};
	\node at (12.6+\shift+0.1*3, -3.9+0.15*3, 0.75*3) {\includegraphics[width=0.1220\textwidth]{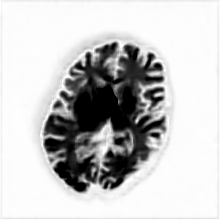}};
	\node at (12.6+\shift+0.1*3, -5.9+0.15*3, 0.75*3) {\includegraphics[width=0.1220\textwidth]{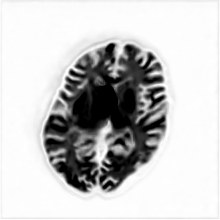}};

        \node at (7.27+\shift+0.3-1.5, 1.5+0.15*3, 0.75*3) {\large$\mathbf{F}_{N}$};
        \node at (7.25+\shift+0.3-1.5, -0.5+0.15*3, 0.75*3) {\large$\mathbf{F}_{N-1}$};
        \node at (7.27+\shift+0.3-1.5, -3.9+0.15*3, 0.75*3) {\large$\mathbf{F}_2$};
        \node at (7.27+\shift+0.3-1.5, -5.9+0.15*3, 0.75*3) {\large$\mathbf{F}_1$};

	\pgfmathsetmacro{\shift}{1.6}
 
        \draw [myarrows, color = orange!80, dotted](12.4+\shift+0.3+2.1, -5.9+0.15*3, 0.75*3) -- (11.4+\shift+2.3, -5.9+0.15*3, 0.75*3);  
        
        \draw [myarrows, color = orange!80, dotted](12.4+\shift+0.3+2.1, -3.9+0.15*3, 0.75*3) -- (11.4+\shift+2.3, -3.9+0.15*3, 0.75*3);  
        
        \draw [myarrows, color = orange!80, dotted](12.4+\shift+0.3+2.1, -0.5+0.15*3, 0.75*3) -- (11.4+\shift+2.3, -0.5+0.15*3, 0.75*3);  
        
        \draw [myarrows, color = orange!80, dotted](12.4+\shift+0.3+2.1, 1.5+0.15*3, 0.75*3) -- (11.4+\shift+2.3, 1.5+0.15*3, 0.75*3);  
        
        \draw [mylines, color = orange!80, dotted](12.4+\shift+0.3+2.1, -5.9+0.15*3, 0.75*3) -- (12.4+\shift+0.3+2.1, 1.5+0.15*3, 0.75*3); 
        
        \draw [mylines, color = orange!80, dotted](12.4+\shift+0.3+2.1, -2.2+0.15*3, 0.75*3) -- (13.7+\shift+0.3+2.3, -2.2+0.15*3, 0.75*3); 
        
        \node at (12.+\shift+3.6, -2.2+0.4+0.15*3, 0.75*3) {\large\cref{eq: Brain-loss}};

    \pgfmathsetmacro{\i}{16.8}
    \pgfmathsetmacro{\j}{-2.2}
    \draw[black,fill=gray!35, line width = 0.02mm] (\i+\shift+0.1*3,\j+0.15*3,0.1*3) -- ++(-\cubex,0,0) -- ++(0,-\cubey,0) -- ++(\cubex,0,0) -- cycle;
    \draw[black,fill=gray!35, line width = 0.02mm] (\i+\shift+0.1*3,\j+0.15*3,0.1*3) -- ++(0,0,-\cubez) -- ++(0,-\cubey,0) -- ++(0,0,\cubez) -- cycle;
    \draw[black,fill=gray!35, line width = 0.02mm] (\i+\shift+0.1*3,\j+0.15*3,0.1*3) -- ++(-\cubex,0,0) -- ++(0,0,-\cubez) -- ++(\cubex,0,0) -- cycle;

    \node at (16.8+\shift+0.1*3, -2.2+0.15*3, 0.75*3) {\includegraphics[width=0.1220\textwidth]{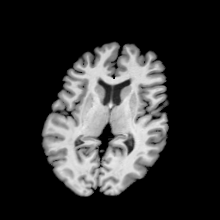}};
    \node at (16.8+\shift+0.1*3, -3.5+0.15*3, 0.75*3) {\large$T$};

    \pgfmathsetmacro{\dx}{6.9}
    \pgfmathsetmacro{\dy}{0}
    \draw [myarrows, color = matcha!150](-0.5+\dx, -7.35+\dy) -- (8+\dx, -7.35+\dy);  
    \node[anchor=north] at (3.7+\dx, -7.4+\dy) {\textcolor{matcha!150}{\large\textbf{Feature Channel}}};



		\pgfmathsetmacro{\x}{19.2}
		\pgfmathsetmacro{\y}{1.64}
		\pgfmathsetmacro{\dx}{1}
		\pgfmathsetmacro{\dya}{0.65} 
		\pgfmathsetmacro{\dyb}{0.52} 
		
		 \draw[thick, color = black] (\x-0.05, \y + 0.35) -- (29.3, \y + 0.35);
		 \node at (15.52, 2) {\hypertarget{alg}{\color{white}~}}; 
		\node at (\x+3.15, \y){\begin{tabular}{c}{\large {\bf{Alg. 1: }} Pseudocode for \texttt{Brain-ID}}\end{tabular}};
		 \draw[thick, color = black] (\x-0.05, \y - 0.3) -- (29.3, \y - 0.3);
		 
		 \node at (\x + 2.94, \y - \dya - 0.02) {\begin{tabular}{c}{\large \color{olive!110}{\# $\mathcal{F}$: feature extraction network}} \end{tabular}};
		 \node at (\x + 2.48, \y - \dya - 0.9*\dyb- 0.02) {\small\begin{tabular}{c}{\large \color{olive!110}{\large \# $\mathcal{L}$: linear activation layer}} \end{tabular}};
		 \node at (\x + 4.87, \y - \dya - 1.8*\dyb- 0.02) {\small\begin{tabular}{c}{\large \color{olive!110}{\# $\{ \Theta_i\}_1^N$: generation params (contrast, resolution, ...)}} \end{tabular}};

		 \draw[thin, color = black] (\x + 0.68, \y - \dya - 3.5 * \dyb) -- (\x + 0.68, \y - \dya - 16. * \dyb)  -- (\x + 0.68 + 0.18, \y - \dya - 16. * \dyb) ;
   
		 \node at (\x + 0.2, \y - \dya - 3 * \dyb) {\begin{tabular}{c}{\small{\bf{1}}}\end{tabular}};
		 \node at (\x + 4.65, \y - \dya - 3 * \dyb) {\small\begin{tabular}{c}{\large \hypertarget{code: loader}{{\bf{for }}$(L,~T)$ in loader:~~\color{olive!110}{\# load 1 subject sample}}} \end{tabular}};
 
		 \node at (\x + 0.2, \y - \dya - 4 * \dyb) {\begin{tabular}{c}{\small{\bf{2}}}\end{tabular}};
		 \node at (\x + 4.7, \y - \dya - 4 * \dyb) {\small\begin{tabular}{c}{\large \color{olive!110}\# generate subject-level deformation field} \end{tabular}};
	   
		 \node at (\x + 0.2, \y - \dya - 5 * \dyb) {\begin{tabular}{c}{\hypertarget{code: 3}{\small{\bf{3}}}} \end{tabular}};
		 \node at (\x + 4.9, \y - \dya - 5 * \dyb) {\small\begin{tabular}{c}{\large $\phi = \textit{get}\_\textit{random}\_\textit{deformation()}$ via \cref{eq: deform}} \end{tabular}};
   
		 \node at (\x + 0.2, \y - \dya - 6 * \dyb) {\begin{tabular}{c}{\hypertarget{code: 4}{\small{\bf{4}}}} \end{tabular}};
		 \node at (\x + 4.15, \y - \dya - 6 * \dyb) {\small\begin{tabular}{c}{\large \color{olive!110}\# generate $N$ intra-subject samples} \end{tabular}};
		 
		 \node at (\x + 0.2, \y - \dya - 7 * \dyb) {\begin{tabular}{c}{\small{\bf{5}}} \end{tabular}};
		 \node at (\x + 2.7, \y - \dya - 7 * \dyb) {\small\begin{tabular}{c}{\large intra$\_$samples = []} \end{tabular}};

		 \draw[thin, color = black] (\x + 1.25, \y - \dya - 8.5 * \dyb) -- (\x + 1.25, \y - \dya - 10. * \dyb)  -- (\x + 1.25 + 0.18, \y - \dya - 10. * \dyb) ;
   
		 \node at (\x + 0.2, \y - \dya - 8 * \dyb) {\begin{tabular}{c}{\small{\bf{6}}} \end{tabular}};
		 \node at (\x + 4.01, \y - \dya - 8 * \dyb) {\small\begin{tabular}{c}{\large \textbf{for} $(i, ~\Theta_i)$ in enumerate$\big(\{\Theta_i\}_1^N\big)$:} \end{tabular}};
		 
		 \node at (\x + 0.2, \y - \dya - 9 * \dyb) {\begin{tabular}{c}{\hypertarget{code: 7}{\small{\bf{7}}}} \end{tabular}};
		 \node at (\x + 4.45, \y - \dya - 9 * \dyb) {\small\begin{tabular}{c}{\large $S_i = \textit{random}\_\textit{generate}(L\,|\,\phi, \,\Theta_i)$} \end{tabular}};
		 
		 \node at (\x + 0.2, \y - \dya - 10 * \dyb) {\begin{tabular}{c}{\hypertarget{code: 8}{\small{\bf{8}}}} \end{tabular}};
		 \node at (\x + 3.9, \y - \dya - 10 * \dyb) {\small\begin{tabular}{c}{\large intra$\_$samples.\textit{append}($S_i$)} \end{tabular}};
		 		 

		 \node at (\x + 0.2, \y - \dya - 11 * \dyb) {\begin{tabular}{c}{\hypertarget{code: 9}{\small{\bf{9}}}} \end{tabular}};
		\node at (\x + 4.55, \y - \dya - 11 * \dyb) {\small\begin{tabular}{c}{\large \color{olive!110}\# formulate all samples to a mini-batch} \end{tabular}};

		 \node at (\x + 0.14, \y - \dya - 12 * \dyb) {\begin{tabular}{c}{\hypertarget{code: 10}{\small{\bf{10}}}} \end{tabular}};
		 \node at (\x + 3.42, \y - \dya - 12 * \dyb) {\small\begin{tabular}{c}{\large $\mathbf{x} = concat(\text{intra}\_\text{samples})$} \end{tabular}};
		 
		 \node at (\x + 0.144, \y - \dya - 13 * \dyb) {\begin{tabular}{c}{\small{\bf{11}}} \end{tabular}};
		 \node at (\x + 3.83, \y - \dya - 13 * \dyb) {\small\begin{tabular}{c}{\large $\mathbf{F} = \mathcal{F}(\mathbf{x})$~~{\color{olive!110}\# feature extraction}} \end{tabular}}; 
		 
		 \node at (\x + 0.14, \y - \dya - 14 * \dyb) {\begin{tabular}{c}{\small{\bf{12}}} \end{tabular}};
		 \node at (\x + 5.03, \y - \dya - 14 * \dyb) {\small\begin{tabular}{c}{\large {\color{olive!110}\# subject-robustness enforcement via \cref{eq: Brain-loss}}} \end{tabular}};

		 \node at (\x + 0.14, \y - \dya - 15 * \dyb) {\begin{tabular}{c}{\small{\bf{13}}} \end{tabular}};
		 \node at (\x + 2.71, \y - \dya - 15 * \dyb) {\small\begin{tabular}{c}{\large loss = $L(\mathcal{L}(\mathbf{F}),~T)$} \end{tabular}};
		  
		 \node at (\x + 0.14, \y - \dya - 16 * \dyb) {\begin{tabular}{c}{\small{\bf{14}}} \end{tabular}};
		 \node at (\x + 2.43, \y - \dya - 16 * \dyb) {\small\begin{tabular}{c}{\large loss$.backward()$} \end{tabular}};
		 
		 \draw[thick, color = black] (\x-0.05, \y - \dya - 16.7 * \dyb) -- (29.3, \y - \dya - 16.7 * \dyb);

	\end{tikzpicture}
	}   
	\vspace{-0.6cm}
\caption{\texttt{Brain-ID}'s contrast-agnostic anatomical representation learning framework.}  
	 \label{fw: train}
\end{figure*}

%% file: sec/method/trainer.tex
\vspace{-0.3cm}
\subsection{Extract Robust and Expressive Subject-specific Features}
\label{sec: framework}

\vspace{-0.2cm}
As mentioned above, we would like the resulting features from \texttt{Brain-ID} to be both robust to intra-subject variations \textit{and} expressive to potential downstream tasks. In this section, we introduce \texttt{Brain-ID}'s learning framework to achieve the two desired properties.

\vspace{-0.5cm}
\subsubsection{Intra-subject Data Generation}
\label{parag: sample}
In order to learn a feature representation that is distinctive to each subject and robust to varying MR contrasts, we observe that enriching \textit{intra}-subject samples leads to better performance~(\cref{tab: ablat}). Specifically, instead of including multiple subjects for a mini-batch during each training iteration as in usual practice, \texttt{Brain-ID} focuses on maximizing the intra-subject variance to improve the subject-robustness of resulting features. As described in Alg.~\hyperlink{alg}{1} (line \hyperlink{code: 3}{4}-\hyperlink{code: 9}{10}), for each training iteration, after randomly selecting a subject and generating its deformation, \texttt{Brain-ID} generates a mini-batch of intra-subject samples ($\{ S_1, \, \dots, \, S_N \}$) with randomly synthesized contrasts, resolutions and corruptions~(\cref{sec: generator}). As will be introduced below, \texttt{Brain-ID} collects losses from all intra-subject samples and conducts back-propagation \textit{at once},  to encourage the subject-specific robustness of its learned features.

We set the intra-subject samples within a mini-batch to have random contrasts and ``mild-to-severe'', \textit{increasing} level of corruptions (\cref{fw: train} (left)), to maximize the intra-subject variance while ensuring the stability of the training process against extreme corruption levels. (In \cref{exp: ablation}, we compare various data generation designs, and provide insights on preventing unstable training.)


\vspace{-0.5cm}
\subsubsection{Anatomy-guided Feature Representation}
\label{parag: supervision}

A richer learning space helps improve the representation robustness against the variance of sample appearances, but not enough for extracting expressive features. For example, a mapping that projects all inputs to zero is perfectly robust, but it does not provide any useful information for potential downstream tasks. Therefore, proper guidance is crucial for feature representation as well. In CIFL~\cite{chua2023contrast}, the authors use a contrastive loss on feature channels to encourage a more discriminative feature representation. However, this method is insufficient, given the extremely complex nature of the human brain anatomy (See comparisons in \cref{sec: exp}). Instead, we propose to use the standard T1w high-resolution structural MR contrast for brain morphometry, i.e., MP-RAGE (magnetization-prepared rapid gradient-echo), as the unique anatomy target to guide \texttt{Brain-ID}'s feature representation learning.

As shown in \cref{fw: train}, a feature extraction backbone ($\mathcal{F}$) first maps the input mini-batch of intra-subject generated samples, $\{ S_1, \, \dots, \, S_N \}$, to their corresponding feature space, $\{ \mathbf{F}_1, \, \dots, \, \mathbf{F}_N \}$. An linear activation layer ($\mathcal{L}$) then projects the features to the current subject's standard contrast (MP-RAGE) space, $T$. The training loss is obtained by summing over the anatomy reconstruction loss of all intra-subject samples in the current mini-batch:
\vspace{-0.2cm}
\begin{equation}
    L = \sum\nolimits_{i}^N ~ \vert \mathcal{L} (\mathbf{F}_i) - T \vert + \lambda \, \vert \nabla \mathcal{L} (\mathbf{F}_i) - \nabla T \vert\,,~\lambda \in R^+\,,
    \label{eq: Brain-loss}
\vspace{-0.2cm}
\end{equation}
where $T$, as the high-resolution, unique anatomy target for all intra-subject samples, encourages the \textit{similarities} of the features extracted from the same subject via reconstruction (1$^{\text{st}}$ term) and gradient difference (2$^{\text{nd}}$ term) \texttt{L1} loss. \texttt{Brain-ID} uses MP-RAGE as the learning target, which formulates both super-resolution and contrast-synthesis problems, and encourages a richer feature representation. (\cref{exp: ablation} provides further insights on how different choices of anatomy guidance would affect the resulting features' properties.)

%% file: sec/method/downstream.tex
\vspace{-0.15cm}
\subsection{Adapting \texttt{Brain-ID} to Downstream Tasks by One Layer}
\label{sec: downstream}
\vspace{-0.15cm}
With the intra-subject data generation and anatomy-guided feature learning design introduced in \cref{sec: framework}, a well-trained \texttt{Brain-ID} model is able to extract robust, high-resolution anatomical features from images with varying deformations, contrasts, resolutions, and artifacts. To minimize the modifications, we propose straightforward \textit{one-layer} solutions adapting \texttt{Brain-ID} features to various brain imaging applications. We later demonstrate that the simple adaptions are effective enough for \texttt{Brain-ID} to achieve state-of-the-art performance across all downstream tasks~(\cref{exp: downstream}), even for small datasets
~(\cref{exp: dataset_size}).

\vspace{-0.45cm}
\subsubsection{Contrast-independent Tasks}
\label{sec: independent}
For tasks where the output should be independent of the input MR contrast, e.g., brain segmentation, we simply add an additional layer following \texttt{Brain-ID} features ($\mathbf{F}$), and fine-tune the model:
\vspace{-0.2cm}
\begin{equation}
    L = task\_loss\_func \big(\mathcal{L} (\mathbf{F}),\, T \big) \,,
    \label{eq: independent-loss}
    \vspace{-0.2cm}
\end{equation}
where $\mathcal{L}$ and $T$ refer to the task-specific activation layer and the contrast-independent ground truth, respectively.

\vspace{-0.5cm}
\subsubsection{Contrast-dependent Tasks}
\label{sec: dependent}
Since \texttt{Brain-ID} features are contrast-agnostic and robust to artifacts, for tasks relevant to the input's contrast/quality, e.g., super-resolution, we concatenate the input image ($I$) that contains the original contrast information with its high-resolution \texttt{Brain-ID} features along the channel dimension, before forwarding into the task-specific activation layer:
\vspace{-0.2cm}
\begin{equation}
    L = task\_loss\_func \big(\mathcal{L} (\mathbf{F} \oplus I),\, T \big) \,,
    \label{eq: dependent-loss}
    \vspace{-0.2cm}
\end{equation}
where $\mathcal{L}$ and $T$ refer to the task-specific activation layer and the contrast-dependent ground truth, respectively.

%% file: sec/exp/main.tex
\vspace{-0.15cm}

\section{Experiments}
\label{sec: exp}
\vspace{-0.2cm}
In this section, we conduct experiments to demonstrate the two properties of \texttt{Brain-ID} as claimed in \cref{sec: method}. \textit{(i)} Robustness~(\cref{exp: feat}), where we propose ``canonical'' and ``atlas-registered'' features to assess both intra- and inter-subject robustness of \texttt{Brain-ID} features; \textit{(ii)} Expressiveness~(\cref{exp: downstream}), where we evaluate the performance of adapting \texttt{Brain-ID} features to a series of common brain imaging applications. We further challenge \texttt{Brain-ID} with reduced-size datasets to explore its ability when limited real data is available for training (\cref{exp: dataset_size}).

\input{sec/exp/setup}


\input{sec/exp/feature} 


\input{sec/exp/fig/fig_task}
\input{sec/exp/fig/tab_task}

\vspace{-0.15cm}
\subsection{Downstream Evaluation}
\vspace{-0.15cm}
\label{exp: downstream}
Following the standard evaluation protocol for feature representation learning~\cite{Caron2021EmergingPI}, we fine-tune \texttt{Brain-ID} features for downstream tasks by the proposed one-layer adaption~(\cref{sec: downstream}). (More setup details and results are in \cref{app: task}.)

\vspace{0.2cm}
\noindent \textbf{Models}~We compare the downstream performance of \texttt{Brain-ID} features with: \textit{(i-ii)}~\texttt{SAMSEG}~\cite{Puonti2016FastAS,Cerri2020ACM} and \texttt{FastSurfer}~\cite{Henschel2019FastSurferA} (only works on T1w), state-of-the-art classical and machine-learning-based brain segmentation models, respectively; \textit{(iii)}~\texttt{SynthSR}~\cite{Iglesias2023SynthSRAP}, state-of-the-art, contrast-agnostic T1w synthesis model. We also provide fine-tuned \texttt{SynthSR} (\texttt{SynthSR}-FT) results for further comparison; \textit{(iv)}~\texttt{SCRATCH}: a baseline with the same data generation/architecture as \texttt{Brain-ID}, yet trained from \textit{scratch}. \texttt{SCRATCH} is used to validate \texttt{Brain-ID}'s effectiveness. 
\textit{(v)}~\hyperlink{cifl}{\texttt{CIFL}} \cite{chua2023contrast}: a baseline with the same data generation/architecture as \texttt{Brain-ID} and \texttt{SCRATCH}, yet initialized with \texttt{CIFL}'s pre-trained features. We use \texttt{CIFL} to demonstrate the superiority of \texttt{Brain-ID}'s anatomy-guided representation.

\input{sec/exp/contrast_inde} 

\input{sec/exp/fig/fig_comp}

\input{sec/exp/contrast_de} 
\input{sec/exp/additional}


%% file: sec/exp/setup.tex
\vspace{-0.3cm}
\subsection{Datasets, Metrics and Implementation Details}
\label{sec: setup}


\vspace{-0.1cm}
\subsubsection{Datasets}
\label{sec: datasets}
For \texttt{Brain-ID} features pre-training, we use 2045 3D segmentations (``anatomies'', or IDs) from the public ADNI dataset~\cite{Jack2008TheAD}. For downstream evaluation, we use six public datasets covering T1w and T2w MRI, FLAIR MRI, and CT: ADNI~\cite{Jack2008TheAD}, ADNI3~\cite{Weiner2017TheAD}, HCP~\cite{Essen2012TheHC}, ADHD200~\cite{Brown2012ADHD200GC}, AIBL~\cite{Fowler2021FifteenYO}, OASIS3~\cite{LaMontagne2018OASIS3LN}.

\vspace{-0.5cm}
\subsubsection{Metrics}
\label{sec: metrics}
We evaluate individual tasks from different aspects. For feature similarity measurements, we use \texttt{L1} distance, and \texttt{(MS-)SSIM} (multi-scale structural similarity)~\cite{Wang2003MultiscaleSS,Wang_2021_CVPR,Liu2021SelfappearanceaidedDE}. For reconstruction and super-resolution, we use \texttt{L1}, \texttt{PSNR} (peak signal-to-noise ratio) and \texttt{(MS-)SSIM}. For segmentation, we use Dice scores. For bias field estimation, we use the normalized \texttt{L2} distance (\texttt{norm-L2}) to avoid possible arbitrary scalings from nonuniformity correction~\cite{Chua2009EvaluationOP}. (\cref{app: dataset} contains further details on metrics and datasets preprocessing.)


\vspace{-0.5cm}
\subsubsection{Implementation Details}
\label{sec: implement}
As a general feature representation model, \texttt{Brain-ID} can use any backbone to extract brain features. For fairer comparison, we adopt the same five-level 3D UNet~\cite{Ronneberger2015UNetCN} (with 64 feature channels in the last layer) as utilized in state-of-the-art models we compare with in~\cref{exp: downstream}. During feature pre-training~(\cref{sec: framework}), a linear regression layer is added for anatomy supervision (\cref{eq: Brain-loss}). For downstream adaptions, a task-specific adaption layer is added (\cref{sec: downstream}). (More implementation details and training recipes are in \cref{app: implement}.)

%% file: sec/exp/feature.tex
\vspace{-0.2cm}
\subsection{Intra/Inter-subject Feature Robustness}
\label{exp: feat}
\vspace{-0.1cm}
In this section, we examine the robustness of \texttt{Brain-ID} features, i.e., the 64 features from the last layer. For comparable and reproducible results, we use our data generator (\cref{sec: generator}) to prepare (deform and corrupt) 1000 testing samples from 100 randomly selected subjects in ADNI~\cite{Jack2008TheAD} (T1w) testing set, with 10 intra-subject samples for each subject. (\hypertarget{cifl}{More} details are in \cref{app: feat}.)

We compare our features with \texttt{CIFL}~\cite{chua2023contrast} which is, to our best knowledge, the only similar work to \texttt{Brain-ID}, that also aims to learn contrast-agnostic brain features. Note the original \texttt{CIFL} method has only experimented on 2D images, and does not have our intra-subject data generation design. For a fairer comparison, we trained \texttt{CIFL} \emph{with its own contrastive-learning design}, while using the same data generator and network architecture as \texttt{Brain-ID}.

\input{sec/exp/fig/fig_feat}
\input{sec/exp/fig/tab_feat}

\vspace{-0.1cm}
\subsubsection{Intra-subject Feature Robustness}
\label{exp: feat-intra}
\vspace{-0.2cm}
Ideally, \texttt{Brain-ID} features computed from the same subject should be structurally \textit{identical}, regardless of their appearances (contrast, resolution, noises, etc). To assess the intra-subject feature robustness, we first define their corresponding ``canonical'' features.

\noindent \textit{``Canonical'' Features.}~We compute $\phi^{-1}$, the inverse of $S$' generated deformation as in \cref{eq: deform}, and map back each sample's ($S$) \texttt{Brain-ID} features ($\mathbf{F} = \mathbf{F}(S \, \vert \, \phi)$) to their original domain without any deformation:
\vspace{-0.2cm}
\begin{equation}
    \mathbf{F}^{-1} = \phi^{-1} \circ \mathbf{F} (S\, \vert \, \phi)\,.
    \label{eq: inv_deform}
\vspace{-0.2cm}
\end{equation}

As shown in Fig.~\hyperlink{fig: feat_dist-a}{4}-(a), for the same subject (ID-$i$), although the inputs (row) are processed by varying deformations and corruptions, their \texttt{Brain-ID}'s ``canonical'' features are of similar looking along each feature channel (column). In \cref{tab: feat}, we quantify the feature similarity: For each testing subject, we first generate a sample \textit{without} any deformation or corruption and use its \texttt{Brain-ID} feature as the gold-standard reference for all other ``canonical'' features computed from intra-subject samples that may have different levels of deformation and corruptions. The final results are obtained by scores averaged over all 64 feature channels of all the intra-subject samples.

\vspace{-0.2cm}
\subsubsection{Inter-subject Feature Robustness}
\label{exp: feat-inter}
\vspace{-0.2cm}
We further assess the feature robustness among anatomies, based on the assumption that features from different subjects should ideally be \textit{structurally similar}, once being registered to a common domain.

\noindent \textit{``Atlas-registered'' Features.}~We register~\cite{Modat2010FastFD} features ($\mathbf{F} = \mathbf{F}(S \, \vert \, \phi)$) from all subjects to a common brain atlas~\cite{Fonov2009Unbiased,Fonov2011Unbiased}:
\vspace{-0.2cm}
\begin{equation}
    \overline{\mathbf{F}} = \psi_{S\to A} \circ \mathbf{F} (S\, \vert \, \phi)\, ,
    \label{eq: atlas_deform}
\vspace{-0.2cm}
\end{equation}
where $\psi_{S\to A}$ denotes the mapping from $S$ to the atlas, $A$.

As shown in Fig.~\hyperlink{fig: feat_dist-b}{4}-(b), even though the input samples (row) from different subjects have varying anatomical structures, after being registered to a common domain, their ``atlas-registered'' \texttt{Brain-ID} features still appear consistent (column). In \cref{tab: feat}, we further provide quantitative results of the structural similarity measured between features from different subjects. Similar to \cref{exp: feat-intra}, we randomly select a testing subject and generate a sample \textit{without} any deformation or corruption. We use its ``atlas-registered'' \texttt{Brain-ID} feature as the gold standard for all ``atlas-registered'' features computed from other samples that may have different levels of deformation and corruption. The final results are obtained by scores averaged over all samples of the 100 testing subjects. 


In both intra/inter-subject aspects, \texttt{Brain-ID} outperforms \texttt{CIFL} by a large margin, which validates its feature robustness against deformations, contrasts, resolutions, artifacts (intra-subject), and anatomies (inter-subject). 

%% file: sec/exp/fig/fig_feat.tex
\begin{figure}[t]

\centering 

\resizebox{0.95\linewidth}{!}{
	\begin{tikzpicture} 
        
		\tikzstyle{myarrows}=[line width=0.6mm,draw=blue!50,-triangle 45,postaction={draw, line width=0.05mm, shorten >=0.02mm, -}]
		\tikzstyle{mylines}=[line width=0.8mm]
  


	\pgfmathsetmacro{\shift}{-1.}
 
        \node at (\shift+1.05, 3+0.35,0.1*3) {\hypertarget{fig: feat_dist-a}{~}};
 
	\foreach \i in {-1, 1, 3}
	{
	\node at (\shift+1.05,\i+0.15,0.1*3) {$\mathbf{F}^{-1}$};
	\node at (\shift+1.05,\i-0.45,0.1*3) {\small\cref{eq: inv_deform}};
        \draw [->, color = matcha!150, line width = 0.5mm](\shift+0.6,\i-0.15,0.1*3) -- (\shift+1.5,\i-0.15,0.1*3);  
	}

        \node at (\shift+1.05, -4+0.35,0.1*3) {\hypertarget{fig: feat_dist-b}{~}};
	\foreach \i in {-8, -6, -4}
	{
	\node at (\shift+1.05,\i+0.15,0.1*3) {$\overline{\mathbf{F}}$};
	\node at (\shift+1.05,\i-0.45,0.1*3) {\small\cref{eq: atlas_deform}};
        \draw [->, color = matcha!150, line width = 0.5mm](\shift+0.6,\i-0.15,0.1*3) -- (\shift+1.5,\i-0.15,0.1*3);  
	}
 
	\node at (\shift+0.1*3, -9.3+0.15*3, 0.75*3) {Input};
	\node at (\shift+0.1*3, -8+0.15*3, 0.75*3) {\includegraphics[width=0.16\textwidth]{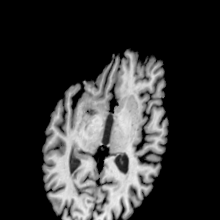}};
	\node at (\shift+0.1*3, -6+0.15*3, 0.75*3) {\includegraphics[width=0.16\textwidth]{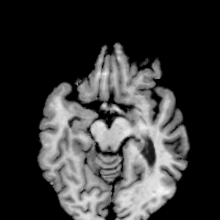}};
	\node at (\shift+0.1*3, -4+0.15*3, 0.75*3) {\includegraphics[width=0.16\textwidth]{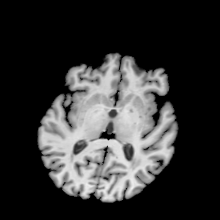}};
 
	\node at (\shift+0.1*3, -2.3+0.15*3, 0.75*3) {Input};
	\node at (\shift+0.1*3, -1+0.15*3, 0.75*3) {\includegraphics[width=0.16\textwidth]{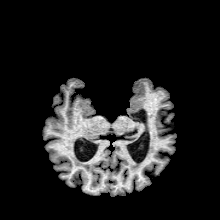}};
	\node at (\shift+0.1*3, 1+0.15*3, 0.75*3) {\includegraphics[width=0.16\textwidth]{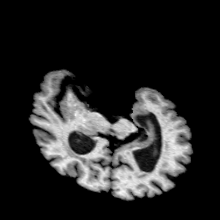}};
	\node at (\shift+0.1*3, 3+0.15*3, 0.75*3) {\includegraphics[width=0.16\textwidth]{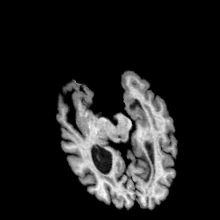}};
 
        \node at (0.+\shift+0.3-1.9, 1+0.15*3, 0.75*3) {ID-$i$};
        
        \node at (0.+\shift+0.6-1.9, -4+0.15*3, 0.75*3) {ID-$ii$}; 
        \node at (0.+\shift+0.6-1.9, -6+0.15*3, 0.75*3) {ID-$iii$};
        \node at (0.+\shift+0.6-1.9, -8+0.15*3, 0.75*3) {ID-$iv$};
 
    \draw [decorate,decoration={brace,amplitude=5pt,mirror,raise=6ex},line width=2.pt,color = gray] (-0.9+\shift, 3.5) -- (-0.9+\shift, -2.3);
    
    \pgfmathsetmacro{\dx}{2.5}
    \pgfmathsetmacro{\dy}{-2.1}
    \draw [myarrows, color = matcha!150](-2.1+\dx, -7.45+\dy) -- (9.+\dx, -7.45+\dy);  
    \node[anchor=north] at (1.+\dx, -7.6+\dy+0.1) {(b)};
    \node[anchor=north] at (3.+\dx, -7.6+\dy+0.1) {\textcolor{matcha!150}{\textbf{Feature Channel}}}; 
    \pgfmathsetmacro{\dy}{4.9}
    \draw [myarrows, color = matcha!150](-2.1+\dx, -7.45+\dy) -- (9.+\dx, -7.45+\dy);  
    \node[anchor=north] at (1.+\dx, -7.6+\dy+0.1) {(a)};
    \node[anchor=north] at (3.+\dx, -7.6+\dy+0.1) {\textcolor{matcha!150}{\textbf{Feature Channel}}}; 

 
 
	\pgfmathsetmacro{\shift}{-3.2}

	\node at (5.2+\shift+0.1*3, 3+0.15*3, 0.75*3) {\includegraphics[width=0.16\textwidth]{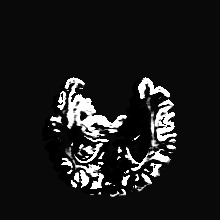}};
	\node at (5.2+\shift+0.1*3, 1+0.15*3, 0.75*3) {\includegraphics[width=0.16\textwidth]{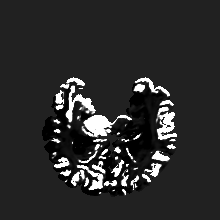}};
	\node at (5.2+\shift+0.1*3, -1+0.15*3, 0.75*3) {\includegraphics[width=0.16\textwidth]{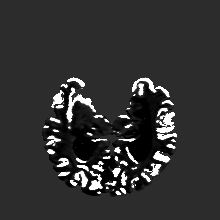}};
	\node at (5.2+\shift+0.1*3, -4+0.15*3, 0.75*3) {\includegraphics[width=0.16\textwidth]{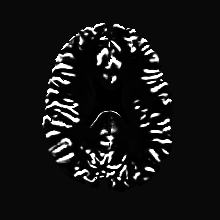}};
	\node at (5.2+\shift+0.1*3, -6+0.15*3, 0.75*3) {\includegraphics[width=0.16\textwidth]{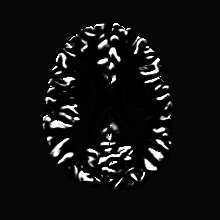}};
	\node at (5.2+\shift+0.1*3, -8+0.15*3, 0.75*3) {\includegraphics[width=0.16\textwidth]{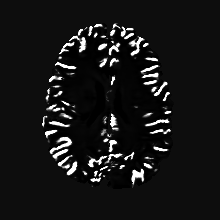}};
        
	\node at (7.2+\shift+0.1*3, 3+0.15*3, 0.75*3) {\includegraphics[width=0.16\textwidth]{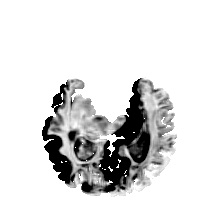}};
	\node at (7.2+\shift+0.1*3, 1+0.15*3, 0.75*3) {\includegraphics[width=0.16\textwidth]{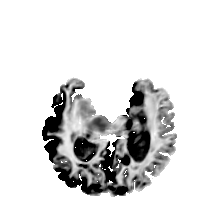}};
	\node at (7.2+\shift+0.1*3, -1+0.15*3, 0.75*3) {\includegraphics[width=0.16\textwidth]{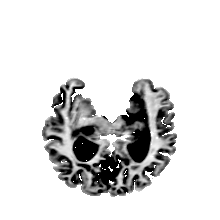}};
	\node at (7.2+\shift+0.1*3, -4+0.15*3, 0.75*3) {\includegraphics[width=0.16\textwidth]{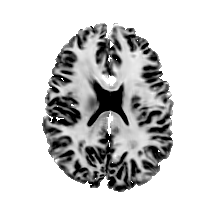}};
	\node at (7.2+\shift+0.1*3, -6+0.15*3, 0.75*3) {\includegraphics[width=0.16\textwidth]{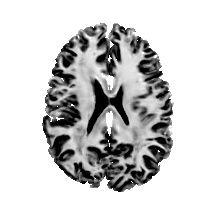}};
	\node at (7.2+\shift+0.1*3, -8+0.15*3, 0.75*3) {\includegraphics[width=0.16\textwidth]{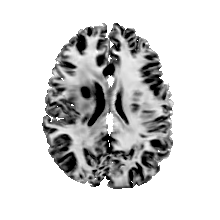}};
 
	\node at (9.2+\shift+0.1*3, 3+0.15*3, 0.75*3) {\includegraphics[width=0.16\textwidth]{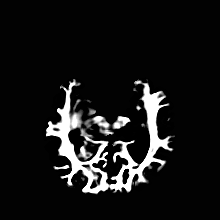}};
	\node at (9.2+\shift+0.1*3, 1+0.15*3, 0.75*3) {\includegraphics[width=0.16\textwidth]{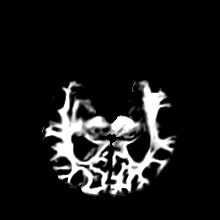}};
	\node at (9.2+\shift+0.1*3, -1+0.15*3, 0.75*3) {\includegraphics[width=0.16\textwidth]{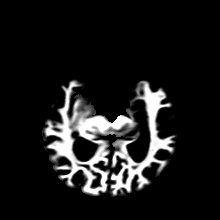}};
	\node at (9.2+\shift+0.1*3, -4+0.15*3, 0.75*3) {\includegraphics[width=0.16\textwidth]{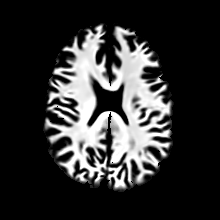}};
	\node at (9.2+\shift+0.1*3, -6+0.15*3, 0.75*3) {\includegraphics[width=0.16\textwidth]{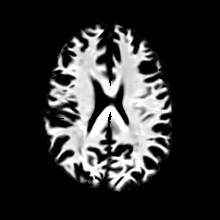}};
	\node at (9.2+\shift+0.1*3, -8+0.15*3, 0.75*3) {\includegraphics[width=0.16\textwidth]{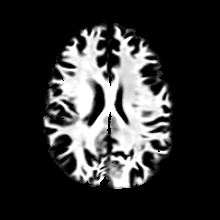}};
 
	\node at (11.2+\shift+0.1*3, 3+0.15*3, 0.75*3) {\includegraphics[width=0.16\textwidth]{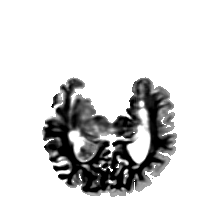}};
	\node at (11.2+\shift+0.1*3, 1+0.15*3, 0.75*3) {\includegraphics[width=0.16\textwidth]{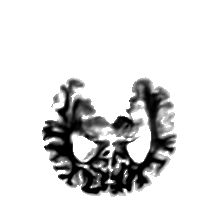}};
	\node at (11.2+\shift+0.1*3, -1+0.15*3, 0.75*3) {\includegraphics[width=0.16\textwidth]{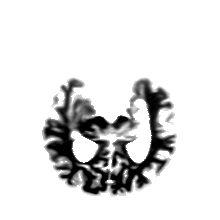}};
	\node at (11.2+\shift+0.1*3, -4+0.15*3, 0.75*3) {\includegraphics[width=0.16\textwidth]{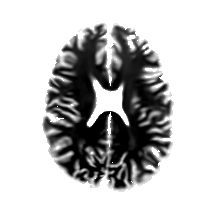}};
	\node at (11.2+\shift+0.1*3, -6+0.15*3, 0.75*3) {\includegraphics[width=0.16\textwidth]{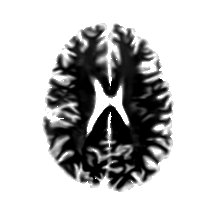}};
	\node at (11.2+\shift+0.1*3, -8+0.15*3, 0.75*3) {\includegraphics[width=0.16\textwidth]{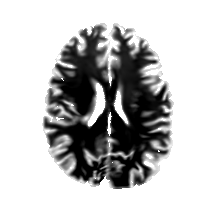}};

	\node at (13.2+\shift+0.1*3, 3+0.15*3, 0.75*3) {\includegraphics[width=0.16\textwidth]{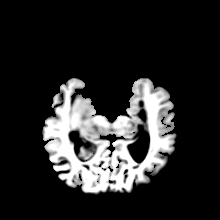}};
	\node at (13.2+\shift+0.1*3, 1+0.15*3, 0.75*3) {\includegraphics[width=0.16\textwidth]{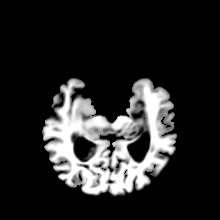}};
	\node at (13.2+\shift+0.1*3, -1+0.15*3, 0.75*3) {\includegraphics[width=0.16\textwidth]{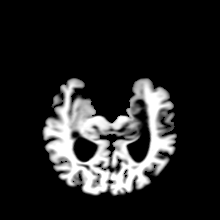}};
	\node at (13.2+\shift+0.1*3, -4+0.15*3, 0.75*3) {\includegraphics[width=0.16\textwidth]{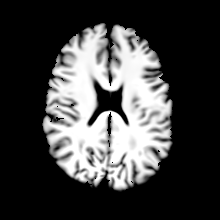}};
	\node at (13.2+\shift+0.1*3, -6+0.15*3, 0.75*3) {\includegraphics[width=0.16\textwidth]{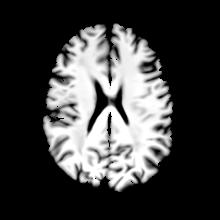}};
	\node at (13.2+\shift+0.1*3, -8+0.15*3, 0.75*3) {\includegraphics[width=0.16\textwidth]{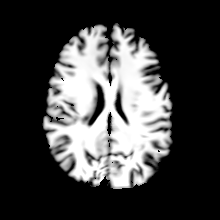}};

    \pgfmathsetmacro{\shift}{-1.2}
    \foreach \y in {-8, -6, -4, -1, 1, 3}{
    \draw[black,fill=black] (12.6+\shift+0.1*3, \y+0.15*3, 0.75*3)  circle (0.8pt);
    \draw[black,fill=black] (12.8+\shift+0.1*3, \y+0.15*3, 0.75*3)  circle (0.8pt);
    \draw[black,fill=black] (13.+\shift+0.1*3, \y+0.15*3, 0.75*3)  circle (0.8pt);
}
        

	\end{tikzpicture}
	}  
	\vspace*{-0.3cm}
\caption{(a) Intra-subject and (b) inter-subject robustness of \texttt{Brain-ID} features.}  
	 \label{fig: feat_dist}
  
\end{figure}
  

%% file: sec/exp/fig/tab_feat.tex
\begin{table}[tb]
    \caption{Evaluations on intra/inter-subject feature robustness.}
    \vspace*{-0.25cm}
    \label{tab: feat}
\resizebox{0.7\linewidth}{!}{
\centering 
    \begin{tabular}{ccccc} 
       \toprule \\[-3ex] 
       
       {\bf Mode} & {\bf Method} & \texttt{L1} ($\downarrow$) & \texttt{SSIM} ($\uparrow$) & \texttt{MS-SSIM} ($\uparrow$)  \\ [-0.2ex]
     \midrule\\[-3ex]
     
       \multicolumn{1}{c}{\multirow{2}{*}{Intra}} & {\texttt{CIFL}~\cite{chua2023contrast}} &  0.122 ($\pm 0.031$)  & 0.511 ($\pm 0.298$) & 0.531 ($\pm 0.367$)  \\ 
        & {\texttt{Brain-ID}} & \textbf{0.011} ($\pm 0.001$)  & \textbf{0.858} ($\pm 0.041$) & \textbf{0.921} ($\pm 0.008$)  \\ 
      \hline\\[-2.3ex]
      
        \multicolumn{1}{c}{\multirow{2}{*}{Inter}} & 
 {\texttt{CIFL}~\cite{chua2023contrast}} & 0.230 ($\pm 0.039$) & 0.552 ($\pm 0.240$) & 0.524 ($\pm 0.291$)  \\ 
        & {\texttt{Brain-ID}} & \textbf{0.014} ($\pm 0.013$) & \textbf{0.843} ($\pm 0.044$) & \textbf{0.913} ($\pm 0.011$) \\ 
      
\bottomrule  \\ [-3.6ex] 
    \end{tabular} 
}
\end{table}

%% file: sec/exp/fig/fig_task.tex
\begin{figure}[t]

\centering 

\resizebox{0.96\linewidth}{!}{
	\begin{tikzpicture}
        \hspace*{-0.5cm}
        
		\tikzstyle{myarrows}=[line width=0.8mm,draw=blue!50,-triangle 45,postaction={draw, line width=0.05mm, shorten >=0.02mm, -}]
		\tikzstyle{mylines}=[line width=0.8mm]
  


	\pgfmathsetmacro{\shift}{-0.9}

 
	\node at (\shift+1.45, 5.2+0.15*3, 0.75*3) {Recon};
	\node at (\shift+1.45, 4.8+0.15*3, 0.75*3) {(FLAIR)};
	\node at (\shift+1.45, 3.2+0.15*3, 0.75*3) {SR};
	\node at (\shift+1.45, 2.8+0.15*3, 0.75*3) {(T1w)};
	\node at (\shift+1.45, 1.2+0.15*3, 0.75*3) {Seg};
	\node at (\shift+1.45, 0.8+0.15*3, 0.75*3) {(CT)};
        
	\pgfmathsetmacro{\shift}{-3.2}
	\node at (5.2+\shift+0.1*3, 6.2+0.15*3, 0.75*3) {Input};
	\node at (7.2+\shift+0.1*3, 6.2+0.15*3, 0.75*3) {\footnotesize\texttt{SynthSR}/\texttt{SAMSEG}};
	\node at (9.2+\shift+0.1*3, 6.2+0.15*3, 0.75*3) {\texttt{CIFL}};
	\node at (11.2+\shift+0.1*3, 6.2+0.15*3, 0.75*3) {\texttt{SCRATCH}};
	\node at (13.2+\shift+0.1*3, 6.2+0.15*3, 0.75*3) {\texttt{Brain-ID}};
	\node at (15.2+\shift+0.1*3, 6.2+0.15*3, 0.75*3) {Ground Truth};

	\node at (5.2+\shift+0.1*3, 5+0.15*3, 0.75*3) {\includegraphics[width=0.16\textwidth]{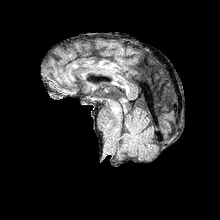}};
	\node at (5.2+\shift+0.1*3, 3+0.15*3, 0.75*3) {\includegraphics[width=0.16\textwidth]{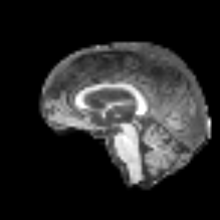}};
	\node at (5.2+\shift+0.1*3, 1+0.15*3, 0.75*3) {\includegraphics[width=0.16\textwidth]{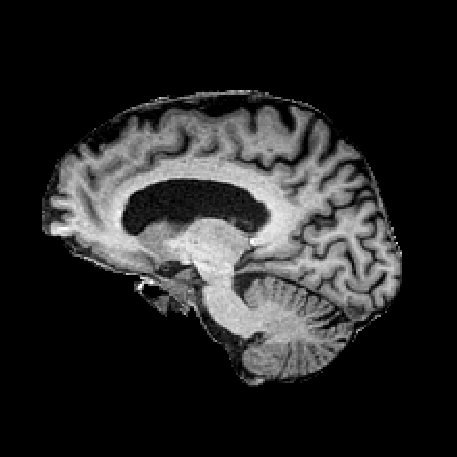}};

	\node at (7.2+\shift+0.1*3, 5+0.15*3, 0.75*3) {\includegraphics[width=0.16\textwidth]{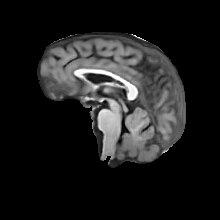}};
        \draw[fill=none, draw=mint!150, line width=0.4mm](7.2+\shift+0.1*3+0.3, 5+0.15*3-0.2, 0.75*3) circle (0.3);
	\node at (7.2+\shift+0.1*3, 3+0.15*3, 0.75*3) {\includegraphics[width=0.16\textwidth]{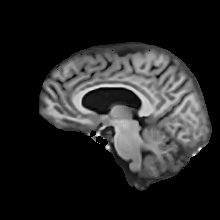}};
        \draw[fill=none, draw=mint!150, line width=0.4mm](7.2+\shift+0.1*3+0.4 , 3+0.15*3-0.3, 0.75*3) circle (0.25);
	\node at (7.2+\shift+0.1*3, 1+0.15*3, 0.75*3) {\includegraphics[width=0.16\textwidth]{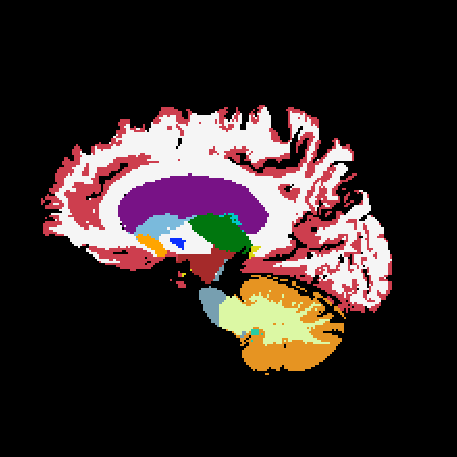}};
        \draw[fill=none, draw=mint!150, line width=0.4mm](7.2+\shift+0.1*3-0.2, 1+0.15*3+0.35, 0.75*3) circle (0.25);

	\node at (9.2+\shift+0.1*3, 5+0.15*3, 0.75*3) {\includegraphics[width=0.16\textwidth]{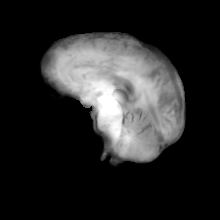}};
	\node at (9.2+\shift+0.1*3, 3+0.15*3, 0.75*3) {\includegraphics[width=0.16\textwidth]{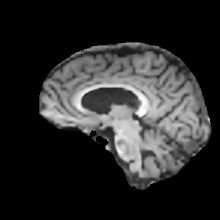}};
        \draw[fill=none, draw=mint!150, line width=0.4mm](9.2+\shift+0.1*3+0.4 , 3+0.15*3-0.3, 0.75*3) circle (0.25);
	\node at (9.2+\shift+0.1*3, 1+0.15*3, 0.75*3) {\includegraphics[width=0.16\textwidth]{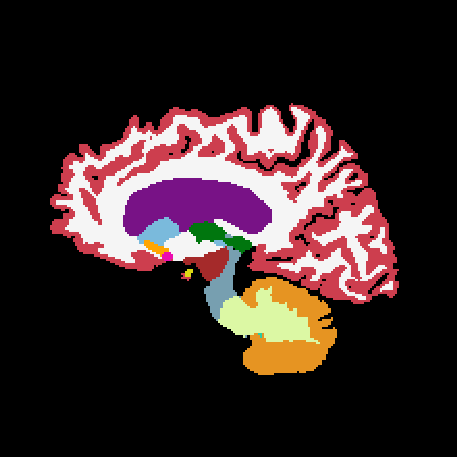}};
        \draw[fill=none, draw=mint!150, line width=0.4mm](9.2+\shift+0.1*3-0.07, 1+0.15*3-0.09, 0.75*3) circle (0.23);

	\node at (11.2+\shift+0.1*3, 5+0.15*3, 0.75*3) {\includegraphics[width=0.16\textwidth]{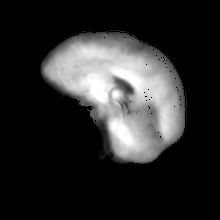}};
	\node at (11.2+\shift+0.1*3, 3+0.15*3, 0.75*3) {\includegraphics[width=0.16\textwidth]{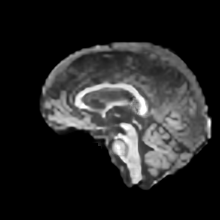}};
	\node at (11.2+\shift+0.1*3, 1+0.15*3, 0.75*3) {\includegraphics[width=0.16\textwidth]{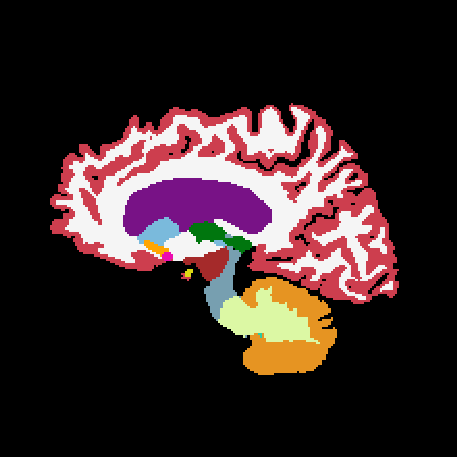}};

	\node at (13.2+\shift+0.1*3, 5+0.15*3, 0.75*3) {\includegraphics[width=0.16\textwidth]{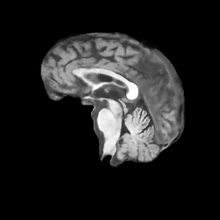}};
        \draw[fill=none, draw=mint!150, line width=0.4mm](13.2+\shift+0.1*3+0.3, 5+0.15*3-0.2, 0.75*3) circle (0.3);
	\node at (13.2+\shift+0.1*3, 3+0.15*3, 0.75*3) {\includegraphics[width=0.16\textwidth]{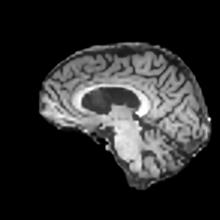}};
        \draw[fill=none, draw=mint!150, line width=0.4mm](13.2+\shift+0.1*3+0.4 , 3+0.15*3-0.3, 0.75*3) circle (0.25);
	\node at (13.2+\shift+0.1*3, 1+0.15*3, 0.75*3) {\includegraphics[width=0.16\textwidth]{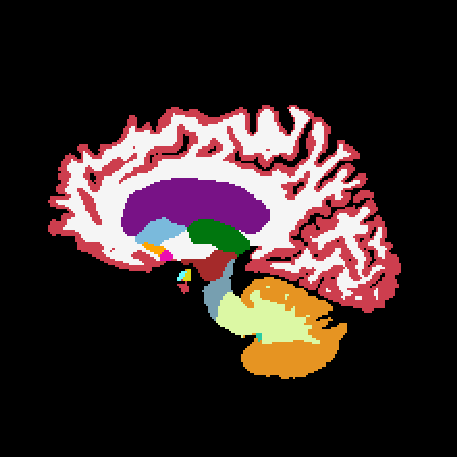}};

	\node at (15.2+\shift+0.1*3, 5+0.15*3, 0.75*3) {\includegraphics[width=0.16\textwidth]{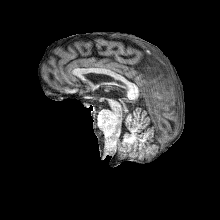}};
	\node at (15.2+\shift+0.1*3, 3+0.15*3, 0.75*3) {\includegraphics[width=0.16\textwidth]{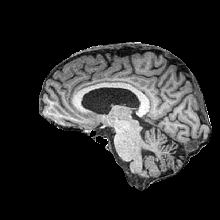}};

	\node at (5.2+\shift+0.1*3, 1+0.15*3, 0.75*3) {\includegraphics[width=0.16\textwidth]{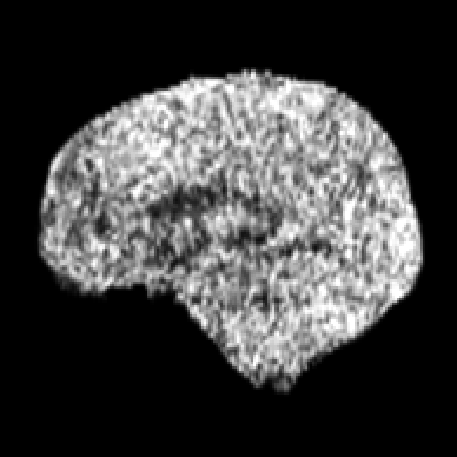}};
	\node at (7.2+\shift+0.1*3, 1+0.15*3, 0.75*3) {\includegraphics[width=0.16\textwidth]{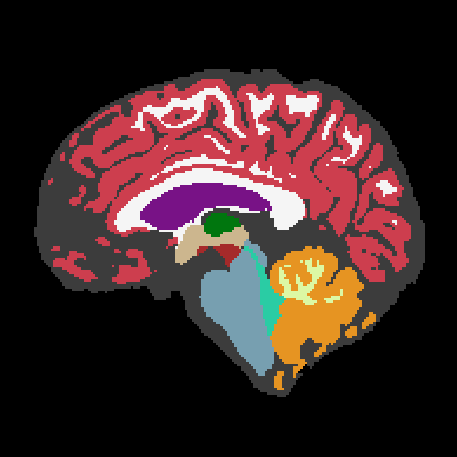}};
	\node at (9.2+\shift+0.1*3, 1+0.15*3, 0.75*3) {\includegraphics[width=0.16\textwidth]{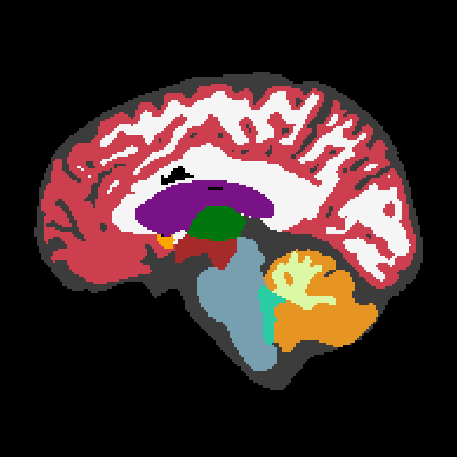}};
        \draw[fill=none, draw=mint!150, line width=0.4mm](9.2+\shift+0.1*3-0.2, 1+0.15*3+0.2, 0.75*3) circle (0.25);
	\node at (11.2+\shift+0.1*3, 1+0.15*3, 0.75*3) {\includegraphics[width=0.16\textwidth]{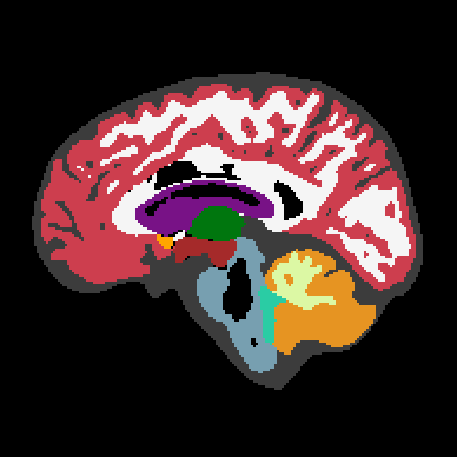}};
        \draw[fill=none, draw=mint!150, line width=0.4mm](11.2+\shift+0.1*3-0.2, 1+0.15*3+0.2, 0.75*3) circle (0.25);
        \draw[fill=none, draw=mint!150, line width=0.4mm](11.2+\shift+0.1*3+0.05, 1+0.15*3-0.3, 0.75*3) circle (0.25);
	\node at (13.2+\shift+0.1*3, 1+0.15*3, 0.75*3) {\includegraphics[width=0.16\textwidth]{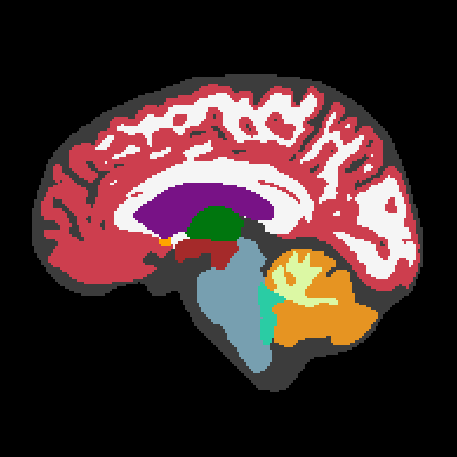}};
	\node at (15.2+\shift+0.1*3, 1+0.15*3, 0.75*3) {\includegraphics[width=0.16\textwidth]{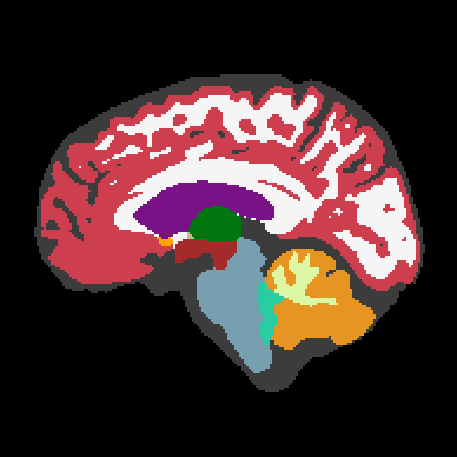}};
 
        \draw[fill=none, draw=mint!150, line width=0.4mm](7.2+\shift+0.1*3-0.64, 1+0.15*3+0., 0.75*3) circle (0.27);


	\end{tikzpicture}
	}  
	\vspace{-0.3cm}
\caption{Qualitative comparisons on downstream tasks of reconstruction (Recon), super-resolution (SR), and segmentation (Seg), between \texttt{Brain-ID}, the baseline \texttt{SCRATCH}, and the state-of-the-art methods \texttt{CIFL}~\cite{chua2023contrast}, \texttt{SynthSR}~\cite{Iglesias2023SynthSRAP} (Recon and SR), \texttt{SAMSEG}~\cite{Cerri2020ACM} (Seg). The visualized testing examples are taken from: AIBL-FLAIR~\cite{Fowler2021FifteenYO} for Recon, AIBL-T1w~\cite{Fowler2021FifteenYO} for SR, and OASIS-CT~\cite{LaMontagne2018OASIS3LN} for Seg. The \textcolor{mint!200}{mint} circles highlight some details.} 

	 \label{fig: task} 
\end{figure}

%% file: sec/exp/fig/tab_task.tex
\begin{table}[t]
    \caption{Comparisons of \texttt{Brain-ID} with state-of-the-art approaches on downstream applications. (``train/test'' refers to the number of subjects in the training/testing set.)} 
    \vspace*{-0.25cm}
    \label{tab: task}
\resizebox{0.95\linewidth}{!}{
\centering 
    \begin{tabular}{cllccccccccc}
       \toprule \\[-3ex] 
      \multicolumn{1}{c}{\multirow{2}{*}{\thead{\normalsize \textbf{Modality}\\\normalsize (Contrast)}}} & \multicolumn{1}{c}{\multirow{2}{*}{\thead{\normalsize \textbf{Dataset}\\\normalsize (Train/Test)}}} & \multicolumn{1}{c}{\multirow{2}{*}{\textbf{Method}}} & \multicolumn{4}{c}{\textbf{Reconstruction}} & \multicolumn{1}{c}{\textbf{Segmentation}} & \multicolumn{3}{c}{\textbf{Super-resolution} (from 3~/~5~/~7~$mm$)} & \multicolumn{1}{c}{\textbf{Bias Field}} \\ [-0.7ex]
        \cmidrule(lr){4-7}
        \cmidrule(lr){8-8}
        \cmidrule(lr){9-11}
        \cmidrule(lr){12-12}
         & & & \texttt{L1} ($\downarrow$) & \texttt{PSNR} ($\uparrow$) & \texttt{SSIM} ($\uparrow$) & \texttt{MS-SSIM} ($\uparrow$) & \texttt{Dice} ($\uparrow$) & \texttt{PSNR} ($\uparrow$) & \texttt{SSIM} ($\uparrow$) & \texttt{MS-SSIM} ($\uparrow$) & \texttt{norm-L2} ($\downarrow$) \\ [-0.2ex]
     \midrule[\heavyrulewidth]\\[-3ex]

        \multicolumn{1}{c}{\multirow{35}{*}{\thead{\normalsize{MR}\\ \vspace*{-0.2cm}~\\ \normalsize{(T1w)}}}} & {\multirow{7}{*}{\thead{\normalsize \textit{ADNI}~\cite{Jack2008TheAD} \\\vspace*{-0.2cm}~\\ \normalsize (1841/204)}}} 
        & {\texttt{SAMSEG}~\cite{Cerri2020ACM}} & - & - & - & - & 0.795 & - & - & - & -   \\ 
        & & {\texttt{FastSurfer}~\cite{Henschel2019FastSurferA}} & - & - & - & - &  0.803 & - & - & - &  -   \\ 
        & & {\texttt{SynthSR}~\cite{Iglesias2023SynthSRAP}} & 0.014 & 26.78 & 0.980 & 0.980 & - & 27.01~/~25.81~/~21.79 & 0.977~/~0.943~/~0.857 & 0.981~/~0.968~/~0.869 &  -   \\ %
        & & {\texttt{SynthSR}-FT~~} & \textbf{0.010} & 29.54 & 0.984 & 0.920 & 0.825 & 30.46~/~28.03~/~23.91 & 0.979~/~0.960~/~0.898 & 0.985~/~0.973~/~0.900 &  0.051   \\ %
        & & {\texttt{CIFL}~\cite{chua2023contrast}}& 0.013 & 26.97 & 0.973 & 0.961 & 0.820 & 30.12~/~27.21~/~23.80 & 0.978~/~0.912~/~0.887 & 0.979~/~0.950~/~0.842 & 0.053  \\ %
        & & {\texttt{SCRATCH}} & 0.011 & 27.54 & 0.979 & 0.957 & 0.816 & 30.29~/~27.94~/~24.57 & 0.975~/~0.959~/~0.920  & 0.982~/~0.964~/~0.866 & 0.056   \\ %
        & & {\texttt{Brain-ID}} & 0.012 & \textbf{33.82} & \textbf{0.993} & \textbf{0.989} & \textbf{0.837} & \textbf{31.30}~/~\textbf{28.62}~/~\textbf{24.68} & \textbf{0.983}~/~\textbf{0.961}~/~\textbf{0.928}  & \textbf{0.987}~/~\textbf{0.980}~/~\textbf{0.947} & \textbf{0.048}     \\ %
        \cmidrule(lr){2-12}
        
        & \multirow{7}{*}{\thead{\normalsize \textit{HCP}~\cite{Essen2012TheHC} \\\vspace*{-0.2cm}~\\  \normalsize (808/87)}}
        & {\texttt{SAMSEG}~\cite{Cerri2020ACM}} & - & - & - & - & 0.810 & - & - & - &  -   \\ 
        & & {\texttt{FastSurfer}~\cite{Henschel2019FastSurferA}} & - & - & - & - &  0.819 & - & - & - &  -   \\ 
        & & {\texttt{SynthSR}~\cite{Iglesias2023SynthSRAP}} & 0.033 & 22.13 & 0.854 & 0.901 & - & 22.21~/~20.83~/~19.93 & 0.848~/~0.802~/~0.747 & 0.899~/~0.864~/~0.789 &  -   \\ %
        & & {\texttt{SynthSR}-FT~~} & 0.025 & 27.11 & 0.935 & 0.917 & 0.810 & 25.90~/~25.45~/~25.10 & 0.889~/~0.872~/~0.831 & 0.919~/~0.911~/~0.889 & 0.052    \\ %
        & & {\texttt{CIFL}~\cite{chua2023contrast}} & 0.029 & 26.42 & 0.932 & 0.913 & 0.804 & 25.98~/~25.11~/~23.99 & 0.930~/~0.883~/~0.871 & 0.936~/~0.927~/~0.880 & 0.059   \\ %
        & & {\texttt{SCRATCH}} & 0.023 & 27.32 & 0.923 & 0.909 & 0.792 & 26.69~/~26.41~/~25.05 & 0.932~/~0.896~/~0.848 & 0.942~/~0.930~/~\textbf{0.892} & 0.069  \\ %
        &  & {\texttt{Brain-ID}} & \textbf{0.020} & \textbf{27.47} & \textbf{0.957} & \textbf{0.929} &  \textbf{0.843} & \textbf{29.70}~/~\textbf{27.90}~/~\textbf{26.12} & \textbf{0.957}~/~\textbf{0.901}~/~\textbf{0.883}  & \textbf{0.973}~/~\textbf{0.938}~/~0.890 & \textbf{0.047}   \\ %
        \cmidrule(lr){2-12}
        
        & \multirow{7}{*}{\thead{\normalsize \textit{ADNI3}~\cite{Weiner2017TheAD} \\ \vspace*{-0.2cm}~\\ \normalsize (298/33)}}
        & {\texttt{SAMSEG}~\cite{Cerri2020ACM}} & - & - & - & - & 0.769  & - & - & - &  -   \\ 
        & & {\texttt{FastSurfer}~\cite{Henschel2019FastSurferA}} & - & - & - & - &  0.796 & - & - & - &  -   \\ 
        & & {\texttt{SynthSR}~\cite{Iglesias2023SynthSRAP}} & 0.023 & 23.60 & 0.928 & 0.909 & - & 23.13~/~22.40~/~22.27 & 0.921~/~0.907~/~0.876 & 0.903~/~0.892~/~0.871 &  -   \\ %
        & & {\texttt{SynthSR}-FT~~} & 0.021 & 27.11 & 0.950 & 0.969 & 0.820 & 27.95~/~27.34~/~26.18 & 0.920~/~0.913~/~0.890 & 0.922~/~0.890~/~0.879 & 0.134    \\ %
        & & {\texttt{CIFL}~\cite{chua2023contrast}} & 0.028 & 28.52 & 0.961 & 0.970 & 0.819 & 27.81~/~27.10~/~26.32 & 0.900~/~0.877~/~0.864 & 0.812~/~0.793~/~0.769 & 0.102    \\ %
        & & {\texttt{SCRATCH}} & 0.033 & 27.28 & 0.957 & 0.963 & 0.816 & 27.99~/~27.37~/~26.63 & 0.893~/~0.889~/~0.876 & 0.794~/~0.785~/~0.753 & 0.128   \\ %
        & & {\texttt{Brain-ID}} & \textbf{0.021} & \textbf{29.89} & \textbf{0.966} & \textbf{0.976} & \textbf{0.843} & \textbf{30.17}~/~\textbf{28.23}~/~\textbf{26.89} & \textbf{0.962}~/~\textbf{0.923}~/~\textbf{0.892} & \textbf{0.973}~/~\textbf{0.921}~/~\textbf{0.883}  & \textbf{0.093}    \\ %
        \cmidrule(lr){2-12}
        
        & \multirow{7}{*}{\thead{\normalsize \textit{ADHD200}~\cite{Brown2012ADHD200GC} \\ \vspace*{-0.2cm}~\\ \normalsize (865/96)}}
        & {\texttt{SAMSEG}~\cite{Cerri2020ACM}} & - & - & - & - & 0.784  & - & - & - &  -  \\ 
        & & {\texttt{FastSurfer}~\cite{Henschel2019FastSurferA}} & - & - & - & - &  0.801 & - & - & - &  -   \\ 
        & & {\texttt{SynthSR}~\cite{Iglesias2023SynthSRAP}} & 0.035 & 21.67 & 0.882 & 0.853 & - & 21.42~/~21.13~/~21.41 & 0.876~/~0.846~/~0.809 & 0.851~/~0.831~/~0.805 &  -   \\ 
        & & {\texttt{SynthSR}-FT~~} & 0.016 & 26.90 & 0.917 & 0.905 & 0.812 & 28.55~/~27.43~/~25.94 & 0.880~/~0.859~/~0.823 & 0.860~/~0.845~/~0.811 & 0.110   \\ %
        & & {\texttt{CIFL}~\cite{chua2023contrast}} & 0.013 & 28.69 & 0.932 & 0.921 & 0.810 & 28.11~/~26.12~/~25.11 & 0.865~/~0.848~/~\textbf{0.831} & 0.828~/~0.800~/~0.789 &  0.109   \\ %
        & & {\texttt{SCRATCH}} & \textbf{0.011} & 31.55 & 0.986 & 0.985 & 0.796 & 29.41~/~27.89~/~26.58 & 0.858~/~0.847~/~0.814 & 0.853~/~0.845~/~\textbf{0.827}  &  0.113   \\ %
        & & {\texttt{Brain-ID}}  & \textbf{0.011} & \textbf{32.48} & \textbf{0.996} & \textbf{0.996} & \textbf{0.847} & \textbf{29.50}~/~\textbf{28.56}~/~\textbf{26.87} & \textbf{0.898}~/~\textbf{0.862}~/~0.811 & \textbf{0.887}~/~\textbf{0.850}~/~0.823 & \textbf{0.107}   \\ 
        \cmidrule(lr){2-12}
        
        & \multirow{7}{*}{\thead{\normalsize \textit{AIBL}~\cite{Fowler2021FifteenYO}\\ \vspace*{-0.2cm}~\\ \normalsize (601/67)}}
        & {\texttt{SAMSEG}~\cite{Cerri2020ACM}} & - & - & - & - & 0.799  & - & - & - &  -   \\ 
        & & {\texttt{FastSurfer}~\cite{Henschel2019FastSurferA}} & - & - & - & - &  0.802 & - & - & - &  -   \\ 
        & & {\texttt{SynthSR}~\cite{Iglesias2023SynthSRAP}} & 0.026 & 22.95 & 0.916 & 0.912 & - & 22.57~/~21.82~/~21.78 & 0.917~/~0.892~/~0.856 & 0.906~/~0.893~/~0.869 &  -   \\ %
        & & {\texttt{SynthSR}-FT~~} & 0.014 & 29.89 & 0.941 & 0.922 & 0.810 & 29.15~/~27.89~/~26.77 & 0.932~/~0.906~/~0.878 & 0.927~/~0.900~/~0.871 & 0.118   \\ %
        & & {\texttt{CIFL}~\cite{chua2023contrast}} & 0.011 & 30.12 & 0.938 & 0.925 & 0.816 & 29.46~/~27.97~/~26.73 & 0.907~/~0.889~/~0.870 & 0.891~/~0.872~/~0.857 & 0.094   \\ %
        & & {\texttt{SCRATCH}} & 0.011 & 30.53 & 0.932 & 0.913 & 0.814 & 29.43~/~28.10~/~26.82 & 0.900~/~0.886~/~0.863 & 0.884~/~0.870~/~0.853 & 0.128   \\ %
        & & {\texttt{Brain-ID}} & \textbf{0.009} & \textbf{33.73} & \textbf{0.972} & \textbf{0.963} &  \textbf{0.851} & \textbf{29.75}~/~\textbf{28.58}~/~\textbf{27.09} & \textbf{0.957}~/~\textbf{0.922}~/~\textbf{0.890} & \textbf{0.975}~/~\textbf{0.956}~/~\textbf{0.937} & \textbf{0.088}  \\ %
        \hline\\[-2.3ex]
        \hline\\[-2.3ex]


        \multicolumn{1}{c}{\multirow{12}{*}{\thead{\normalsize{MR}\\ \vspace*{-0.2cm}~\\ \normalsize{(T2w)}}}} & 
        \multirow{6}{*}{\thead{\normalsize \textit{HCP}~\cite{Essen2012TheHC} \\ \vspace*{-0.2cm}~\\ \normalsize (808/87)}}
        & {\texttt{SAMSEG}~\cite{Cerri2020ACM}} & - & - & - & - &  0.782 & - & - & - &  -   \\ 
        & & {\texttt{SynthSR}~\cite{Iglesias2023SynthSRAP}} & 0.034 & 21.46 & 0.833 & 0.885 & - & - & - & - &  -   \\ %
        & & {\texttt{SynthSR}-FT~~} & 0.028 & 26.10 & 0.894 & 0.898 & 0.755 & 28.43~/~25.36~/~22.10 & 0.942~/~0.851~/~0.801 & 0.977~/~0.892~/~0.798 & 0.140 \\ %
        & & {\texttt{CIFL}~\cite{chua2023contrast}} & 0.036 & 25.12 & 0.891 & 0.879 & 0.787 & 28.26~/~25.78~/~21.73 & \textbf{0.959}~/~0.893~/~0.797 & 0.982~/~0.906~/~0.742 & 0.138    \\ %
        &  & {\texttt{SCRATCH}} & 0.038 & 24.99 & 0.873 & 0.866 & 0.756 & 28.12~/~24.52~/~21.58 & 0.945~/~0.863~/~0.783 & 0.980~/~0.893~/~0.738 & \textbf{0.136}   \\ %
        & & {\texttt{Brain-ID}} & \textbf{0.016} & \textbf{28.10} & \textbf{0.934} & \textbf{0.935} &  \textbf{0.805} & \textbf{30.26}~/~\textbf{26.11}~/~\textbf{24.10}  & \textbf{0.959}~/~\textbf{0.902}~/~\textbf{0.832} & \textbf{0.987}~/~\textbf{0.953}~/~\textbf{0.912} & \textbf{0.136} \\ %
        \cmidrule(lr){2-12}
        
        & \multirow{6}{*}{\thead{\normalsize \textit{AIBL}~\cite{Fowler2021FifteenYO} \\ \vspace*{-0.2cm}~\\ \normalsize (272/30)}}
        & {\texttt{SAMSEG}~\cite{Cerri2020ACM}} & - & - & - & - & 0.763  & - & - & - &  -   \\ 
        & & {\texttt{SynthSR}~\cite{Iglesias2023SynthSRAP}} & 0.033 & 20.08 & 0.805 & 0.831 & - & - & - & - &  -   \\ %
        & & {\texttt{SynthSR}-FT~~} & 0.024 & 22.93 & 0.815 & 0.840 & 0.719 & 30.15~/~28.09~/~26.92 & 0.940~/~0.912~/~0.893 & 0.966~/~0.948~/~0.922 & 0.200  \\ %
        & & {\texttt{CIFL}~\cite{chua2023contrast}} & 0.022 & 23.71 & 0.820 & 0.839 & 0.721 & 29.98~/~27.01~/~25.66 & 0.931~/~0.899~/~0.861 & 0.967~/~0.941~/~0.905 & 0.193   \\ %
        & & {\texttt{SCRATCH}} & \textbf{0.020} & 22.27 & 0.849 & 0.851 & 0.714 & 31.91~/~29.20~/~\textbf{27.35} & 0.954~/~0.934~/~0.896 & 0.982~/~0.969~/~0.947 &  0.197  \\ %
        & & {\texttt{Brain-ID}} & 0.022 & \textbf{23.99} & \textbf{0.861} & \textbf{0.850} & \textbf{0.782} & \textbf{32.26}~/~\textbf{29.90}~/~27.09  & \textbf{0.973}~/~\textbf{0.937}~/~\textbf{0.900} & \textbf{0.988}~/~\textbf{0.971}~/~\textbf{0.948} & \textbf{0.148} \\ %
        \hline\\[-2.3ex]
        \hline\\[-2.3ex]


        \multicolumn{1}{c}{\multirow{12}{*}{\thead{\normalsize{MR}\\ \vspace*{-0.2cm}~\\ \normalsize{(FLAIR)}}}} & 
        \multirow{6}{*}{\thead{\normalsize \textit{ADNI3}~\cite{Weiner2017TheAD} \\ \vspace*{-0.2cm}~\\ \normalsize (298/33)}}
        & {\texttt{SAMSEG}~\cite{Cerri2020ACM}} & - & - & - & - & 0.718  & - & - & - &  -   \\ 
        & & {\texttt{SynthSR}~\cite{Iglesias2023SynthSRAP}} & 0.026 & 22.77 & 0.919 & 0.895 & - & - & - & - &  -   \\ %
        & & {\texttt{SynthSR}-FT~~} & 0.021 & 22.34 & 0.921 & \textbf{0.900} & 0.753 & 30.11~/~28.70~/~25.32 & 0.930~/~0.899~/~0.864 & 0.948~/~0.910~/~0.878 & 0.251   \\ %
        & & {\texttt{CIFL}~\cite{chua2023contrast}} & 0.020 & 21.29 & 0.911 & 0.897 & 0.761 & \textbf{32.72}~/~29.00~/~26.99 & 0.949~/~0.906~/~0.873 & 0.953~/~0.919~/~0.889 & 0.237  \\ %
        & & {\texttt{SCRATCH}} & 0.025 & 20.80 & 0.901 & 0.862 & 0.759 & 32.36~/~28.71~/~28.00 & 0.945~/~0.915~/~0.877 & 0.941~/~0.936~/~0.917 & 0.236   \\ %
        & & {\texttt{Brain-ID}} & \textbf{0.017} & \textbf{26.44} & \textbf{0.927} & 0.892 & \textbf{0.786} & 32.18~/~\textbf{30.00}~/~\textbf{28.28} & \textbf{0.959}~/~\textbf{0.921}~/~\textbf{0.883} & \textbf{0.982}~/~\textbf{0.965}~/~\textbf{0.921} & \textbf{0.227} \\ %
        \cmidrule(lr){2-12}
        
        & \multirow{6}{*}{\thead{\normalsize \textit{AIBL}~\cite{Fowler2021FifteenYO} \\ \vspace*{-0.2cm}~\\ \normalsize (302/34)}}
        & {\texttt{SAMSEG}~\cite{Cerri2020ACM}} & - & - & - & - & 0.710  & - & - & - &  -   \\ 
        & & {\texttt{SynthSR}~\cite{Iglesias2023SynthSRAP}} & 0.029 & 21.77 & 0.902 & 0.892 & - & - & - & - &  -   \\ %
        & & {\texttt{SynthSR}-FT~~} & 0.024 & 25.80 & 0.914 & 0.900 & 0.735 & 26.20~/~24.91~/~23.17 & 0.860~/~0.832~/~0.795 & 0.933~/~0.910~/~0.894 & 0.126  \\ %
        & & {\texttt{CIFL}~\cite{chua2023contrast}} & 0.026 & 27.11 & 0.901 & 0.870 & 0.721 & 27.72~/~26.85~/~25.99 & 0.851~/~0.828~/~0.804 & 0.939~/~0.905~/~0.901 & 0.109  \\ %
        & & {\texttt{SCRATCH}} & 0.023 & 26.84 & 0.898 & 0.867 & 0.720 & 28.02~/~26.52~/~25.43 & 0.878~/~0.838~/~0.799 & 0.943~/~0.901~/~0.891 & 0.115 \\ %
        & & {\texttt{Brain-ID}} & \textbf{0.019} & \textbf{27.25} & \textbf{0.936} & \textbf{0.912} & \textbf{0.767} & \textbf{28.69}~/~\textbf{27.63}~/~\textbf{26.69} & \textbf{0.949}~/~\textbf{0.906}~/~\textbf{0.866} & \textbf{0.971}~/~\textbf{0.925}~/~\textbf{0.916} & \textbf{0.103} \\ %
        \hline\\[-2.3ex]
        \hline\\[-2.3ex]

        
        \multicolumn{1}{c}{\multirow{6}{*}{CT}} 
        & \multirow{6}{*}{\thead{\normalsize \textit{OASIS3}~\cite{LaMontagne2018OASIS3LN} \\ \vspace*{-0.2cm}~\\ \normalsize  (795/88)}}
        & {\texttt{SAMSEG}~\cite{Cerri2020ACM}} & - & - & - & - & 0.701  & - & - & - &  -   \\ 
        & & {\texttt{SynthSR}~\cite{Iglesias2023SynthSRAP}} & 0.041 & 20.93 & 0.758 & 0.786 & - & - & - & - &  -   \\ %
        & & {\texttt{SynthSR}-FT~~} & 0.030 & 23.76 & 0.797 & 0.845 & 0.700 & 25.30~/~22.18~/~20.35 & 0.892~/~0.813~/~0.760 & 0.896~/~0.801~/~0.764 & -   \\ %
        & & {\texttt{CIFL}~\cite{chua2023contrast}} & 0.027 & 24.91 & 0.819 & 0.871 & 0.718 & 25.99~/~23.70~/~22.83 & 0.909~/~\textbf{0.820}~/~0.779 & 0.969~/~0.816~/~0.775 & -  \\ %
        & & {\texttt{SCRATCH}} & 0.025 & 24.35 & 0.811 & 0.872 & 0.709 & 26.34~/~23.64~/~22.62 & 0.905~/~0.818~/~0.770 & 0.873~/~0.818~/~0.780 & -  \\ %
        & & {\texttt{Brain-ID}} & \textbf{0.023} & \textbf{25.49} & \textbf{0.891} & \textbf{0.895} & \textbf{0.765} & \textbf{26.74}~/~\textbf{24.01}~/~\textbf{23.09} & \textbf{0.910}~/~0.818~/~\textbf{0.792} & \textbf{0.974}~/~\textbf{0.824}~/~\textbf{0.799} & - \\ 
    
\midrule[\heavyrulewidth] \\ [-3.2ex]
\multicolumn{12}{l}{\small (1) ``-'' means not applicable: \texttt{FastSurfer}~\cite{Henschel2019FastSurferA} and \texttt{SAMSEG}~\cite{Cerri2020ACM} are for segmentation-only; \texttt{SynthSR}~\cite{Iglesias2023SynthSRAP} only synthesize T1w MRI; CT does not have bias fields.} \\ [-0.5ex]
\multicolumn{12}{l}{\small (2) \texttt{SynthSR}-FT: \texttt{SynthSR} fine-tuned on each task/dataset, for a fairer comparison.} \\ [-0.5ex] 

\bottomrule  \\ [-3.6ex]  
    \end{tabular}   
}

\end{table}

%% file: sec/exp/contrast_inde.tex
\vspace{-0.3cm}
\subsubsection{Contrast-independent Applications}
\label{exp: independent} 

\noindent \vspace{0.1cm}\\ \textbf{- Anatomy Reconstruction/Contrast Synthesis}~As one of the most common types of MRI scans, T1w MR images highlight the differences between gray and white matter, and are mostly often utilized to image the anatomy of the brain, spinal cord, bones and joints~\cite{Symms2004ARO}. For each dataset, we train models to reconstruct their T1w MRI counterparts with the \texttt{L1} loss.

\noindent  \textbf{- Brain Segmentation}~ For each dataset, we use SynthSeg~\cite{Billot2021SynthSegSO} to obtain the segmentation labels with 30 brain anatomical regions~\cite{Fischl2002WholeBS}, as the gold standard segmentation target. We train our compared models to predict the brain segmentation labels, with the soft dice loss and cross-entropy loss~\cite{Billot2021SynthSegSO}.

As shown in \cref{fig: task}, \texttt{Brain-ID} clearly outperforms \texttt{CIFL} and \texttt{SCRATCH} in the reconstruction task. As highlighted by the {\color{mint!200}mint} circles, it reveals the anatomy details and produces more fine-grained results than the strong state-of-the-art method \texttt{SynthSR}~\cite{Iglesias2023SynthSRAP}. \texttt{Brain-ID} also achieves better segmentation results, especially in the smaller and more challenging regions. Quantitative comparisons are in \cref{tab: task}, where \texttt{Brain-ID} obtains the best scores across most metrics. Notably, \texttt{Brain-ID} achieves greater gains over baseline models, \texttt{SCRATCH} and \texttt{CIFL}, particularly on smaller datasets (e.g., ADNI3-T1/FLAIR, AIBL-FLAIR, where \texttt{Brain-ID}'s robust and rich features quickly adapt to specific tasks/datasets.

%% file: sec/exp/fig/fig_comp.tex
\begin{figure}[t]

\centering 

\resizebox{\linewidth}{!}{
	\begin{tikzpicture}
        
		\tikzstyle{myarrows}=[line width=0.6mm,draw=blue!50,-triangle 45,postaction={draw, line width=0.05mm, shorten >=0.02mm, -}]
		\tikzstyle{mylines}=[line width=0.8mm]
  

	\pgfmathsetmacro{\shift}{-1.}
 
        \node at (\shift+1.05, 3+0.35,0.1*3) {\hypertarget{fig: feat_dist-a}{~}};
 
	\foreach \i in {-1, 1, 3}
	{
        \draw [<->, color = matcha!150, line width = 0.5mm](\shift+0.6,\i-0.15,0.1*3) -- (\shift+1.5,\i-0.15,0.1*3);  
	}

        
        \node at (0.+\shift+0.5-1.9, 3.2+0.15*3, 0.75*3) {T1w};
        \node at (0.+\shift+0.5-1.9, 2.8+0.15*3, 0.75*3) {(HF)};
        \node at (0.+\shift+0.5-1.9, 1.2+0.15*3, 0.75*3) {T1w};
        \node at (0.+\shift+0.5-1.9, 0.8+0.15*3, 0.75*3) {(\textbf{\textit{LF}})}; 
        \node at (0.+\shift+0.5-1.9, -0.8+0.15*3, 0.75*3) {\textbf{\textit{FLAIR}}};
        \node at (0.+\shift+0.5-1.9, -1.2+0.15*3, 0.75*3) {(HF)};

	\node at (\shift+0.1*3, 3+0.15*3, 0.75*3) {\includegraphics[width=0.16\textwidth]{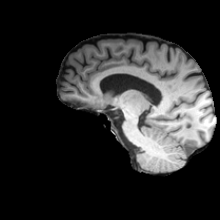}};
	\node at (\shift+0.1*3, 1+0.15*3, 0.75*3) {\includegraphics[width=0.16\textwidth]{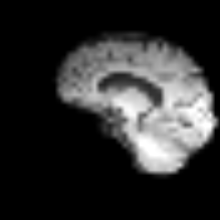}};
	\node at (\shift+0.1*3, -1+0.15*3, 0.75*3) {\includegraphics[width=0.16\textwidth]{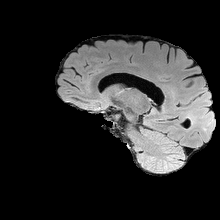}};

 
 
	\pgfmathsetmacro{\shift}{-3.2}

	\node at (5.3+\shift+0.1*3, 3+0.15*3, 0.75*3) {\includegraphics[width=0.16\textwidth]{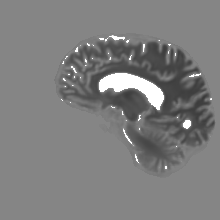}};
	\node at (5.3+\shift+0.1*3, 1+0.15*3, 0.75*3) {\includegraphics[width=0.16\textwidth]{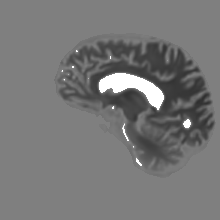}};
	\node at (5.3+\shift+0.1*3, -1+0.15*3, 0.75*3) {\includegraphics[width=0.16\textwidth]{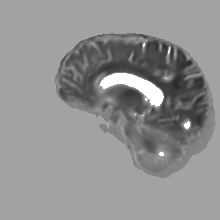}};
        \node at (5.3+\shift+0.1*3-0.5, 3-0.45+0.15*3, 0.75*3) {\textcolor{green}{\LARGE\cmark}};
        \node at (5.3+\shift+0.1*3-0.5, 1-0.45+0.15*3, 0.75*3) {\textcolor{green}{\LARGE\cmark}};
        \node at (5.3+\shift+0.1*3-0.5, -1-0.45+0.15*3, 0.75*3) {\textcolor{green}{\LARGE\cmark}};
        
	\node at (7.3+\shift+0.1*3, 3+0.15*3, 0.75*3) {\includegraphics[width=0.16\textwidth]{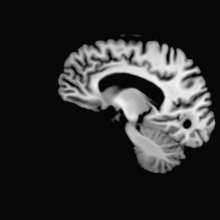}};
	\node at (7.3+\shift+0.1*3, 1+0.15*3, 0.75*3) {\includegraphics[width=0.16\textwidth]{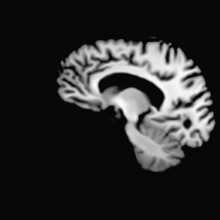}};
	\node at (7.3+\shift+0.1*3, -1+0.15*3, 0.75*3) {\includegraphics[width=0.16\textwidth]{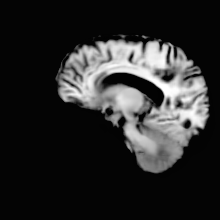}};
        \node at (7.3+\shift+0.1*3-0.5, 3-0.45+0.15*3, 0.75*3) {\textcolor{green}{\LARGE\cmark}};
        \node at (7.3+\shift+0.1*3-0.5, 1-0.45+0.15*3, 0.75*3) {\textcolor{green}{\LARGE\cmark}};
        \node at (7.3+\shift+0.1*3-0.5, -1-0.45+0.15*3, 0.75*3) {\textcolor{green}{\LARGE\cmark}};
 
	\node at (9.8+\shift+0.1*3, 3+0.15*3, 0.75*3) {\includegraphics[width=0.16\textwidth]{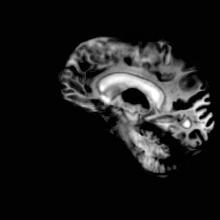}};
	\node at (9.8+\shift+0.1*3, 1+0.15*3, 0.75*3) {\includegraphics[width=0.16\textwidth]{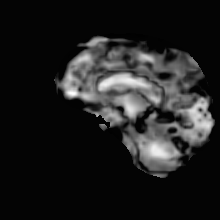}};
	\node at (9.8+\shift+0.1*3, -1+0.15*3, 0.75*3) {\includegraphics[width=0.16\textwidth]{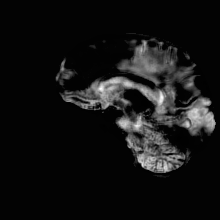}};
        \node at (9.8+\shift+0.1*3-0.5, 3-0.45+0.15*3, 0.75*3) {\textcolor{green}{\LARGE\cmark}};
        \node at (9.8+\shift+0.1*3-0.5, 1-0.45+0.15*3, 0.75*3) {\textcolor{red}{\LARGE\xmark}};
        \node at (9.8+\shift+0.1*3-0.5, -1-0.45+0.15*3, 0.75*3) {\textcolor{red}{\LARGE\xmark}};
 
	\node at (11.8+\shift+0.1*3, 3+0.15*3, 0.75*3) {\includegraphics[width=0.16\textwidth]{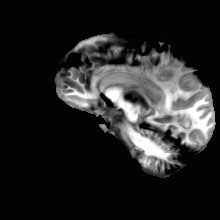}};
	\node at (11.8+\shift+0.1*3, 1+0.15*3, 0.75*3) {\includegraphics[width=0.16\textwidth]{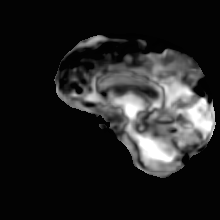}};
	\node at (11.8+\shift+0.1*3, -1+0.15*3, 0.75*3) {\includegraphics[width=0.16\textwidth]{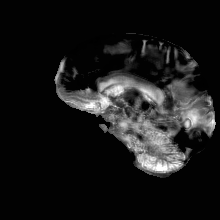}};
        \node at (11.8+\shift+0.1*3-0.5, 3-0.45+0.15*3, 0.75*3) {\textcolor{green}{\LARGE\cmark}};
        \node at (11.8+\shift+0.1*3-0.5, 1-0.45+0.15*3, 0.75*3) {\textcolor{red}{\LARGE\xmark}};
        \node at (11.8+\shift+0.1*3-0.5, -1-0.45+0.15*3, 0.75*3) {\textcolor{red}{\LARGE\xmark}};

    \pgfmathsetmacro{\dx}{-7.65}
    \pgfmathsetmacro{\dy}{5.28}
        \node[anchor=north] at (6.1+\dx, -7.65+\dy) {{\textbf{Input}}}; 
    \pgfmathsetmacro{\dy}{4.9}
        \node[anchor=north] at (6.1+\dx, -7.6+\dy) {(\textbf{HF}: high-field; \textbf{LF}: low-field)};

    \pgfmathsetmacro{\dx}{-3.5}
    \pgfmathsetmacro{\dy}{4.9}
        \draw [decorate,decoration={brace,amplitude=5pt,mirror,raise=6ex},line width=2.pt,color = matcha!150] (3.9+\dx, -1.6) -- (8.2+\dx, -1.6);
        \node[anchor=north] at (6.1+\dx, -7.6+\dy) {\textcolor{matcha!150}{\textbf{\texttt{Brain-ID} features}}};

    \pgfmathsetmacro{\dx}{1.}
        \draw [decorate,decoration={brace,amplitude=5pt,mirror,raise=6ex},line width=2.pt,color = gray!80] (3.9+\dx, -1.6) -- (8.2+\dx, -1.6);
        \node[anchor=north] at (6.1+\dx, -7.6+\dy) {\textcolor{gray!80}{\textbf{\texttt{SCRATCH} features}}}; 


	\end{tikzpicture}
	}  
	\vspace*{-0.6cm}
\caption{\texttt{SCRATCH}, which is well trained on HF T1w scans, produces highly descriptive features for HF T1w images ($1^{\text{st}}$ row), but \textit{does not preserve} the same high quality useful for downstream tasks when handling LF ($2^{\text{nd}}$ row) or other contrasts ($3^{\text{rd}}$ row).} 
	 \label{fig: feat_comp}
  
\end{figure}
  

%% file: sec/exp/contrast_de.tex
\vspace{-0.35cm}
\subsubsection{Contrast-dependent Applications}
\label{exp: dependent} 
We also evaluate \texttt{Brain-ID} on two brain imaging tasks that are dependent on the input modality/quality.

\vspace{0.12cm}
\noindent \textbf{- Image Super-resolution}~We use the standard 1$mm$ isotropic images from all datasets as the ground truth high-resolution target, and follow the strategy in SynthSR~\cite{Iglesias2023SynthSRAP} where the input samples are randomly resampled and corrupted during training. For inference, we downsample the original images into 3$mm$, 5$mm$, and 7$mm$ isotropic images as inputs, and evaluate the output quality.

\vspace{0.12cm}
\noindent \textbf{- Bias Field Estimation}~The bias field is a smooth,  low-frequency multiplicative signal that corrupts MRI images, which affects image analysis tasks such as segmentation or texture analysis~\cite{Juntu2005BiasFC}. Bias field estimation is often needed as a pre-processing step to correct corrupted MRI images~\cite{Chua2009EvaluationOP}. We apply randomly simulated bias fields to the input samples~(\cref{sec: corrupt}), and train all models with the \texttt{L2} loss. During inference, we pre-generate and apply the bias fields on the testing data for reproducibility, and evaluate the bias field estimation performance.

\vspace{0.1cm}
With a simple one-layer adaption~(\cref{sec: dependent}), \texttt{Brain-ID}'s contrast-agnostic anatomical features achieve state-of-the-art performance on contrast-dependent tasks~(\cref{tab: task}, \cref{fig: task}). As shown in Fig.~\ref{fig: feat_comp}, \texttt{Brain-ID} obtains contrast/resolution-robustness that \textit{cannot} be achieved by models trained from real images (due to the \textit{limited variability} in their appearance), regardless of the backbone choices. 
Such robust features greatly improve super-resolution, resulting in higher (\texttt{MS}-)\texttt{SSIM} scores. \texttt{Brain-ID}'s gains are less obvious for bias field estimation, probably due to the fact that the bias field is approximately \textit{independent} of brain anatomy.

%% file: sec/exp/additional.tex
\subsection{Practical Value and Broader Impact}
\label{exp: dataset_size}
\vspace{-0.05cm}

\input{sec/exp/fig/comp_curve}
\subsubsection{Low-resolution data} Trained on synthetic images, \texttt{Brain-ID} features are not only robust to various contrasts/modalities, but also extremely robust to low-resolution data (Fig.~\ref{fig: feat_comp}) and provide huge potential in clinical MRI (including big retrospective data~\cite{Billot2023Robust}) and portable low-field MRI~\cite{Iglesias2022Quantitative}.

\vspace{-0.4cm}
\subsubsection{Small-size Datasets}
To investigate the effectiveness of \texttt{Brain-ID} features for limited-size datasets, we assess its performance across all four downstream tasks on subsets of ADNI3~\cite{Weiner2017TheAD} training set (298 cases originally). As shown in \cref{fig: comp_curve}, we reduce the training set size percentage from 100\% to an extreme 10\%. With only 20\% of the data ({\textcolor{mint!200}{mint}} line), \texttt{Brain-ID} still achieves better performance compared to its full-sized baseline, ``100\%'' \texttt{SCRATCH} ({\textcolor{myyellow!200}{yellow}} line), over all tasks. 30 training samples (10\%, {\textcolor{myblue!200}{blue}} line) may be at the edge of obtaining acceptable results for \texttt{Brain-ID}, as the model becomes less stable and the performance drops occur more frequently -- especially for brain segmentation (\cref{fig: comp_curve}-(b)).


\vspace{-0.16cm}
\subsection{Additional Experiments and Further Discussions}
\label{exp: ablation}
\vspace{-0.1cm}

\input{sec/exp/fig/tab_ablat}


\subsubsection{Choice of Anatomy Guidance}~We explore other anatomical learning targets than MP-RAGE (\cref{parag: supervision}): \textit{(i)}~segmentation labels; \textit{(ii)}~both segmentation labels and MP-RAGE. As shown in \cref{tab: ablat}, adding segmentation guidance improves robustness, yet inhibits the feature expressiveness and affects the downstream performance; we observe features obtained from segmentation guidance produce less high-frequency texture than \texttt{Brain-ID}, especially \textit{within} the same-label regions. (Please refer to the visualizations in Fig.~\ref{app-fig: feat_comp}, \cref{app: ablat}.)

\vspace{-0.5cm}
\subsubsection{Data Generation Design}~As shown in \cref{tab: ablat}, training with all mildly-corrupted samples results in reduced feature robustness and further harms the downstream performance, yet using all extremely corrupted samples leads to unstable training and model collapse. We train \texttt{Brain-ID} on samples of gradually increased corruptions (\cref{fw: train} (left)). (\cref{app: implement} lists the full parameter setups.)

\vspace{-0.5cm}
\subsubsection{Batch Size}~\texttt{Brain-ID} computes training loss of all intra-subject samples in a mini-batch at once (\cref{eq: Brain-loss}). We observe that larger batches help improve feature robustness~(\cref{tab: ablat}), yet do not further enhance the downstream performance.

\vspace{-0.5cm}
\subsubsection{Limitations}~\texttt{Brain-ID} is designed to capture the unique anatomy of subjects, it works exceptionally well on fundamental brain imaging tasks such as reconstruction, segmentation, and super-resolution, with a simple one-layer adaption. However, as \texttt{Brain-ID}'s data generation is solely based on anatomies, we observe that it is not adept at handling images with extensive pathology regions. Our future work will focus on contrast-agnostic, pathology-encoded representations.

%% file: sec/exp/fig/comp_curve.tex
\begin{figure}[t]

\centering 

\resizebox{0.9\linewidth}{!}{
	\begin{tikzpicture}
        \hspace*{-0.5cm}
        
		\tikzstyle{myarrows}=[line width=0.2mm,draw=blue!50,-triangle 45,postaction={draw, line width=0.05mm, shorten >=0.02mm, -}]
		\tikzstyle{mylines}=[line width=0.3mm]
  

 
	\pgfmathsetmacro{\shift}{-3.2}

	\node at (5.2+\shift+0.1*3, 5.8+0.15*3, 0.75*3) {\includegraphics[width=0.50\textwidth]{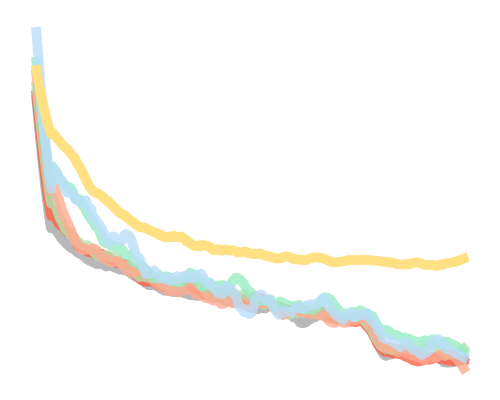}};
	\node at (5.2+\shift+0.1*3, -1+0.15*3, 0.75*3) {\includegraphics[width=0.50\textwidth]{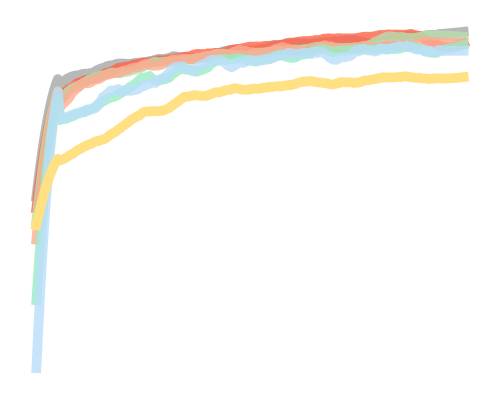}};
        
	\node at (12+\shift+0.1*3, 5.8+0.15*3, 0.75*3) {\includegraphics[width=0.50\textwidth]{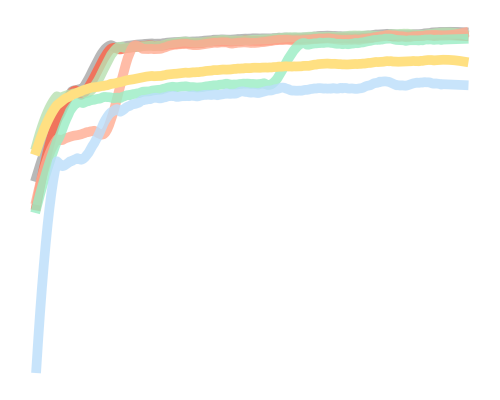}};
	\node at (12+\shift+0.1*3, -1+0.15*3, 0.75*3) {\includegraphics[width=0.50\textwidth]{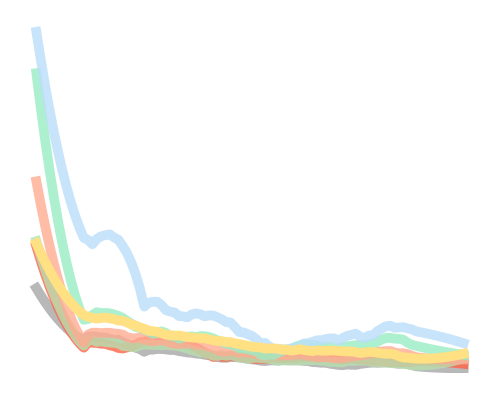}};


    \pgfmathsetmacro{\dx}{0}
    \pgfmathsetmacro{\dy}{0}
    
    \pgfmathsetmacro{\x}{-1.3}
    \pgfmathsetmacro{\y}{-3.7}
    
    \node[anchor=north] at (2.75+\x+\dx, -0.8+6.8+\y+\dy) {(a) Reconstruction}; 
    \node[anchor=north] at (2.75+6.8+\x+\dx, -0.8+6.8+\y+\dy) {(b) Segmentation}; 
    \node[anchor=north] at (2.75+\x+\dx, -0.8+\y+\dy) {(c) Super-resolution}; 
    \node[anchor=north] at (2.75+6.8+\x+\dx, -0.8+\y+\dy) {(d) Bias Field Estimation}; 

    \node[anchor=north] at (\x+\dx, 5.2+6.8+\y+\dy) {{\texttt{L1} ($\downarrow$)}};
    \node[anchor=north] at (6.8+\x+\dx, 5.2+6.8+\y+\dy) {{\texttt{Dice} ($\uparrow$)}};
    \node[anchor=north] at (\x+\dx, 5.2+\y+\dy) {{\texttt{SSIM} ($\uparrow$)}}; 
    \node[anchor=north] at (6.8+\x+\dx, 5.2+\y+\dy) {{\texttt{norm-L2} ($\downarrow$)}};  


    \pgfmathsetmacro{\a}{7.4}
    \pgfmathsetmacro{\b}{-3.1}
    \draw [mylines, dashed, color = black](6.8+\x+\dx, -0.1+\b+\dy) -- (5.7+6.8+\x+\dx, -0.1+\b+\dy);
    \draw [mylines, dashed, color = black](6.8+\x+\dx, -0.3+\b+\dy) -- (5.7+6.8+\x+\dx, -0.3+\b+\dy);
    \node at (6.8-0.5+\x+\dx, \b+\dy) {\small{0.128}};
    \node at (6.8-0.5+\x+\dx, -0.4+\b+\dy) {\bf\small{0.093}};

    \pgfmathsetmacro{\dx}{-3.3}
    \pgfmathsetmacro{\dy}{-4.45}
    \pgfmathsetmacro{\ddx}{2.5}
    \pgfmathsetmacro{\ddy}{1.5}
    
    \node at (-6.5+\a+\x+\dx, 2.1+6.8+1.5*\ddy+0.3+0.08+\dy) {(\% of ADNI3~\cite{Weiner2017TheAD})};

    \fill[fill=myyellow!110] (-7.5+\a+\x+\dx, 1.8+6.8+\ddy+\dy) rectangle ++(2.0, 0.16);
    \node at (-6.5+\a+\x+\dx, 2.1+6.8+\ddy+0.08+\dy) {\texttt{SCRATCH}};  
    \node at (-6.5+\a+\x+\dx, 2.1+6.8+\ddy+0.3+0.08+\dy) {100\%}; 
    
    \fill[fill=gray!80] (-7.5+\a+\x+\dx, 1.8+6.8+\dy) rectangle ++(2.0, 0.16);
    \node at (-6.5+\a+\x+\dx, 2.1+6.8+0.08+\dy) {\texttt{{Brain-ID}}}; 
    \node at (-6.5+\a+\x+\dx, 2.1+6.8+0.3+0.08+\dy) {100\%}; 
    
    \fill[fill=tomato!80] (-7.5+\a+\x+\dx, 1.8+6.8-\ddy+\dy) rectangle ++(2.0, 0.16);
    \node at (-6.5+\a+\x+\dx, 2.1+6.8-\ddy+0.08+\dy) {\texttt{{Brain-ID}}}; 
    \node at (-6.5+\a+\x+\dx, 2.1+6.8-\ddy+0.3+0.08+\dy) {80\%}; 
    
    \fill[fill=matcha] (-7.5+\a+\x+\dx, 1.8+6.8-2*\ddy+\dy) rectangle ++(2.0, 0.16);
    \node at (-6.5+\a+\x+\dx, 2.1+6.8-2*\ddy+0.08+\dy) {\texttt{{Brain-ID}}}; 
    \node at (-6.5+\a+\x+\dx, 2.1+6.8-2*\ddy+0.3+0.08+\dy) {60\%}; 
    
    \fill[fill=myorange] (-7.5+\a+\x+\dx, 1.8+6.8-3*\ddy+\dy) rectangle ++(2.0, 0.16);
    \node at (-6.5+\a+\x+\dx, 2.1+6.8-3*\ddy+0.08+\dy) {\texttt{{Brain-ID}}}; 
    \node at (-6.5+\a+\x+\dx, 2.1+6.8-3*\ddy+0.3+0.08+\dy) {40\%}; 
    
    \fill[fill=mint] (-7.5+\a+\x+\dx, 1.8+6.8-4*\ddy+\dy) rectangle ++(2.0, 0.16);
    \node at (-6.5+\a+\x+\dx, 2.1+6.8-4*\ddy+0.08+\dy) {\texttt{{Brain-ID}}}; 
    \node at (-6.5+\a+\x+\dx, 2.1+6.8-4*\ddy+0.3+0.08+\dy) {20\%}; 

    \fill[fill=myblue] (-7.5+\a+\x+\dx, 1.8+6.8-5*\ddy+\dy) rectangle ++(2.0, 0.16);
    \node at (-6.5+\a+\x+\dx, 2.1+6.8-5*\ddy+0.08+\dy) {\texttt{{Brain-ID}}}; 
    \node at (-6.5+\a+\x+\dx, 2.1+6.8-5*\ddy+0.3+0.08+\dy) {10\%};

    \pgfmathsetmacro{\dx}{0}
    \pgfmathsetmacro{\dy}{0}
    
    \pgfmathsetmacro{\a}{0.9}
    \pgfmathsetmacro{\b}{0.}
    \draw [mylines, dashed, color = black](6.8+\x+\dx, 6.42+0.3+\b+\dy) -- (5.7+6.8+\x+\dx, 6.42+0.3+\b+\dy);
    \draw [mylines, dashed, color = black](6.8+\x+\dx, 6.41+0.95+\b+\dy) -- (5.7+6.8+\x+\dx, 6.41+0.95+\b+\dy);
    \node at (6.8-0.5+\x+\dx, 6.42+0.3+\b+\dy) {\small{0.816}};
    \node at (6.8-0.5+\x+\dx, 6.41+0.95+\b+\dy) {\bf\small{0.843}}; 
    
    \pgfmathsetmacro{\a}{7.4}
    \pgfmathsetmacro{\b}{-3.1}
    \draw [mylines, dashed, color = black](\x+\dx, 0.92+6.8+\b+\dy) -- (5.7+\x+\dx, 0.92+6.8+\b+\dy);
    \draw [mylines, dashed, color = black](\x+\dx, -0.2+6.8+\b+\dy) -- (5.7+\x+\dx, -0.2+6.8+\b+\dy);
    \node at (-0.5+\x+\dx, 0.92+6.8+\b+\dy) {\small{0.043}}; 

    \pgfmathsetmacro{\a}{0.9}
    \pgfmathsetmacro{\b}{-3.1}
    \draw [mylines, dashed, color = black](\x+\dx, 3.1+\b+\dy) -- (5.7+\x+\dx, 3.1+\b+\dy);
    \draw [mylines, dashed, color = black](\x+\dx, 3.6+\b+\dy) -- (5.7+\x+\dx, 3.6+\b+\dy);
    \node at (-0.5+\x+\dx, 3.1+\b+\dy) {\small{0.893}};
    \node at (-0.5+\x+\dx, 3.6+\b+\dy) {\bf\small{0.962}};


	\foreach \dx/\dy in {0/0, 6.8/0, 0/6.8, 6.8/6.8}
	{
    \draw [myarrows, color = black](\x+\dx, \y+\dy) -- (5.7+\x+\dx, \y+\dy);  
    \draw [myarrows, color = black](\x+\dx, \y+\dy) -- (\x+\dx, 4.7+\y+\dy); 
    
    \node[anchor=north] at (5.2+\x+\dx, -0.3+\y+\dy) {{\textit{Epoch}}}; 
    
    \draw [mylines, color = black](1+\x+\dx, \y+\dy) -- (1+\x+\dx, 0.1+\y+\dy); 
    \node[] at (1+\x+\dx, -0.2\y+\dy) {\small{50}};  
    \draw [mylines, color = black](3+\x+\dx, \y+\dy) -- (3+\x+\dx, 0.1+\y+\dy);  
    \node[] at (3+\x+\dx, -0.2\y+\dy) {\small{150}};   
    \draw [mylines, color = black](5+\x+\dx, \y+\dy) -- (5+\x+\dx, 0.1+\y+\dy); 
    \node[] at (5+\x+\dx, -0.2\y+\dy) {\small{250}};   
    
}

	\end{tikzpicture}
	}  
	\vspace*{-0.25cm}
\caption{Adapting \texttt{Brain-ID} to small datasets. The horizontal (vertical) axes indicate training epochs (evaluation scores of corresponding tasks). Results are obtained by evaluating models collected throughout the epochs, on ADNI3~\cite{Weiner2017TheAD} full testing set.}  
 
	 \label{fig: comp_curve}
  
\end{figure}
  

%% file: sec/exp/fig/tab_ablat.tex
\begin{table}[t]
    \caption{Comparison between \texttt{Brain-ID} and its variants.} 
    \vspace*{-0.25cm}
    \label{tab: ablat}
\resizebox{0.75\linewidth}{!}{
\centering 
    \begin{tabular}{lccccc}
       \toprule \\[-3ex] 
      \multicolumn{1}{c}{\multirow{2}{*}{\thead{\normalsize\textbf{Model Setup}\\\textit{\normalsize[Comparison Target]}}}} & \multicolumn{2}{c}{\textbf{Feature Robustness (Intra)}} & \multicolumn{3}{c}{\textbf{Downstream (Reconstruction)}} \\ [-0.7ex]
        \cmidrule(lr){2-3}
        \cmidrule(lr){4-6}
           & \texttt{SSIM} ($\uparrow$) & \texttt{MS-SSIM} ($\uparrow$) & \texttt{L1} ($\downarrow$) & \texttt{PSNR} ($\uparrow$) & \texttt{SSIM} ($\uparrow$) \\ [-0.2ex]
     \midrule\\[-3ex]
     
        \multicolumn{5}{l}{\textit{[Representation guidance: supervision target for feature learning]}} &  \\ 
        {Segmentation} &  \textbf{0.891} & \textbf{0.963} & 0.029 & 28.13 & 0.958 \\ 
        {Segmentation $+$ MP-RAGE} & 0.863 & 0.940 & 0.023 & 29.05 & 0.964 \\
        {MP-RAGE (\texttt{Brain-ID})} & 0.858 & 0.921 & \textbf{0.021} & \textbf{29.89} & \textbf{0.966} \\ 
         \hline\\[-2.3ex]
         
         \multicolumn{5}{l}{\textit{[Data generation design: corruption levels in intra-subject mini-batches]}} & \\ 
        {All Mild ($\sigma_{\text{noise}}=1,\, \cdots$)} & 0.792 & 0.813 & 0.028 & 28.30 & 0.960 \\ 
        {All Medium ($\sigma_{\text{noise}}=5,\, \cdots$)} & 0.831 & 0.899 & 0.022 & 29.67 & 0.964 \\ 
        {All Severe ($\sigma_{\text{noise}}=10,\, \cdots$)} & N/A & N/A & N/A & N/A & N/A \\ 
        {Mild to Severe (\texttt{Brain-ID})} & \textbf{0.858} & \textbf{0.921} & \textbf{0.021} & \textbf{29.89} & \textbf{0.966} \\ 
         \hline\\[-2.3ex]
         
        \multicolumn{5}{l}{\textit{[Mini-batch size: number of intra-subject samples]}} & \\ 
        2 & 0.826 & 0.897 & 0.025 & 28.83 & 0.959 \\ 
        3 & 0.841 & 0.902 & 0.022 & 29.30 & 0.962 \\ 
        {4~(\texttt{Brain-ID})} & 0.858 & 0.921 & \textbf{0.021} & 29.89 & \textbf{0.966} \\ 
        5 & \textbf{0.860} & \textbf{0.929} & \textbf{0.021} & \textbf{29.93} & 0.965 \\ [-0.5ex]
\bottomrule  \\ [-3.6ex]  
    \end{tabular} 
}
\end{table}

%% file: sec/con.tex
\vspace{-0.1cm}
\section{Conclusion}
\label{sec: con}
\vspace{-0.15cm}
We introduced \texttt{Brain-ID}, a contrast-agnostic anatomical representation learning approach for brain imaging, which is distinctive to each subject's anatomy and resilient to variations in image appearances. \texttt{Brain-ID} is trained entirely on synthetic data, and quickly and effectively adapts to downstream tasks through only a single layer. We validated \texttt{Brain-ID} features' intra/inter-subject robustness, and effectiveness on four downstream applications (reconstruction, segmentation, super-resolution, bias field estimation). Experiments on six public datasets, covering MR and CT, demonstrated that \texttt{Brain-ID} achieves state-of-the-art performance across all tasks and modalities, and preserves high performance on low-resolution data and small-size datasets. We believe \texttt{Brain-ID} unlocks the great potential of robust foundation models, especially for non-calibrated modalities.

%% file: sec/appendix/main.tex
\vspace{1cm}
{
\begin{center} 
\textbf{\LARGE Appendix for \texttt{Brain-ID}}
\end{center}
}

\setcounter{table}{0}
\setcounter{figure}{0}
\setcounter{equation}{0}

\renewcommand\thetable{\thesection.\arabic{table}}
\renewcommand\thefigure{\thesection.\arabic{figure}}
\renewcommand\theequation{\thesection.\arabic{equation}}

\appendix

\input{sec/appendix/dataset}
\input{sec/appendix/implement}
\input{sec/appendix/feat}

\input{sec/appendix/task}

\input{sec/appendix/additional}

%% file: sec/appendix/dataset.tex
\section{Datasets and Metrics}
\label{app: dataset}

\subsubsection{Datasets}
We test and compare our method over various datasets including modalities of MR and CT, the MR images further contain T1-weighted , T2-weighted and FLAIR (fluid-attenuated inversion recovery) images. 

\begin{itemize}
\item \texttt{ADNI}~\cite{Jack2008TheAD}: we use T1-weighted (2045 cases) MRI scans from the Alzheimer’s Disease Neuroimaging Initiative (ADNI). All scans are acquired at 1 $mm$ isotropic resolution from a wide array of scanners and protocols. The dataset contains aging subjects, some diagnosed with mild cognitive impairment (MCI) or Alzheimer's Disease (AD). Many subjects present strong atrophy patterns and white matter lesions.

\item \texttt{HCP}~\cite{Essen2012TheHC}: we use T1-weighted (897 cases) and T2-weighted (897 cases) MRI scans of young subjects from the Human Connectome Project, acquired at 0.7 mm resolution.

\item \texttt{ADNI3}~\cite{Weiner2017TheAD}: we use T1-weighted (331 cases) and FLAIR (331 cases) MRI scans from ADNI3, which continues the previously funded ADNI1, ADNI-GO, and ADNI2 studies to determine the relationships between the clinical, cognitive, imaging, genetic and biochemical biomarker characteristics of the entire spectrum of sporadic late onset AD.

\item \texttt{ADHD200}~\cite{Brown2012ADHD200GC}: we use T1-weighted (961 cases) MRI scans from ADH200 Sample, which is a grassroots initiative dedicated to the understanding of the neural basis of Attention Deficit Hyperactivity Disorder (ADHD).

\item \texttt{AIBL}~\cite{Fowler2021FifteenYO}: we use T1-weighted (668 cases), T2-weighted (302 cases) and FLAIR (336 cases) MRI scans from The Australian Imaging, Biomarkers and Lifestyle (AIBL) Study, which is a study of cognitive impairment (MCI) and Alzheimer’s disease dementia.

\item \texttt{OASIS3}~\cite{LaMontagne2018OASIS3LN}: we use CT (885 cases) scans from OASIS3, which is a longitudinal neuroimaging, clinical, and cognitive dataset for normal aging and AD. For our experiments, we use CT and T1-weighted MRI pair with the earliest date, from each subject.

\end{itemize}

\subsubsection{Data for Synthetic Generator}

\texttt{Brain-ID}'s synthetic generator uses (1) brain segmentation labels, for random-contrast input images generation~(\cref{sec: generator}), and (2) MP-RAGE, the target ground truth for anatomy-guided supervision~(\cref{sec: framework}). In this work, we use the segmentation maps of training images from ADNI~\cite{Jack2008TheAD}, as well as their corresponding MP-RAGE images. Note that we do not use any type of real images from ADNI as input for \texttt{Brain-ID}'s pre-training.

\subsubsection{Data Preprocessing}

For all datasets, we skull-strip all the images using SynthStrip~\cite{Hoopes2022SynthStripSF}, and resample them to 1 $mm$ isotropic resolution. For all the images, except T1-weighted MRI, in each dataset, we use NiftyReg~\cite{Modat2010FastFD} rigid registration to register all images to their same-subject T1-weighted MRI counterparts. The gold-standard brain segmentation maps are obtained by performing SynthSeg~\cite{Billot2021SynthSegSO} on the T1-weighted MR images of all the subjects.

\subsubsection{Metrics}
We resort to various metrics for evaluating individual tasks across multiple aspects: 

\begin{itemize}
\item \texttt{L1}: the average $L1$ distance, is used for intra/inter-subject feature distance evaluation (\cref{exp: feat}), and the overall prediction correctness of anatomy reconstruction (\cref{exp: independent}), super-resolution and bias-field estimation (\cref{exp: dependent}).

\item \texttt{normL2}: the normalized $L2$ distance for bias field~\cite{Chua2009EvaluationOP} is defined as:
\begin{equation}
    \text{\texttt{normL2}} = \sqrt{ \frac{\sum_x\big(w B_{\text{est}}(x) - B_{\text{true}}(x)\big)^2 }{\sum_x B_{\text{true}}^2(x)} }\,,
    \label{app-eq: norml2}
\end{equation}
where $w$ is the normalization coefficient obtained by:
\begin{equation}
    w = \frac{ \sum_x B_{\text{true}}(x) B_{\text{est}}(x)}{ \sum_x B_{\text{est}}^2 (x) }\,, \quad x \in \Omega\,,
\end{equation}
$\Omega$ refers to the brain domain, $B_{\text{est}}$ and $B_{\text{true}}$ are the estimated and ground truth bias fields, respectively. Normalization is necessary for the evaluation of bias field estimation (\cref{exp: dependent}) because nonuniformity correction may result in arbitrary scaling of the bias field.

\item \texttt{PSNR}: the peak signal-to-noise ratio (PSNR) that indicates the fidelity of predictions. It is used in anatomy reconstruction (\cref{exp: independent}) and super-resolution (\cref{exp: dependent}).

\item \texttt{(MS-)SSIM}: the structural similarity scores between the generated and real images. \texttt{MS-SSIM} is a variant of \texttt{SSIM} focusing on multiple scales of the images that are shown to correlate well with human perception~\cite{Wang2003MultiscaleSS,Wang_2021_CVPR,Liu2021SelfappearanceaidedDE}. They are used in intra/inter-subject feature distance evaluation (\cref{exp: feat}), reconstruction (\cref{exp: independent}), and super-resolution (\cref{exp: dependent}),

\item \texttt{Dice}: the similarity score between predicted and ground truth segmentations, and it is used in brain segmentation evaluation  (\cref{exp: independent}).

\end{itemize}

%% file: sec/appendix/implement.tex
\section{Implementation Details}
\label{app: implement}


\subsubsection{Model Architecture}
As mentioned in~\cref{sec: exp}, \texttt{Brain-ID} can use any backbone to extract brain features. We use the five-level 3D UNet~\cite{Ronneberger2015UNetCN} as \texttt{Brain-ID}'s backbone for feature extraction, with 64 feature channels in the last layer.
\begin{itemize}
\item During the feature pre-training stage~(\cref{sec: framework}), a linear regression layer is added following the feature outputs for anatomy supervision (\cref{eq: Brain-loss}). 

\item During downstream tasks adaptions, the regression layer for anatomy supervision is abandoned, instead, a task-specific activation layer is added following the feature outputs (\cref{sec: downstream}). Specifically, a linear regression layer is added for the tasks, anatomy reconstruction/contrast synthesis, image super-resolution, and bias field estimation. An additional softmax activation is added for segmentation probability outputs.
\end{itemize}


\subsubsection{Feature Backbone Pre-training}
We pre-train \texttt{Brain-ID} on the synthetic data from our generator (\cref{sec: generator}) for 300,000 iterations, with a patch size of $128^3$ and a mini-batch size (i.e., number of intra-subject augmented samples) of 4. We use the synchronized AdamW optimization, with a base learning rate of $10^{-4}$ and a linear warm-up in the first 2,000 iterations followed by a multi-step learning schedule (learning rate drops at 160,000 and 240,000 iterations) with a multiplier of 0.1. 


\subsubsection{Synthetic Data Generator}

\input{sec/appendix/fig/tab_setup}

As shown in \cref{fw: train}, \texttt{Brain-ID} simulates its training samples of increasing corruption levels, from mild to severe. \cref{tab: ablat} also explores the effects of different levels of sample corruption on feature robustness and downstream performance. In \cref{app-tab: setup}, we list the generator parameters for mild, medium, and severe data corruption levels, respectively. Note that (1) for each level, the setup parameters only control the corruption value ranges, since the simulation is randomized, there could still be mildly corrupted samples generated under the ``severe'' settings; (2) The random deformation fields are independent of data corruption levels.

%% file: sec/appendix/fig/tab_setup.tex
\begin{table}[t]
\resizebox{0.6\linewidth}{!}{
\centering 
    \begin{tabular}{lcccc}
       \toprule \\[-3ex] 
        \multirow{2}{*}{\textbf{Category}} & \multirow{2}{*}{\textbf{Param}} & \multicolumn{3}{c}{\textbf{Corruption Level}} \\ [-0.2ex]
        \cmidrule(lr){3-5}
        & & Mild & Medium & Severe \\ [-0.2ex]
        \midrule\\[-3ex]
     
        \multirow{6}{*}{\thead{Deformation}} & affine-rotation$_{max}$ & 15 & $=$ & $=$ \\ 
        & affine-shearing$_{max}$ & 0.2 & $=$ & $=$ \\ 
        & affine-scaling$_{max}$ & 0.2 & $=$ & $=$ \\
        & nonlinear-scale $\mu_{min}$ & 0.03 & $=$ & $=$ \\
        & nonlinear-scale $\mu_{max}$ & 0.06 & $=$ & $=$ \\
        & nonlinear-scale $\sigma_{max}$ & 4 & $=$ & $=$ \\
         \hline\\[-2.3ex]

        \multirow{2}{*}{\thead{Resolution}} & $p_{\text{low-field}}$ & 0.1 & 0.3 & 0.5 \\ 
        & $p_{\text{anisotropic}}$ & 0 & 0.1 & 0.25 \\
         \hline\\[-2.3ex]

        \multirow{4}{*}{\thead{Bias Field}} & $\mu_{min}$ & 0.01 & 0.02 & 0.02 \\ 
        & $\mu_{max}$ & 0.02 & 0.03 & 0.04 \\
        & $\sigma_{min}$ & 0.01 & 0.05 & 0.1 \\ 
        & $\sigma_{max}$ & 0.05 & 0.3 & 0.6 \\
         \hline\\[-2.3ex]

        \multirow{2}{*}{\thead{Noises}} & $\sigma_{min}$ & 0.01 & 0.5 & 5 \\ 
        & $\sigma_{max}$ & 1 & 5 & 15 \\
        
\bottomrule \\[-3.5ex]
    \end{tabular}  
    \caption{\texttt{Brain-ID} synthetic generator setups: mild, medium and severe levels. $p$ denotes probability, $\mu$ and $\sigma$ refer to the mean and variance of the Gaussian distributions, respectively.} 
    \label{app-tab: setup}
}
\end{table}

%% file: sec/appendix/feat.tex
\section{Feature Robustness Evaluation}
\label{app: feat}

For the evaluation of feature robustness, we use T1-weighted MRI of 100 randomly selected subjects from ADNI~\cite{Jack2008TheAD}. To challenge the model's robustness against data corruptions, and meanwhile obtain comparable and reproducible results, we use our data generator (\cref{sec: generator}) to pre-augment all the input images. Note that in this section, there is no contrast simulation step within data augmentation, and only the random deformation and data corruptions are applied. For each selected subject, we generate intra-augmented samples. All samples are generated with ``medium'' corruption settings as listed in \cref{app-tab: setup}.

%% file: sec/appendix/task.tex
\section{Downstream Task Comparisons}
\label{app: task}

For all the downstream tasks, the model architecture and date generation strategy used for \texttt{Brain-ID}, \texttt{SCRATCH} and \texttt{CIFL} are the same. The only difference between the three compared models lies in their initial weights. \texttt{Brain-ID} and \texttt{CIFL} are initialized by their pre-trained weights from training on synthetic data as described in \cref{sec: generator,sec: framework}.

For the state-of-the-art comparisons, we consider \texttt{FastSurfer}~\cite{Henschel2019FastSurferA} and \texttt{SAMSEG} \cite{Puonti2016FastAS,Cerri2020ACM} for brain segmentation, and \texttt{SynthSR}~\cite{Iglesias2023SynthSRAP} for anatomy reconstruction/contrast synthesis and super-resolution for T1-weighted images.

\begin{itemize}
    \item \texttt{FastSurfer}~\cite{Henschel2019FastSurferA} is a state-of-the-art brain segmentation model, which is designed for segmentation on T1-weighted images. Therefore, we only report the segmentation performance of \texttt{FastSurfer} on T1-weighted MRI. In addition, since \texttt{FastSurfer} does not predict cerebrospinal fluid (CSF), we remove the CSF label during the \texttt{Dice} score computation.
    \item \texttt{SAMSEG}~\cite{Puonti2016FastAS,Cerri2020ACM} is a state-of-the-art, multi-modal brain segmentation model, which works on both MR and CT images. Similar to \texttt{FastSurfer}, \texttt{SAMSEG} does not predict the CSF label either, the CSF label is therefore removed during the \texttt{Dice} score computation of \texttt{SAMSEG}.
    \item \texttt{SynthSR}~\cite{Iglesias2023SynthSRAP} is a state-of-the-art, contrast-agnostic model for anatomy reconstruction/contrast synthesis. For input MRI images with any contrast and resolution, \texttt{SynthSR} generates their corresponding high-resolution, 1~$mm$ isotropic T1-weighted MRI. In our comparisons, we apply \texttt{SynthSR} on our anatomy reconstruction/contrast synthesis task, as well as the image super-resolution task of T1-weighted MRI. 
\end{itemize}

%% file: sec/appendix/additional.tex
\section{Additional Experimental Results}
\label{app: ablat}

\subsubsection{Feature Representation Learning}

\input{sec/appendix/fig/fig_feat}

As discussed in \cref{exp: ablation}, in \texttt{Brain-ID} we adopt the high-resolution MP-RAGE scan as the anatomy guidance for brain feature representation. Experimental comparisons in \cref{tab: ablat} illustrate that incorporating segmentation as the target for anatomical supervision in learning brain feature representation leads to reduced high-frequency texture compared to the use of MP-RAGE alone. The visual comparisons presented in  \cref{app-fig: feat_comp} reveal that the features with segmentation guidance indeed encompass anatomical structures, however, they exhibit notably smoother texture \textit{within} each structural region defined by the brain segmentation labels. In contrast, employing MP-RAGE as the target for anatomical supervision inherently entails the tasks of anatomy reconstruction and super-resolution simultaneously. The resulting features from \texttt{Brain-ID} are shown to carry richer information content, as evidenced by the more pronounced high-frequency textures they manifest.


\subsubsection{Downstream Task Evaluation}

\input{sec/appendix/fig/fig_task-recon}
\input{sec/appendix/fig/fig_task-sr}
\input{sec/appendix/fig/fig_task-seg}

In addition to the qualitative results in \cref{showcase,fig: task}, we provide more visualization comparisons of the downstream tasks in \cref{app-fig: task-recon,app-fig: task-sr,app-fig: task-seg}.

%% file: sec/appendix/fig/fig_feat.tex
\begin{figure*}[t]

\centering 

\resizebox{\linewidth}{!}{
	\begin{tikzpicture}
        
		\tikzstyle{myarrows}=[line width=0.6mm,draw=blue!50,-triangle 45,postaction={draw, line width=0.05mm, shorten >=0.02mm, -}]
		\tikzstyle{mylines}=[line width=0.8mm]
  


	\pgfmathsetmacro{\shift}{-1.}
 
	\foreach \i in {-1, 1, 3}
	{
        \draw [->, color = matcha!150, line width = 0.5mm](\shift+0.9,\i-0.15,0.1*3) -- (\shift+1.5,\i-0.15,0.1*3);  
	}
        \draw [-, color = matcha!150, line width = 0.5mm](\shift+0.9,-1-0.15,0.1*3) -- (\shift+0.9,3-0.15,0.1*3); 
        \draw [-, color = matcha!150, line width = 0.5mm](\shift+0.6,1-0.15,0.1*3) -- (\shift+1.05,1-0.15,0.1*3); 
 
	\foreach \i in {-8, -6, -4}
	{
        \draw [->, color = matcha!150, line width = 0.5mm](\shift+0.9,\i-0.15,0.1*3) -- (\shift+1.5,\i-0.15,0.1*3);  
	}
        \draw [-, color = matcha!150, line width = 0.5mm](\shift+0.9,-8-0.15,0.1*3) -- (\shift+0.9,-4-0.15,0.1*3); 
        \draw [-, color = matcha!150, line width = 0.5mm](\shift+0.6,-6-0.15,0.1*3) -- (\shift+1.05,-6-0.15,0.1*3);

	\node at (\shift+0.1*3, -7.3+0.15*3, 0.75*3) {Input};
	\node at (\shift+0.1*3, -6+0.15*3, 0.75*3) {\includegraphics[width=0.16\textwidth]{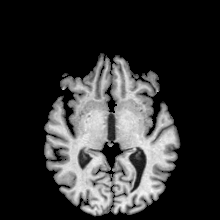}};
 
	\node at (\shift+0.1*3, -0.3+0.15*3, 0.75*3) {Input};
	\node at (\shift+0.1*3, 1+0.15*3, 0.75*3) {\includegraphics[width=0.16\textwidth]{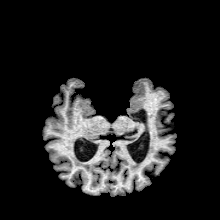}};
 
        \node at (0.+\shift+0.6-1.9, 1+0.15*3, 0.75*3) {ID-$i$};
        
        \node at (0.+\shift+0.6-1.9, -6+0.15*3, 0.75*3) {ID-$ii$};

	\foreach \i in {-0.5, -7.5}
	{
	\node at (\shift+1.95, \i+0.15*3, 0.75*3) {(c)};
	\node at (\shift+1.95, \i+2+0.15*3, 0.75*3) {(b)};
	\node at (\shift+1.95, \i+4+0.15*3, 0.75*3) {(a)};
        }

    
    \pgfmathsetmacro{\dx}{2.5}
    \pgfmathsetmacro{\dy}{-2.1}
    \draw [myarrows, color = matcha!150](-2.1+\dx, -7.45+\dy) -- (9.+\dx, -7.45+\dy);  
    \node[anchor=north] at (3.+\dx, -7.6+\dy+0.1) {\textcolor{matcha!150}{\textbf{Feature Channel}}}; 
    \pgfmathsetmacro{\dy}{4.9}
    \draw [myarrows, color = matcha!150](-2.1+\dx, -7.45+\dy) -- (9.+\dx, -7.45+\dy);  
    \node[anchor=north] at (3.+\dx, -7.6+\dy+0.1) {\textcolor{matcha!150}{\textbf{Feature Channel}}}; 

 
 
	\pgfmathsetmacro{\shift}{-3.2}

	\node at (5.2+\shift+0.1*3, 3+0.15*3, 0.75*3) {\includegraphics[width=0.16\textwidth]{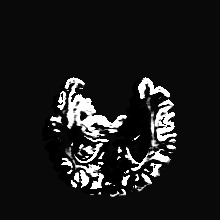}};
	\node at (5.2+\shift+0.1*3, 1+0.15*3, 0.75*3) {\includegraphics[width=0.16\textwidth]{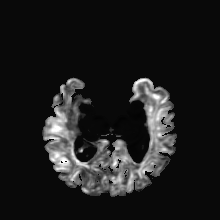}};
	\node at (5.2+\shift+0.1*3, -1+0.15*3, 0.75*3) {\includegraphics[width=0.16\textwidth]{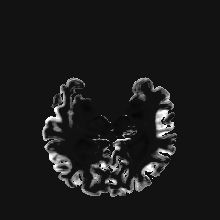}};
	\node at (5.2+\shift+0.1*3, -4+0.15*3, 0.75*3) {\includegraphics[width=0.16\textwidth]{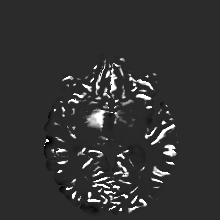}};
	\node at (5.2+\shift+0.1*3, -6+0.15*3, 0.75*3) {\includegraphics[width=0.16\textwidth]{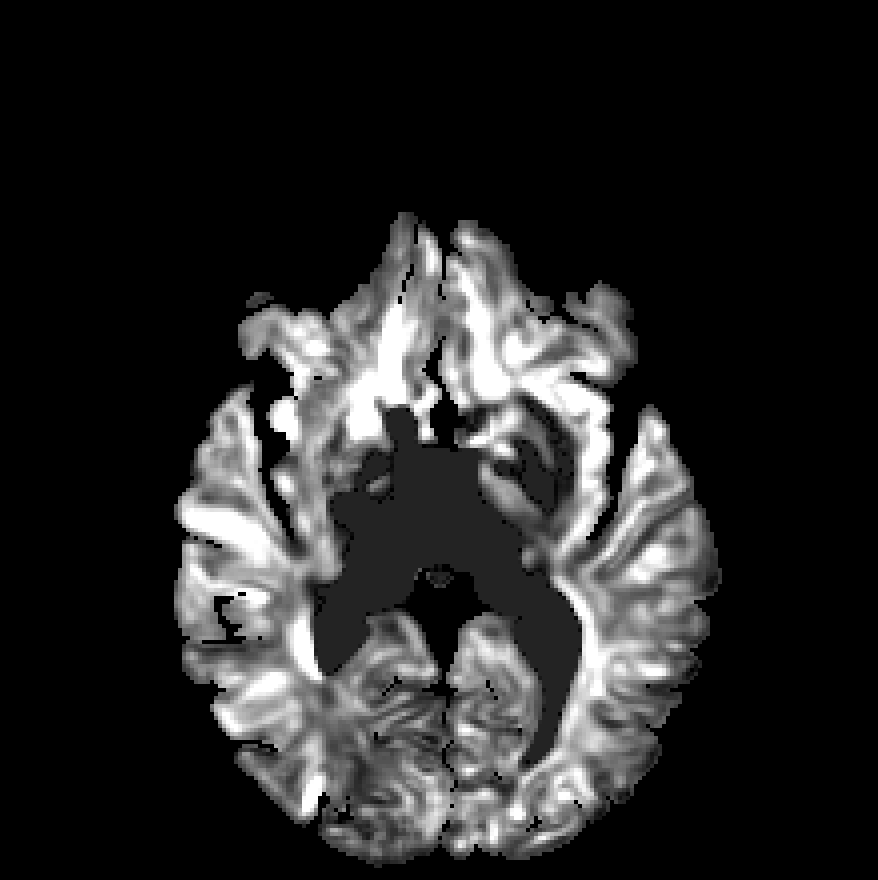}};
	\node at (5.2+\shift+0.1*3, -8+0.15*3, 0.75*3) {\includegraphics[width=0.16\textwidth]{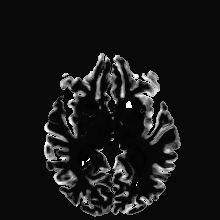}};
        
	\node at (7.2+\shift+0.1*3, 3+0.15*3, 0.75*3) {\includegraphics[width=0.16\textwidth]{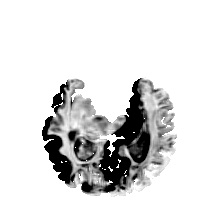}};
	\node at (7.2+\shift+0.1*3, 1+0.15*3, 0.75*3) {\includegraphics[width=0.16\textwidth]{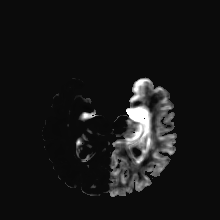}};
	\node at (7.2+\shift+0.1*3, -1+0.15*3, 0.75*3) {\includegraphics[width=0.16\textwidth]{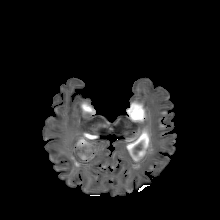}};
	\node at (7.2+\shift+0.1*3, -4+0.15*3, 0.75*3) {\includegraphics[width=0.16\textwidth]{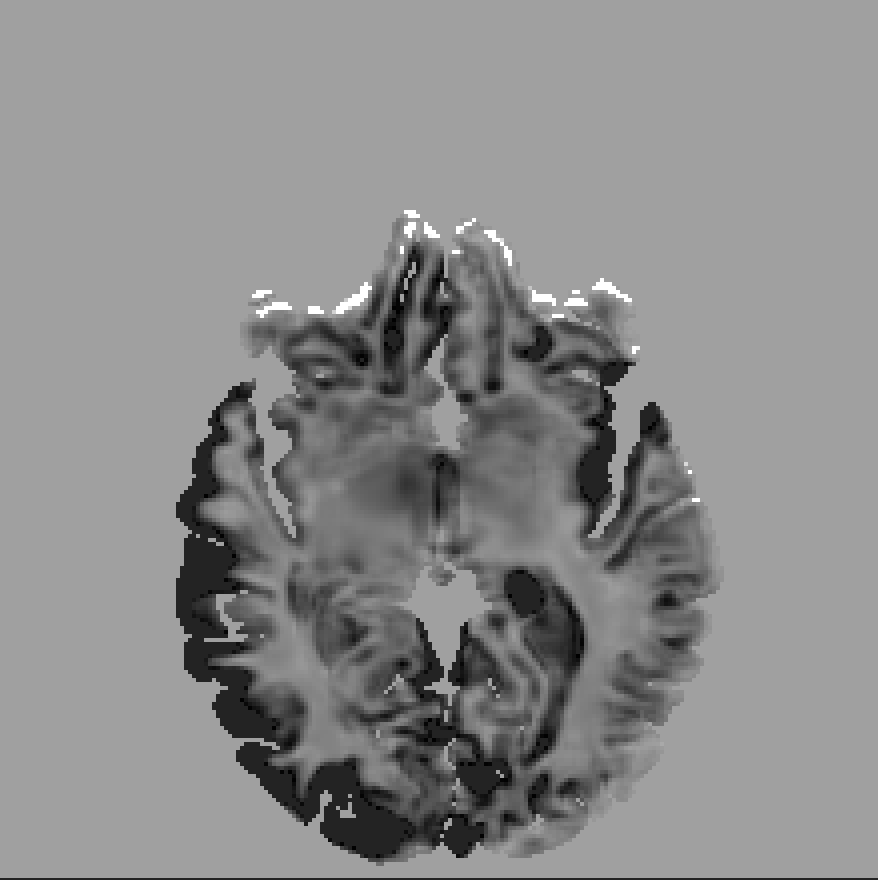}};
	\node at (7.2+\shift+0.1*3, -6+0.15*3, 0.75*3) {\includegraphics[width=0.16\textwidth]{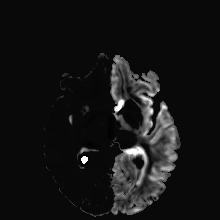}};
	\node at (7.2+\shift+0.1*3, -8+0.15*3, 0.75*3) {\includegraphics[width=0.16\textwidth]{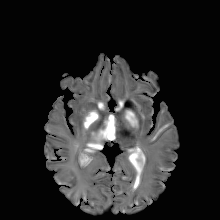}};
 
	\node at (9.2+\shift+0.1*3, 3+0.15*3, 0.75*3) {\includegraphics[width=0.16\textwidth]{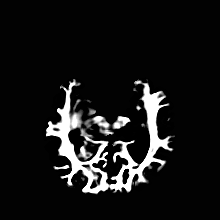}};
	\node at (9.2+\shift+0.1*3, 1+0.15*3, 0.75*3) {\includegraphics[width=0.16\textwidth]{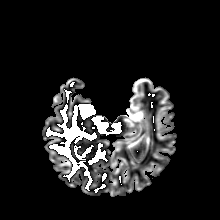}};
	\node at (9.2+\shift+0.1*3, -1+0.15*3, 0.75*3) {\includegraphics[width=0.16\textwidth]{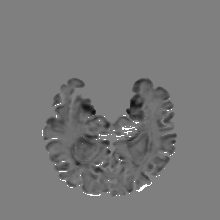}};
	\node at (9.2+\shift+0.1*3, -4+0.15*3, 0.75*3) {\includegraphics[width=0.16\textwidth]{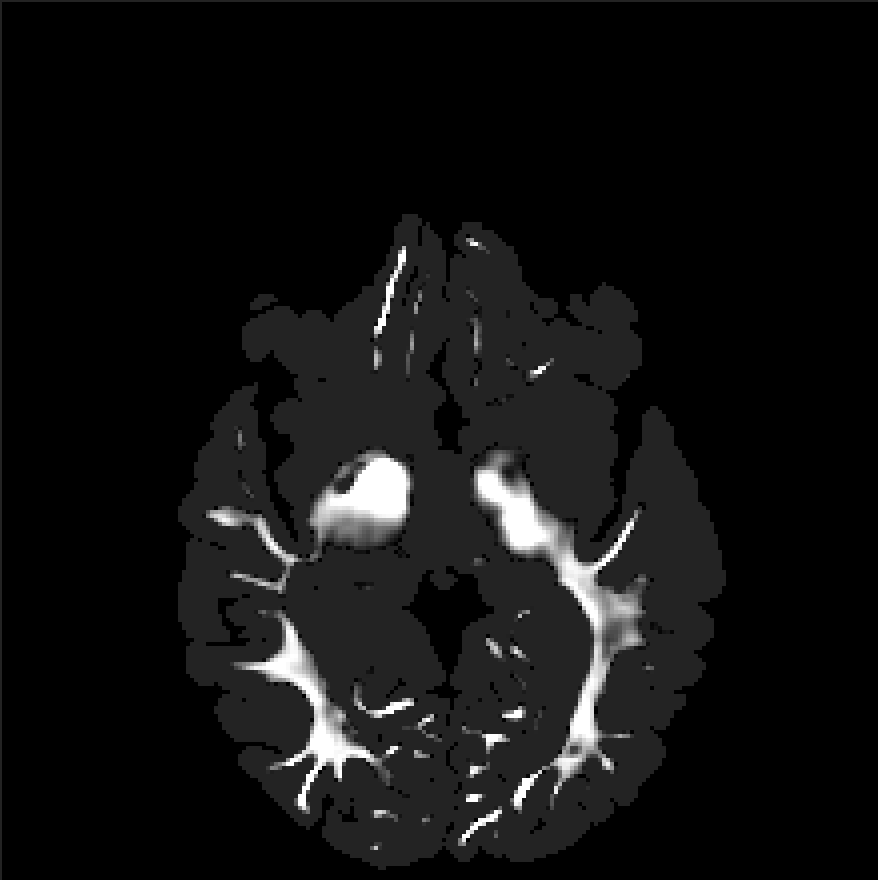}};
	\node at (9.2+\shift+0.1*3, -6+0.15*3, 0.75*3) {\includegraphics[width=0.16\textwidth]{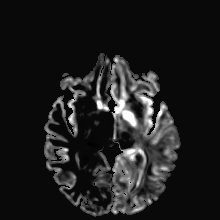}};
	\node at (9.2+\shift+0.1*3, -8+0.15*3, 0.75*3) {\includegraphics[width=0.16\textwidth]{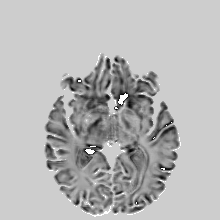}};
 
	\node at (11.2+\shift+0.1*3, 3+0.15*3, 0.75*3) {\includegraphics[width=0.16\textwidth]{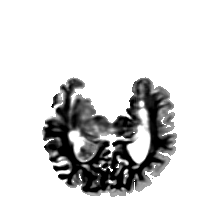}};
	\node at (11.2+\shift+0.1*3, 1+0.15*3, 0.75*3) {\includegraphics[width=0.16\textwidth]{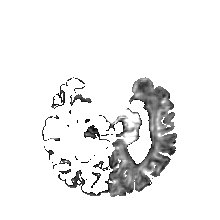}};
	\node at (11.2+\shift+0.1*3, -1+0.15*3, 0.75*3) {\includegraphics[width=0.16\textwidth]{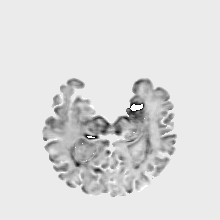}};
	\node at (11.2+\shift+0.1*3, -4+0.15*3, 0.75*3) {\includegraphics[width=0.16\textwidth]{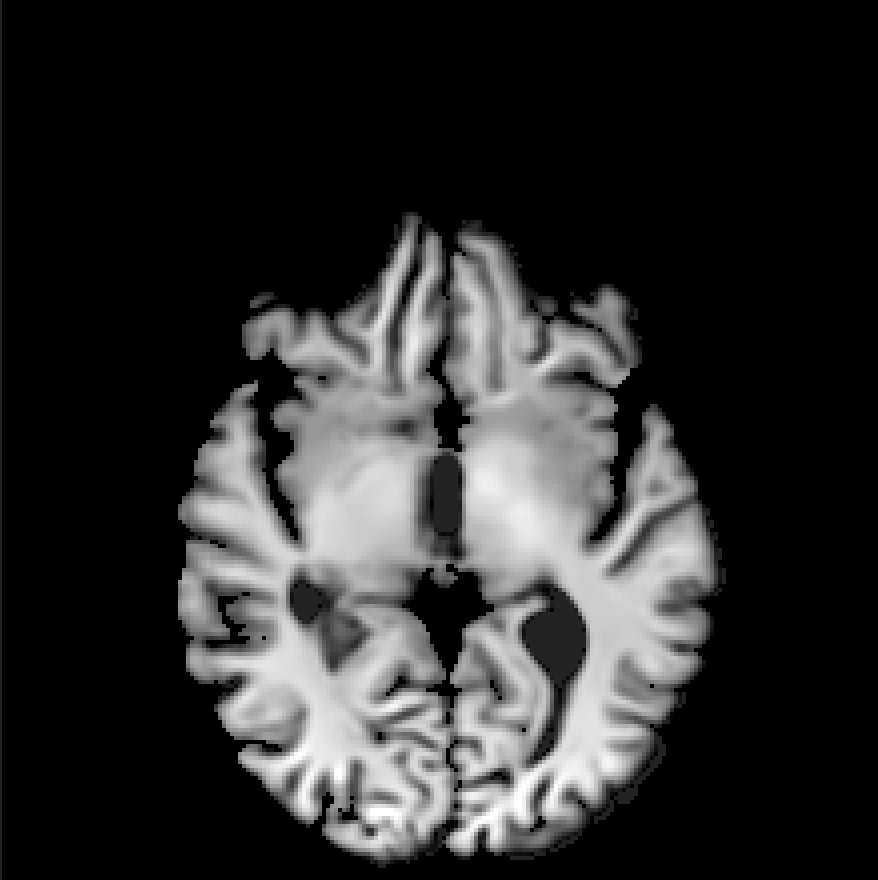}};
	\node at (11.2+\shift+0.1*3, -6+0.15*3, 0.75*3) {\includegraphics[width=0.16\textwidth]{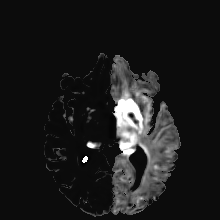}};
	\node at (11.2+\shift+0.1*3, -8+0.15*3, 0.75*3) {\includegraphics[width=0.16\textwidth]{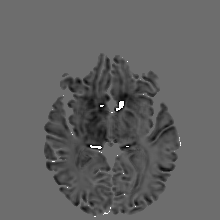}};

	\node at (13.2+\shift+0.1*3, 3+0.15*3, 0.75*3) {\includegraphics[width=0.16\textwidth]{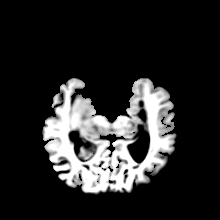}};
	\node at (13.2+\shift+0.1*3, 1+0.15*3, 0.75*3) {\includegraphics[width=0.16\textwidth]{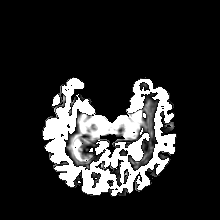}};
	\node at (13.2+\shift+0.1*3, -1+0.15*3, 0.75*3) {\includegraphics[width=0.16\textwidth]{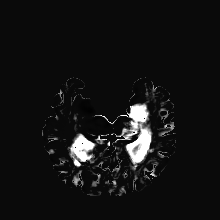}};
	\node at (13.2+\shift+0.1*3, -4+0.15*3, 0.75*3) {\includegraphics[width=0.16\textwidth]{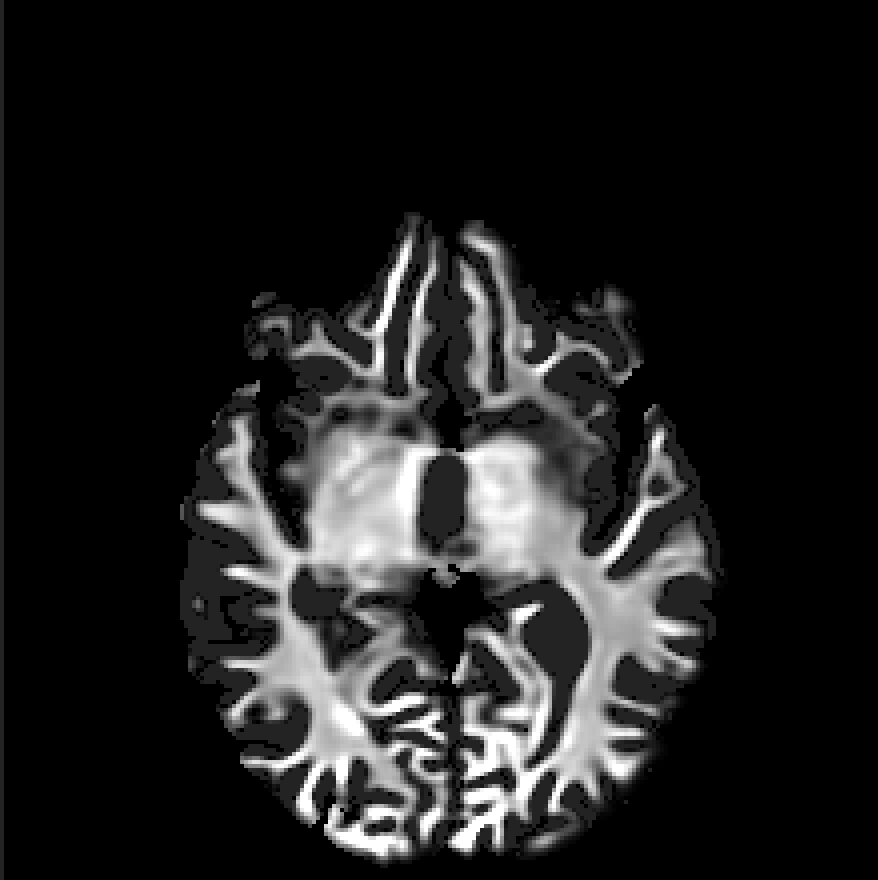}};
	\node at (13.2+\shift+0.1*3, -6+0.15*3, 0.75*3) {\includegraphics[width=0.16\textwidth]{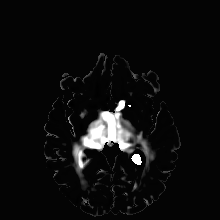}};
	\node at (13.2+\shift+0.1*3, -8+0.15*3, 0.75*3) {\includegraphics[width=0.16\textwidth]{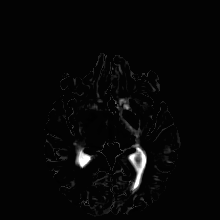}};

    \pgfmathsetmacro{\shift}{-1.2}
    \foreach \y in {-8, -6, -4, -1, 1, 3}{
    \draw[black,fill=black] (12.6+\shift+0.1*3, \y+0.15*3, 0.75*3)  circle (0.8pt);
    \draw[black,fill=black] (12.8+\shift+0.1*3, \y+0.15*3, 0.75*3)  circle (0.8pt);
    \draw[black,fill=black] (13.+\shift+0.1*3, \y+0.15*3, 0.75*3)  circle (0.8pt);
}
        

	\end{tikzpicture}
	}  
\caption{Visualization comparisons for features computed from (a) \texttt{Brain-ID} (MP-RAGE guided) with its variants: (b) segmentation guided and (c) segmentation $+$ MP-RAGE guided feature representation models. Note that although the two testing subjects here for the three models are the same, their respective selected feature channels are different, for the purpose of better showing different frequency levels of features from each model.} 
 
	 \label{app-fig: feat_comp}
  
\end{figure*}
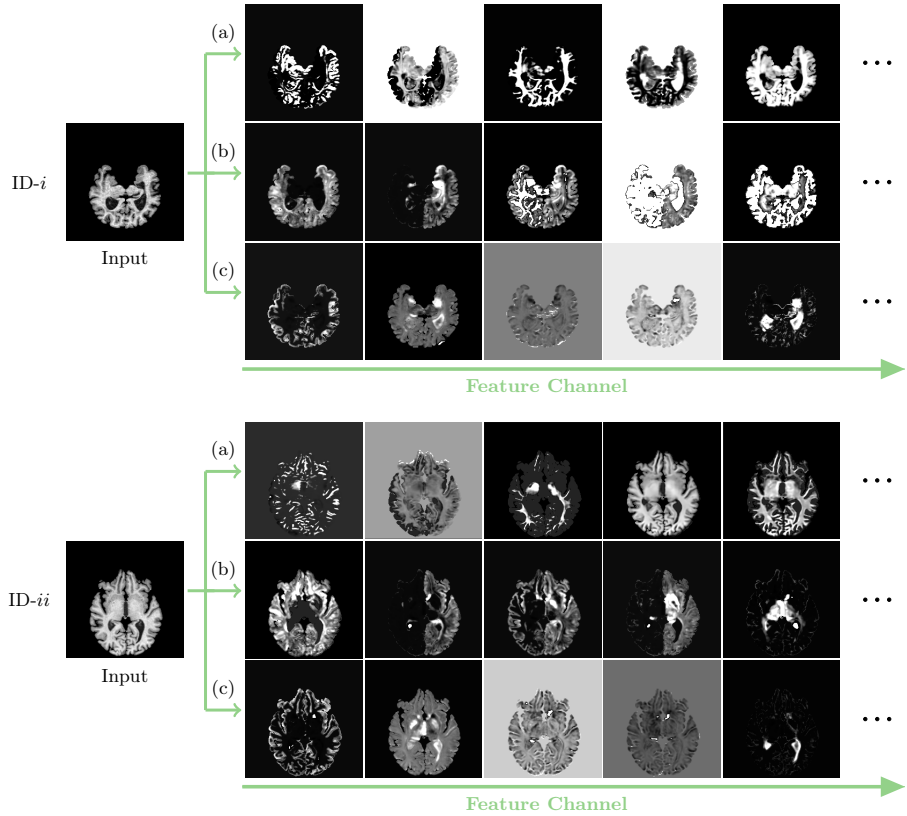

%% file: sec/appendix/fig/fig_task-recon.tex
\begin{figure*}[t]

\centering 

\resizebox{1.1\linewidth}{!}{
	\begin{tikzpicture}
        \hspace*{-0.8cm}
        
		\tikzstyle{myarrows}=[line width=0.8mm,draw=blue!50,-triangle 45,postaction={draw, line width=0.05mm, shorten >=0.02mm, -}]
		\tikzstyle{mylines}=[line width=0.8mm]
  


	\pgfmathsetmacro{\shift}{-0.8}
 
	\node at (\shift+1.4, 5+0.15*3+0.2, 0.75*3) {\small MR};
	\node at (\shift+1.4, 5+0.15*3-0.2, 0.75*3) {\small (T1w)};
	\node at (\shift+1.4, 3.2+0.15*3+0.2, 0.75*3) {\small MR};
	\node at (\shift+1.4, 3.2+0.15*3-0.2, 0.75*3) {\small (T2w)}; 
	\node at (\shift+1.4, 1.4+0.15*3+0.2, 0.75*3) {\small MR};
	\node at (\shift+1.4, 1.4+0.15*3-0.2, 0.75*3) {\small (FLAIR)}; 
	\node at (\shift+1.4, -0.43+0.15*3, 0.75*3) {\small CT};

	\pgfmathsetmacro{\shift}{-3.2}
	\node at (5.2+\shift+0.1*3, 6.3+0.15*3, 0.75*3) {\small Input};
	\node at (7.2+\shift+0.1*3, 6.3+0.15*3, 0.75*3) {\texttt{\small SynthSR}~\cite{Iglesias2023SynthSRAP}};
	\node at (9.2+\shift+0.1*3, 6.3+0.15*3, 0.75*3) {\small \texttt{CIFL}~\cite{chua2023contrast}};
	\node at (11.2+\shift+0.1*3, 6.3+0.15*3, 0.75*3) {\small \texttt{SCRATCH}};
	\node at (13.2+\shift+0.1*3, 6.3+0.15*3, 0.75*3) {\small \texttt{Brain-ID}};
	\node at (15.2+\shift+0.1*3, 6.3+0.15*3, 0.75*3) {\small Ground Truth};

	\node at (5.2+\shift+0.1*3, 5+0.15*3, 0.75*3) {\includegraphics[width=0.16\textwidth]{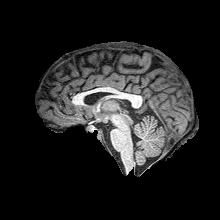}};
	\node at (7.2+\shift+0.1*3, 5+0.15*3, 0.75*3) {\includegraphics[width=0.16\textwidth]{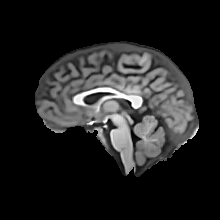}};
        \draw[fill=none, draw=mint!150, line width=0.4mm](7.2+\shift+0.1*3+0.2, 5+0.15*3+0.45, 0.75*3) circle (0.25);
        \draw[fill=none, draw=mint!150, line width=0.4mm](7.2+\shift+0.1*3+0.35, 5+0.15*3-0.25, 0.75*3) circle (0.25);
	\node at (9.2+\shift+0.1*3, 5+0.15*3, 0.75*3) {\includegraphics[width=0.16\textwidth]{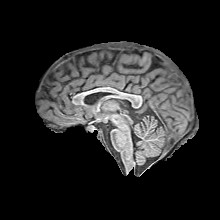}};
        \draw[fill=none, draw=mint!150, line width=0.4mm](9.2+\shift+0.1*3+0.2, 5+0.15*3+0.45, 0.75*3) circle (0.25);
	\node at (11.2+\shift+0.1*3, 5+0.15*3, 0.75*3) {\includegraphics[width=0.16\textwidth]{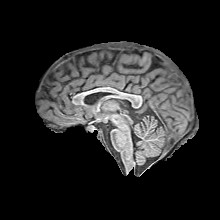}};
        \draw[fill=none, draw=mint!150, line width=0.4mm](11.2+\shift+0.1*3+0.2, 5+0.15*3+0.45, 0.75*3) circle (0.25);
	\node at (13.2+\shift+0.1*3, 5+0.15*3, 0.75*3) {\includegraphics[width=0.16\textwidth]{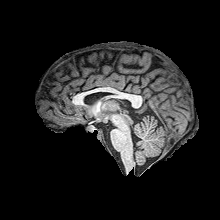}};
	\node at (15.2+\shift+0.1*3, 5+0.15*3, 0.75*3) {\includegraphics[width=0.16\textwidth]{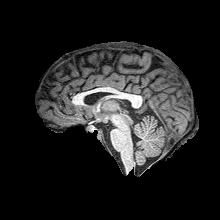}};

	\node at (5.2+\shift+0.1*3, 3.2+0.15*3, 0.75*3) {\includegraphics[width=0.16\textwidth]{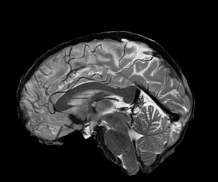}};
	\node at (7.2+\shift+0.1*3, 3.2+0.15*3, 0.75*3) {\includegraphics[width=0.16\textwidth]{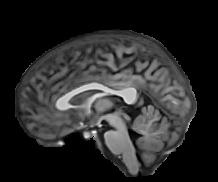}};
        \draw[fill=none, draw=mint!150, line width=0.4mm](7.2+\shift+0.1*3+0.4, 3.2+0.15*3-0.38, 0.75*3) circle (0.29);
	\node at (9.2+\shift+0.1*3, 3.2+0.15*3, 0.75*3) {\includegraphics[width=0.16\textwidth]{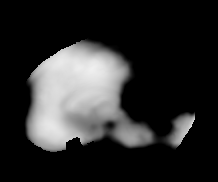}};
	\node at (11.2+\shift+0.1*3, 3.2+0.15*3, 0.75*3) {\includegraphics[width=0.16\textwidth]{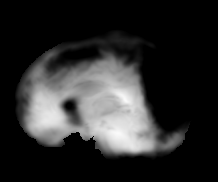}};
	\node at (13.2+\shift+0.1*3, 3.2+0.15*3, 0.75*3) {\includegraphics[width=0.16\textwidth]{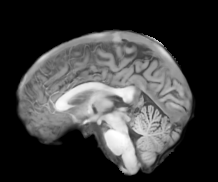}};
	\node at (15.2+\shift+0.1*3, 3.2+0.15*3, 0.75*3) {\includegraphics[width=0.16\textwidth]{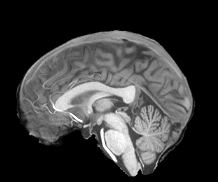}};

	\node at (5.2+\shift+0.1*3, 1.4+0.15*3, 0.75*3) {\includegraphics[width=0.16\textwidth]{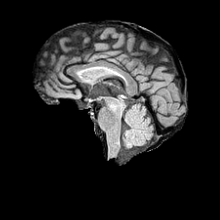}};
	\node at (7.2+\shift+0.1*3, 1.4+0.15*3, 0.75*3) {\includegraphics[width=0.16\textwidth]{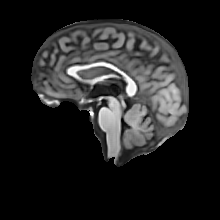}};
        \draw[fill=none, draw=mint!150, line width=0.4mm](7.2+\shift+0.1*3+0.25, 1.4+0.15*3-0.15, 0.75*3) circle (0.26);
	\node at (9.2+\shift+0.1*3, 1.4+0.15*3, 0.75*3) {\includegraphics[width=0.16\textwidth]{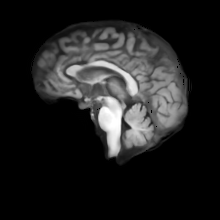}};
        \draw[fill=none, draw=mint!150, line width=0.4mm](9.2+\shift+0.1*3+0.25, 1.4+0.15*3-0.15, 0.75*3) circle (0.26);
	\node at (11.2+\shift+0.1*3, 1.4+0.15*3, 0.75*3) {\includegraphics[width=0.16\textwidth]{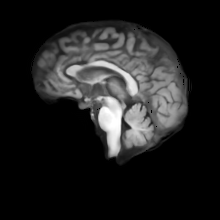}};
        \draw[fill=none, draw=mint!150, line width=0.4mm](11.2+\shift+0.1*3+0.25, 1.4+0.15*3-0.15, 0.75*3) circle (0.26);
	\node at (13.2+\shift+0.1*3, 1.4+0.15*3, 0.75*3) {\includegraphics[width=0.16\textwidth]{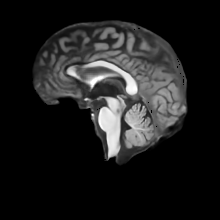}};
	\node at (15.2+\shift+0.1*3, 1.4+0.15*3, 0.75*3) {\includegraphics[width=0.16\textwidth]{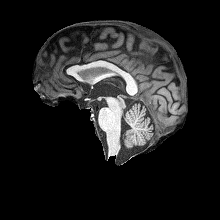}};

	\node at (5.2+\shift+0.1*3, -0.43+0.15*3, 0.75*3) {\includegraphics[width=0.16\textwidth]{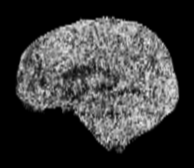}};
	\node at (7.2+\shift+0.1*3, -0.43+0.15*3, 0.75*3) {\includegraphics[width=0.16\textwidth]{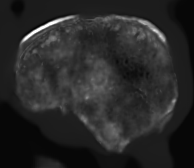}};
	\node at (9.2+\shift+0.1*3, -0.43+0.15*3, 0.75*3) {\includegraphics[width=0.16\textwidth]{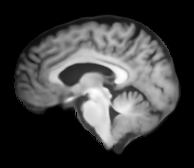}};
        \draw[fill=none, draw=mint!150, line width=0.4mm](9.2+\shift+0.1*3+0.35, -0.43+0.15*3-0.3, 0.75*3) circle (0.28);
	\node at (11.2+\shift+0.1*3, -0.43+0.15*3, 0.75*3) {\includegraphics[width=0.16\textwidth]{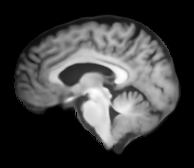}};
        \draw[fill=none, draw=mint!150, line width=0.4mm](11.2+\shift+0.1*3+0.35, -0.43+0.15*3-0.3, 0.75*3) circle (0.28);
	\node at (13.2+\shift+0.1*3, -0.43+0.15*3, 0.75*3) {\includegraphics[width=0.16\textwidth]{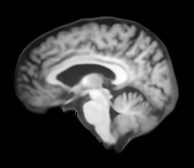}};
	\node at (15.2+\shift+0.1*3, -0.43+0.15*3, 0.75*3) {\includegraphics[width=0.16\textwidth]{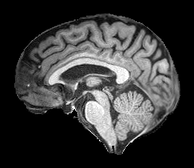}};

  
	\end{tikzpicture}
	}  
    \caption{Qualitative comparisons on the downstream task of anatomy reconstruction/contrast synthesis, between \texttt{Brain-ID}, the baseline \texttt{SCRATCH}, and the state-of-the-art methods \texttt{CIFL}~\cite{chua2023contrast}, \texttt{SynthSR}~\cite{Iglesias2023SynthSRAP}. Each row presents the comparison results of inputs with their respective modality/contrast, as indicated in the listing. The \textcolor{mint!200}{mint} circles highlight some less noticeable details.} 

	 \label{app-fig: task-recon}
\end{figure*}

%% file: sec/appendix/fig/fig_task-sr.tex
\begin{figure*}[t]

\centering 

\resizebox{1.15\linewidth}{!}{
	\begin{tikzpicture}
        \hspace*{-1.55cm}
        
		\tikzstyle{myarrows}=[line width=0.8mm,draw=blue!50,-triangle 45,postaction={draw, line width=0.05mm, shorten >=0.02mm, -}]
		\tikzstyle{mylines}=[line width=0.8mm]
  


	\pgfmathsetmacro{\shift}{-0.8}
 
	\node at (\shift+1.4, 5+0.15*3+0.2, 0.75*3) {\small \textcolor{white}{MR}};
	\node at (\shift+1.4, 5+0.15*3-0.2, 0.75*3) {\small \textcolor{white}{(T1w)}};
	\node at (\shift+1.4, 2.75+0.15*3+0.2, 0.75*3) {\small \textcolor{white}{MR}};
	\node at (\shift+1.4, 2.75+0.15*3-0.2, 0.75*3) {\small \textcolor{white}{(T2w)}}; 
	\node at (\shift+1.4, 0.4+0.15*3+0.2, 0.75*3) {\small \textcolor{white}{MR}};
	\node at (\shift+1.4, 0.4+0.15*3-0.2, 0.75*3) {\small \textcolor{white}{(FLAIR)}};

	\pgfmathsetmacro{\shift}{-3.2}
	\node at (5.2+\shift+0.1*3, 6.3+0.15*3+0.1, 0.75*3) {\small Input};
	\node at (7.2+\shift+0.1*3, 6.3+0.15*3+0.1, 0.75*3) {\small \texttt{SynthSR}~\cite{Iglesias2023SynthSRAP}};
	\node at (9.2+\shift+0.1*3, 6.3+0.15*3+0.1, 0.75*3) {\small \texttt{CIFL}~\cite{chua2023contrast}};
	\node at (11.2+\shift+0.1*3, 6.3+0.15*3+0.1, 0.75*3) {\small \texttt{SCRATCH}};
	\node at (13.2+\shift+0.1*3, 6.3+0.15*3+0.1, 0.75*3) {\small \texttt{Brain-ID}};
	\node at (15.2+\shift+0.1*3, 6.3+0.15*3+0.1, 0.75*3) {\small Ground Truth};

	\node at (5.2+\shift+0.1*3, 5+0.15*3, 0.75*3) {\includegraphics[width=0.16\textwidth]{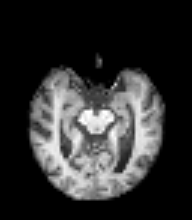}};
	\node at (7.2+\shift+0.1*3, 5+0.15*3, 0.75*3) {\includegraphics[width=0.16\textwidth]{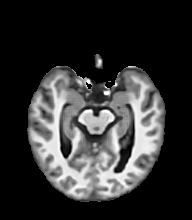}};
        \draw[fill=none, draw=mint!150, line width=0.4mm](7.2+\shift+0.1*3-0.55, 5+0.15*3-0.15, 0.75*3) circle (0.25);
	\node at (9.2+\shift+0.1*3, 5+0.15*3, 0.75*3) {\includegraphics[width=0.16\textwidth]{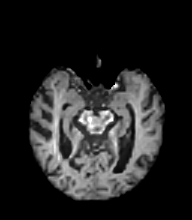}};
        \draw[fill=none, draw=mint!150, line width=0.4mm](9.2+\shift+0.1*3-0., 5+0.15*3-0.65, 0.75*3) circle (0.27);
	\node at (11.2+\shift+0.1*3, 5+0.15*3, 0.75*3) {\includegraphics[width=0.16\textwidth]{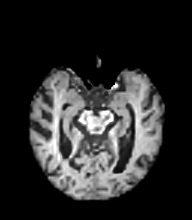}};
        \draw[fill=none, draw=mint!150, line width=0.4mm](11.2+\shift+0.1*3-0., 5+0.15*3-0.65, 0.75*3) circle (0.27);
	\node at (13.2+\shift+0.1*3, 5+0.15*3, 0.75*3) {\includegraphics[width=0.16\textwidth]{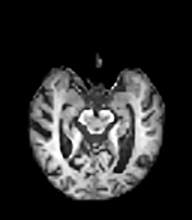}};
	\node at (15.2+\shift+0.1*3, 5+0.15*3, 0.75*3) {\includegraphics[width=0.16\textwidth]{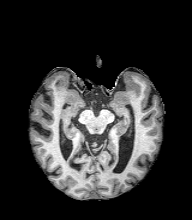}};

	\node at (5.2+\shift+0.1*3, 2.75+0.15*3, 0.75*3) {\includegraphics[width=0.16\textwidth]{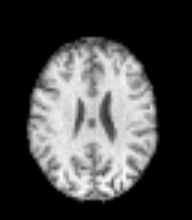}};
	\node at (7.2+\shift+0.1*3, 2.75+0.15*3, 0.75*3) {\includegraphics[width=0.16\textwidth]{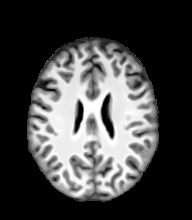}};
        \draw[fill=none, draw=mint!150, line width=0.4mm](7.2+\shift+0.1*3-0.55, 2.75+0.15*3+0.2, 0.75*3) circle (0.25);
	\node at (9.2+\shift+0.1*3, 2.75+0.15*3, 0.75*3) {\includegraphics[width=0.16\textwidth]{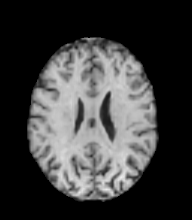}};
        \draw[fill=none, draw=mint!150, line width=0.4mm](9.2+\shift+0.1*3-0.15, 2.75+0.15*3-0.05, 0.75*3) circle (0.27);
	\node at (11.2+\shift+0.1*3, 2.75+0.15*3, 0.75*3) {\includegraphics[width=0.16\textwidth]{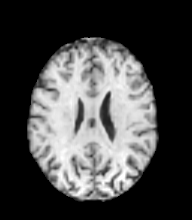}};
        \draw[fill=none, draw=mint!150, line width=0.4mm](11.2+\shift+0.1*3-0.15, 2.75+0.15*3-0.05, 0.75*3) circle (0.27);
	\node at (13.2+\shift+0.1*3, 2.75+0.15*3, 0.75*3) {\includegraphics[width=0.16\textwidth]{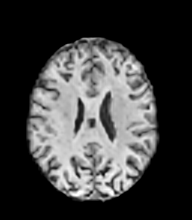}};
	\node at (15.2+\shift+0.1*3, 2.75+0.15*3, 0.75*3) {\includegraphics[width=0.16\textwidth]{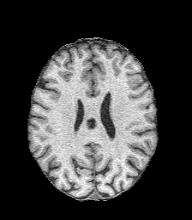}};

	\node at (5.2+\shift+0.1*3, 0.4+0.15*3, 0.75*3) {\includegraphics[width=0.16\textwidth]{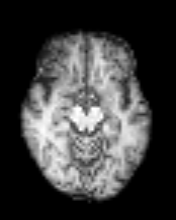}};
	\node at (7.2+\shift+0.1*3, 0.4+0.15*3, 0.75*3) {\includegraphics[width=0.16\textwidth]{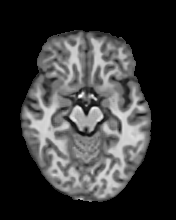}};
        \draw[fill=none, draw=mint!150, line width=0.4mm](7.2+\shift+0.1*3+0.25, 0.4+0.15*3-0.15, 0.75*3) circle (0.26);
	\node at (9.2+\shift+0.1*3, 0.4+0.15*3, 0.75*3) {\includegraphics[width=0.16\textwidth]{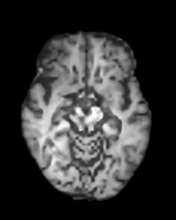}};
        \draw[fill=none, draw=mint!150, line width=0.4mm](9.2+\shift+0.1*3+0.25, 0.4+0.15*3-0.15, 0.75*3) circle (0.26);
	\node at (11.2+\shift+0.1*3, 0.4+0.15*3, 0.75*3) {\includegraphics[width=0.16\textwidth]{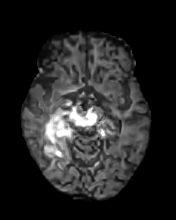}};
        \draw[fill=none, draw=mint!150, line width=0.4mm](11.2+\shift+0.1*3+0.25, 0.4+0.15*3-0.15, 0.75*3) circle (0.26);
	\node at (13.2+\shift+0.1*3, 0.4+0.15*3, 0.75*3) {\includegraphics[width=0.16\textwidth]{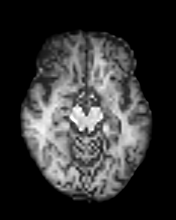}};
	\node at (15.2+\shift+0.1*3, 0.4+0.15*3, 0.75*3) {\includegraphics[width=0.16\textwidth]{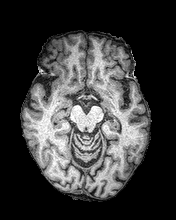}};


	\end{tikzpicture}
	}  
\caption{Qualitative comparisons on the downstream task of image super-resolution, between \texttt{Brain-ID}, the baseline \texttt{SCRATCH}, and the state-of-the-art methods \texttt{CIFL}~\cite{chua2023contrast}, \texttt{SynthSR}~\cite{Iglesias2023SynthSRAP}. Each row corresponds to a different testing subject. The \textcolor{mint!200}{mint} circles highlight some less noticeable details.} 

	 \label{app-fig: task-sr}
\end{figure*}
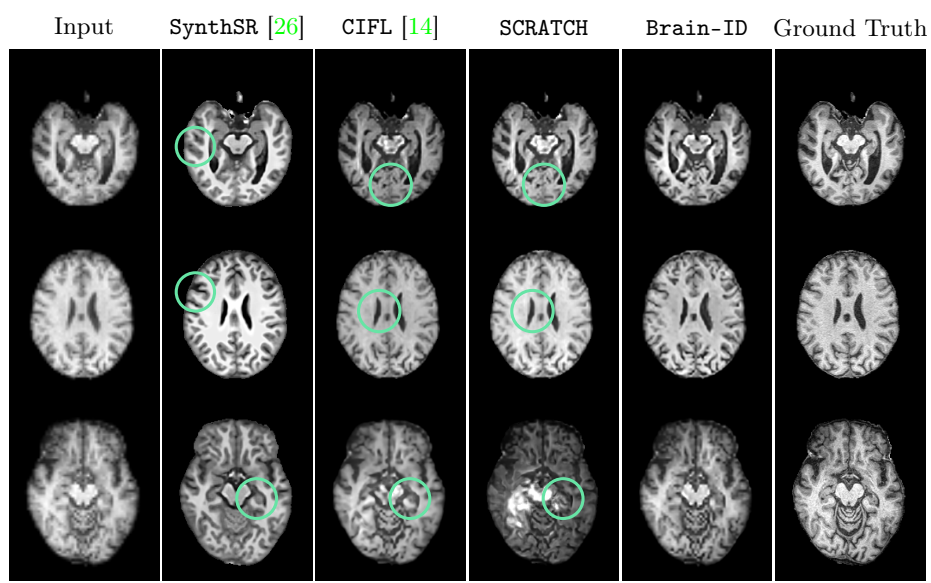

%% file: sec/appendix/fig/fig_task-seg.tex
\begin{figure*}[t]

\centering 

\resizebox{1.1\linewidth}{!}{
	\begin{tikzpicture}
        \hspace*{-0.8cm}
        
		\tikzstyle{myarrows}=[line width=0.8mm,draw=blue!50,-triangle 45,postaction={draw, line width=0.05mm, shorten >=0.02mm, -}]
		\tikzstyle{mylines}=[line width=0.8mm]
  


	\pgfmathsetmacro{\shift}{-0.8}
 
	\node at (\shift+1.4, 5+0.15*3+0.2, 0.75*3) {\small MR};
	\node at (\shift+1.4, 5+0.15*3-0.2, 0.75*3) {\small (T1w)};
	\node at (\shift+1.4, 3+0.15*3+0.2, 0.75*3) {\small MR};
	\node at (\shift+1.4, 3+0.15*3-0.2, 0.75*3) {\small (T2w)}; 
	\node at (\shift+1.4, 1+0.15*3+0.2, 0.75*3) {\small MR};
	\node at (\shift+1.4, 1+0.15*3-0.2, 0.75*3) {\small (FLAIR)}; 
	\node at (\shift+1.4, -1+0.15*3, 0.75*3) {\small CT};

	\pgfmathsetmacro{\shift}{-3.2}
	\node at (5.2+\shift+0.1*3, 6.3+0.15*3, 0.75*3) {\small Input};
	\node at (7.2+\shift+0.1*3, 6.3+0.15*3, 0.75*3) {\small \texttt{SAMSEG}~\cite{Cerri2020ACM}};
	\node at (9.2+\shift+0.1*3, 6.3+0.15*3, 0.75*3) {\small \texttt{CIFL}~\cite{chua2023contrast}};
	\node at (11.2+\shift+0.1*3, 6.3+0.15*3, 0.75*3) {\small \texttt{SCRATCH}};
	\node at (13.2+\shift+0.1*3, 6.3+0.15*3, 0.75*3) {\small \texttt{Brain-ID}};
	\node at (15.2+\shift+0.1*3, 6.3+0.15*3, 0.75*3) {\small Ground Truth};

	\node at (5.2+\shift+0.1*3, 5+0.15*3, 0.75*3) {\includegraphics[width=0.16\textwidth]{fig/task/seg-adni3-t1-255/input.png}};
	\node at (7.2+\shift+0.1*3, 5+0.15*3, 0.75*3) {\includegraphics[width=0.16\textwidth]{fig/task/seg-adni3-t1-255/samseg.png}};
        \draw[fill=none, draw=mint!150, line width=0.4mm](7.2+\shift+0.1*3-0.2, 5+0.15*3+0.35, 0.75*3) circle (0.25);
	\node at (9.2+\shift+0.1*3, 5+0.15*3, 0.75*3) {\includegraphics[width=0.16\textwidth]{fig/task/seg-adni3-t1-255/cifl.png}};
        \draw[fill=none, draw=mint!150, line width=0.4mm](9.2+\shift+0.1*3-0.07, 5+0.15*3-0.09, 0.75*3) circle (0.23);
	\node at (11.2+\shift+0.1*3, 5+0.15*3, 0.75*3) {\includegraphics[width=0.16\textwidth]{fig/task/seg-adni3-t1-255/scr.png}};
        \draw[fill=none, draw=mint!150, line width=0.4mm](11.2+\shift+0.1*3-0.07, 5+0.15*3-0.09, 0.75*3) circle (0.23);
	\node at (13.2+\shift+0.1*3, 5+0.15*3, 0.75*3) {\includegraphics[width=0.16\textwidth]{fig/task/seg-adni3-t1-255/brain-id.png}};
	\node at (15.2+\shift+0.1*3, 5+0.15*3, 0.75*3) {\includegraphics[width=0.16\textwidth]{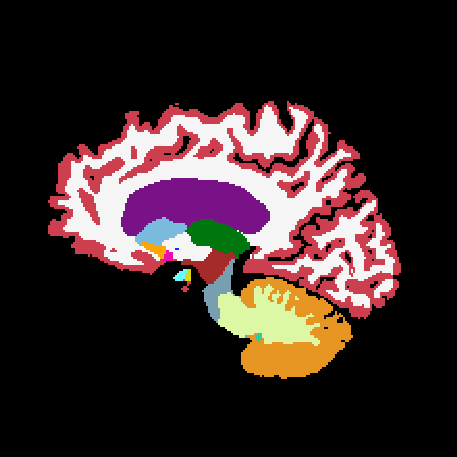}};

	\node at (5.2+\shift+0.1*3, 3+0.15*3, 0.75*3) {\includegraphics[width=0.16\textwidth]{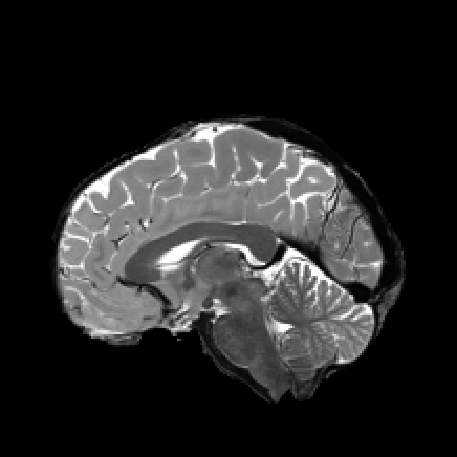}};
	\node at (7.2+\shift+0.1*3, 3+0.15*3, 0.75*3) {\includegraphics[width=0.16\textwidth]{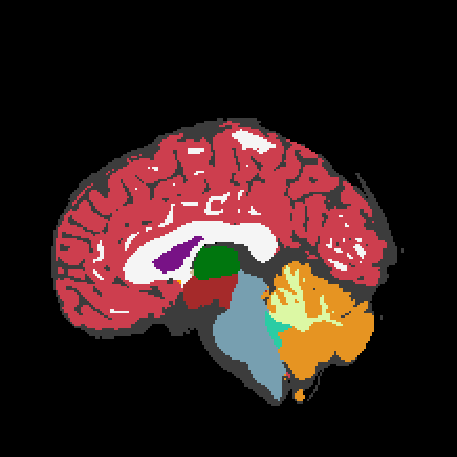}};
        \draw[fill=none, draw=mint!150, line width=0.4mm](7.2+\shift+0.1*3-0.05, 3+0.15*3+0.08, 0.75*3) circle (0.26);
	\node at (9.2+\shift+0.1*3, 3+0.15*3, 0.75*3) {\includegraphics[width=0.16\textwidth]{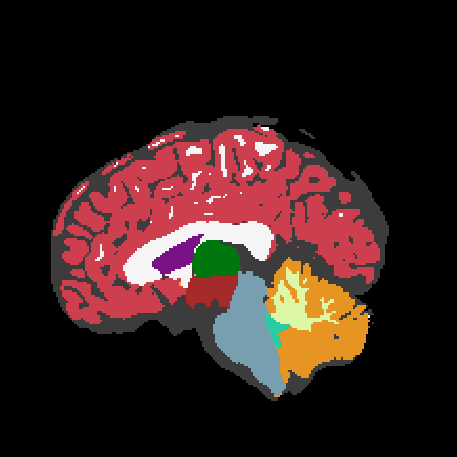}};
        \draw[fill=none, draw=mint!150, line width=0.4mm](9.2+\shift+0.1*3-0.2, 3+0.15*3-0.08, 0.75*3) circle (0.25);
	\node at (11.2+\shift+0.1*3, 3+0.15*3, 0.75*3) {\includegraphics[width=0.16\textwidth]{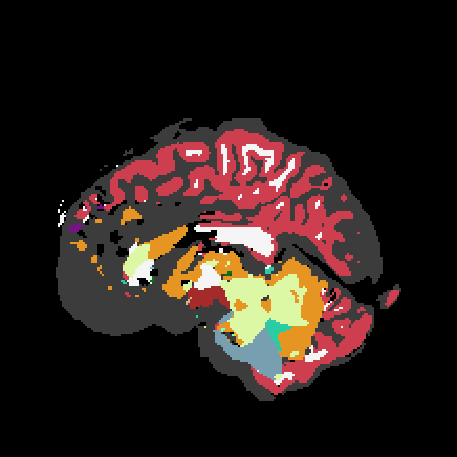}};
	\node at (13.2+\shift+0.1*3, 3+0.15*3, 0.75*3) {\includegraphics[width=0.16\textwidth]{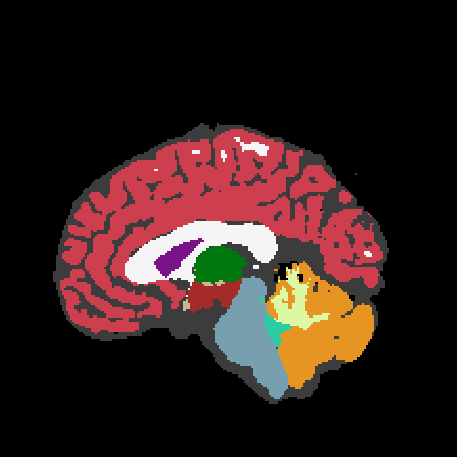}};
	\node at (15.2+\shift+0.1*3, 3+0.15*3, 0.75*3) {\includegraphics[width=0.16\textwidth]{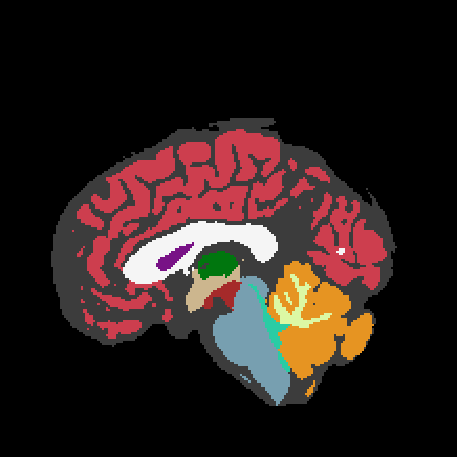}};

	\node at (5.2+\shift+0.1*3, 1+0.15*3, 0.75*3) {\includegraphics[width=0.16\textwidth]{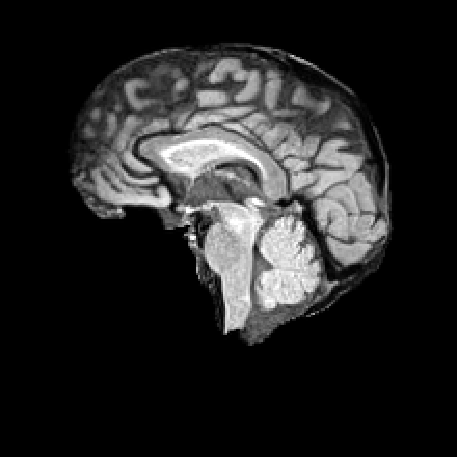}};
	\node at (7.2+\shift+0.1*3, 1+0.15*3, 0.75*3) {\includegraphics[width=0.16\textwidth]{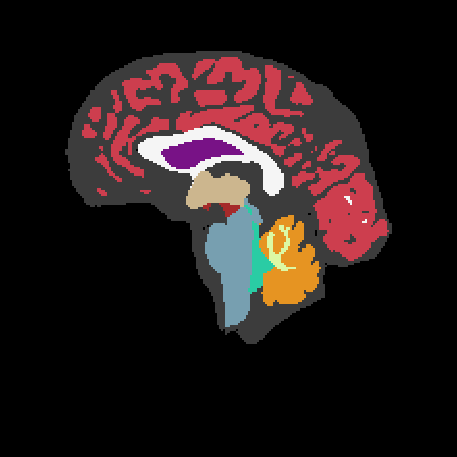}};
        \draw[fill=none, draw=mint!150, line width=0.4mm](7.2+\shift+0.1*3-0.1, 1+0.15*3+0.6, 0.75*3) circle (0.26);
	\node at (9.2+\shift+0.1*3, 1+0.15*3, 0.75*3) {\includegraphics[width=0.16\textwidth]{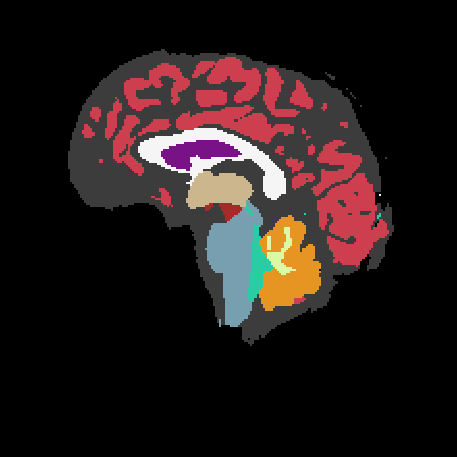}};
        \draw[fill=none, draw=mint!150, line width=0.4mm](9.2+\shift+0.1*3-0.1, 1+0.15*3+0.6, 0.75*3) circle (0.26);
	\node at (11.2+\shift+0.1*3, 1+0.15*3, 0.75*3) {\includegraphics[width=0.16\textwidth]{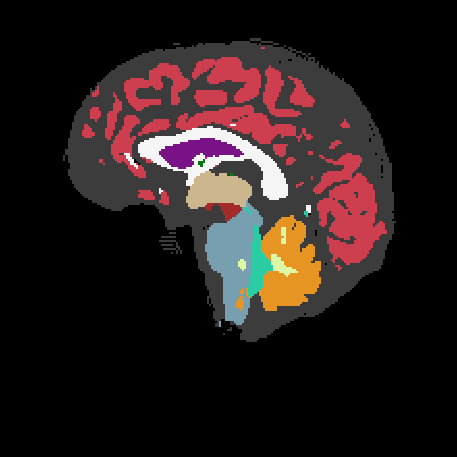}};
        \draw[fill=none, draw=mint!150, line width=0.4mm](11.2+\shift+0.1*3-0.1, 1+0.15*3+0.6, 0.75*3) circle (0.26);
        \draw[fill=none, draw=mint!150, line width=0.4mm](11.2+\shift+0.1*3+0.08, 1+0.15*3-0.15, 0.75*3) circle (0.26);
	\node at (13.2+\shift+0.1*3, 1+0.15*3, 0.75*3) {\includegraphics[width=0.16\textwidth]{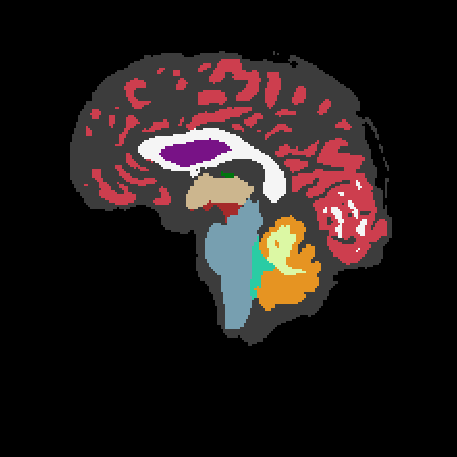}};
	\node at (15.2+\shift+0.1*3, 1+0.15*3, 0.75*3) {\includegraphics[width=0.16\textwidth]{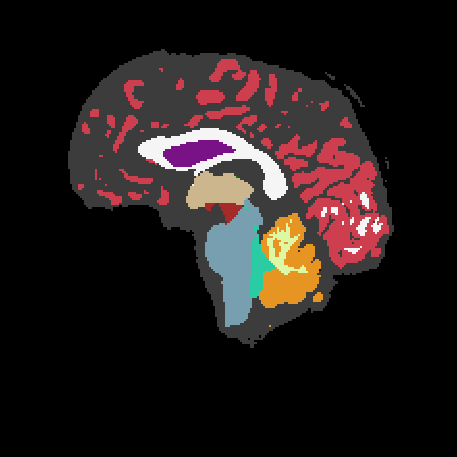}};

	\node at (5.2+\shift+0.1*3, -1+0.15*3, 0.75*3) {\includegraphics[width=0.16\textwidth]{fig/app-seg/oasis-ct-30010/input.png}};
	\node at (7.2+\shift+0.1*3, -1+0.15*3, 0.75*3) {\includegraphics[width=0.16\textwidth]{fig/app-seg/oasis-ct-30010/samseg.png}};
	\node at (9.2+\shift+0.1*3, -1+0.15*3, 0.75*3) {\includegraphics[width=0.16\textwidth]{fig/app-seg/oasis-ct-30010/cifl.png}};
        \draw[fill=none, draw=mint!150, line width=0.4mm](9.2+\shift+0.1*3-0.2, -1+0.15*3+0.2, 0.75*3) circle (0.25);
	\node at (11.2+\shift+0.1*3, -1+0.15*3, 0.75*3) {\includegraphics[width=0.16\textwidth]{fig/app-seg/oasis-ct-30010/scr.png}};
        \draw[fill=none, draw=mint!150, line width=0.4mm](11.2+\shift+0.1*3-0.2, -1+0.15*3+0.2, 0.75*3) circle (0.25);
        \draw[fill=none, draw=mint!150, line width=0.4mm](11.2+\shift+0.1*3+0.05, -1+0.15*3-0.3, 0.75*3) circle (0.25);
	\node at (13.2+\shift+0.1*3, -1+0.15*3, 0.75*3) {\includegraphics[width=0.16\textwidth]{fig/app-seg/oasis-ct-30010/brain-id.png}};
	\node at (15.2+\shift+0.1*3, -1+0.15*3, 0.75*3) {\includegraphics[width=0.16\textwidth]{fig/app-seg/oasis-ct-30010/gt.png}};


	\end{tikzpicture}
	}  
\caption{Qualitative comparisons on the downstream task of brain segmentation, between \texttt{Brain-ID}, the baseline \texttt{SCRATCH}, and the state-of-the-art methods \texttt{CIFL}~\cite{chua2023contrast}, \texttt{SAMSEG}~\cite{Cerri2020ACM}. Each row presents the comparison results of inputs with their respective modality/contrast, as indicated in the listing. The \textcolor{mint!200}{mint} circles highlight some less noticeable details.} 

	 \label{app-fig: task-seg}
\end{figure*}